\documentclass[11pt,a4paper,twocolumn]{article}

\usepackage[margin=1in]{geometry}
\usepackage{amsmath,amssymb}
\usepackage{graphicx}
\usepackage{booktabs}
\usepackage{array}
\usepackage{authblk}
\usepackage{microtype}
\usepackage{makecell}
\usepackage{url}
\usepackage{xurl}
\usepackage{float}
\usepackage{orcidlink}
\usepackage{hyperref}
\usepackage{cleveref}
\hypersetup{
  colorlinks=true,
  linkcolor=blue,
  citecolor=blue,
  urlcolor=blue
}

\newcommand{\GeV}{\text{GeV}}
\newcommand{\MET}{\ensuremath{E_{\mathrm{T}}^{\mathrm{miss}}}}
\newcommand{\score}{\ensuremath{\mathcal{S}}}
\newcolumntype{L}[1]{>{\raggedright\arraybackslash}p{#1}}

\title{%
  Mono-\texorpdfstring{$Z$}{Z} Dark Matter Search with Neural Spline Flows \\ Using CMS Run~2015D Open Data
}

\author[1,$\dagger$]{Hitesh Rasineni\,\orcidlink{0009-0003-4958-0915}}
\author[2,$\dagger$]{Chebrolu Bhavishya\,\orcidlink{0009-0000-9770-0975}}

\affil[1]{VIT-AP University, Amaravati, 522241, India}
\affil[2]{Mohan Babu University, Tirupati, 517102, India}

\date{}

\begin{document}

\maketitle

\begingroup
\renewcommand{\thefootnote}{\fnsymbol{footnote}}
\footnotetext[2]{$\dagger$ These authors contributed equally and share first authorship.}
\endgroup

\begin{abstract}
  We report a search for dark matter~(DM) produced in association with a
  leptonically decaying $Z$~boson at $\sqrt{s}=13$~TeV, using CMS Run~2015D
  open data corresponding to an integrated luminosity of $2.32~\mathrm{fb}^{-1}$
  and simplified-model Monte Carlo simulation.
  Events are selected in the mono-$Z\to\ell^+\ell^-$ final state in two
  parallel lepton-flavor channels, $Z\to\mu^+\mu^-$ and $Z\to e^+e^-$,
  requiring an opposite-sign same-flavor dilepton pair with invariant mass
  $60 < m_{\ell\ell} < 120\,\GeV$.
  A signal region is defined by $\MET \ge 50\,\GeV$,
  $|\Delta\phi(\MET, Z)| > 2.5$, and $n_{\mathrm{jets}} \le 1$;
  a control region with $\MET < 50\,\GeV$ provides the SM density model used in
  the likelihood-ratio score.
  Forty kinematic observables are extracted from MINIAOD and MINIAODSIM inputs,
  cleaned with physics-motivated bounds and data-driven tail clipping, and
  reduced to a 37-dimensional feature vector comprising 35 physics quantities
  augmented with two binary jet-presence indicators.
  Five Neural Spline Flows~(NSFs)~\cite{Durkan2019,Papamakarios2021} are trained
  independently: two channel-specific SM flows learn the background density from
  control-region Drell--Yan events~(70\%/30\% train/validation split), and three
  mediator-specific DM flows learn the signal density from Monte Carlo samples
  with vector, axial-vector, and scalar $s$-channel mediators.
  For each mediator hypothesis $h$, the per-event test statistic is the
  log-likelihood ratio
  \begin{equation*}
    \mathcal{S}_h(\mathbf{x})
    = \log p(\mathbf{x}\mid\mathrm{DM}_h)
    - \log p(\mathbf{x}\mid\mathrm{SM}_{\ell\ell}),
  \end{equation*}
  which concentrates signal sensitivity across the full kinematic phase space
  without requiring a hard upper $\MET$ threshold.
  The fitted signal strengths are nonzero in the nominal fits, driven by a
  high-\MET\ background-modelling residual rather than evidence for a DM signal.
  Combining the $\mu\mu$ and $ee$ channels in a simultaneous SR+VR
  binned profile-likelihood fit, where the validation region
  ($50\le\MET<100\,\GeV$) provides an independent background
  normalisation constraint, we set observed (expected) 95\%~CL upper
  limits on the signal-strength parameter of
  $\mu<0.0177$ ($0.0018$) for the scalar mediator,
  $\mu<0.0362$ ($0.0039$) for the vector mediator, and
  $\mu<0.0498$ ($0.0069$) for the axial-vector mediator,
  corresponding to cross-section upper limits of
  $\sigma_{95}<1.76\times10^{-9}\,\mathrm{pb}$ (scalar),
  $\sigma_{95}<9.90\times10^{-4}\,\mathrm{pb}$ (vector), and
  $\sigma_{95}<9.24\times10^{-2}\,\mathrm{pb}$ and
  $7.89\times10^{-3}\,\mathrm{pb}$ (axial-vector, two benchmark points).
  The observed limits exceed the expected by factors of roughly 7--12 across all
  three mediator hypotheses and three independent background model
  variants, driven by a residual background-modelling discrepancy in
  the high-\MET\ tail ($\MET\ge100\,\GeV$, 2.4--2.5\% of SR events)
  rather than evidence of a DM signal.
  To our knowledge this is the first application of Neural Spline Flow
  likelihood-ratio scoring to a mono-$Z$ DM search on CMS open Run~2015D data
  in both the $\mu\mu$ and $ee$ channels simultaneously.
  \end{abstract}

\section{Introduction}
\label{sec:introduction}

Astrophysical evidence points to a non-baryonic dark component of the Universe, but the
microscopic identity of dark matter (DM) remains unknown.
Among the experimental approaches, high-energy colliders can produce DM in controlled
environments and probe its couplings to Standard Model (SM) particles through missing
transverse momentum ($\MET$) recoiling against visible final-state objects~\cite{Goodman2011}.
The mono-$Z$ topology---a leptonically decaying $Z$ boson balanced by $\MET$---offers a
relatively clean signature: the dilepton system constrains the $Z$ mass, electroweak
backgrounds are smaller than in hadronic mono-jet searches, and the channel is sensitive
to simplified models with $s$-channel mediators that also radiate an on-shell $Z$
boson~\cite{Bell2012,Carpenter2013,DMForum2015}.

Published LHC searches in this final state have progressed from early reinterpretations
and multivariate projections to full Run~2 results from CMS and ATLAS~\cite{Alves2015,CMS2017MonoZ,CMS2021MonoZ}.
These analyses typically bin events in $\MET$ or related kinematic variables, fit parametric
or template background models, and compare data to signal hypotheses defined within the
ATLAS/CMS Dark Matter Forum benchmark framework~\cite{DMForum2015}.
Complementary mono-$Z$ and mono-$Z'$ phenomenology has also been studied through dilepton
mass spectra, angular distributions, and resonance-plus-$\MET$ topologies~\cite{Autran2015,Altmannshofer2015,Yang2017,Elgammal2024}.

We investigate an alternative strategy that keeps the mono-$Z$ event selection but replaces
hand-crafted discriminant variables with a likelihood-ratio score built from learned event
densities.
Neural Spline Flows (NSFs)~\cite{Durkan2019,Papamakarios2021} model the SM background and
three DM mediator hypotheses as continuous distributions over a fixed set of cleaned
kinematic features; the per-event score is the log-density difference between the selected
DM hypothesis and the channel-specific SM hypothesis.
Normalizing flows supply exact density evaluation for correlated inputs~\cite{Papamakarios2021,Reyes2022},
and surjective or simulation-aware extensions have been explored for collider event
modeling and anomaly detection~\cite{Verheyen2022,Andreassen2020}.
Our implementation uses CMS Run~2015D open data~\cite{CERNOpenDataDoubleMuon,CERNOpenDataDoubleEG}
for SM-dominated control samples and publicly released MonoZToLL MC samples with
vector, axial-vector, and scalar mediators~\cite{CERNOpenDataMonoZVector,CERNOpenDataMonoZAxialMx10Mv20,CERNOpenDataMonoZAxialMx50Mv200,CERNOpenDataMonoZScalar}
for signal studies, with one DM flow trained per mediator model and identical
physics-feature definitions in the $\mu\mu$ and $ee$ channels.

Section~\ref{sec:literature} reviews the mono-$Z$ literature and related machine-learning
methods in more detail.
Section~\ref{sec:objective} states the analysis regions and score definition used here;
Section~\ref{sec:outline} summarizes our contributions and the paper structure.
All region definitions, training splits, and numerical results reported below follow our
analysis pipeline unless explicitly attributed to external work.

\subsection{Analysis objective}
\label{sec:objective}

The physics goal is to test whether the number and kinematics of events in a signal region
are consistent with SM production of $Z+\MET$, or require a DM contribution.
The working signal hypotheses are DM pair production with vector, axial-vector, and
scalar mediators, simulated with the same lepton selection as data.

Operationally, we partition events into a control region (CR) and a signal region (SR).
The CR ($\MET<50\,\GeV$) is dominated by Drell--Yan $Z$+jets and is used only to train
SM density models; it is disjoint from the SR by construction.
The SR applies $\MET\ge 50\,\GeV$, $|\Delta\phi(\MET,Z)|>2.5$, and $n_{\mathrm{jets}}\le 1$.
The SR imposes no explicit upper \MET\ requirement; in practice the
cleaned dataset caps \MET\ at approximately 200\,\GeV\
(Section~\ref{sec:datasets}), so events enter the fit through \MET\
values up to this bound. The validation interval ($50\le\MET<100\,\GeV$) additionally serves as a
background-normalisation sideband in the simultaneous SR+VR fit: it constrains the per-channel normalisation nuisances using VR data independently of the SR, providing a genuine sideband-based background prediction.

Two SM flows are trained, one on CR double-muon events and one on CR double-electron events
(70\%/30\% train/validation split in each channel).
Three DM flows are trained, one each on the full vector, axial-vector, and scalar MC
samples, and are shared across lepton channels.
The branch \texttt{lepton\_flavor} ($11$ for electrons, $13$ for muons) selects the channel
but is not included in the NSF input vector.

For an event $\mathbf{x}$ in the SR and mediator hypothesis
$h\in\{\mathrm{vector},\mathrm{axial},\mathrm{scalar}\}$,
\begin{equation}
  \score_h(\mathbf{x}) =
  \log p(\mathbf{x}\mid\mathrm{DM}_h) -
  \log p(\mathbf{x}\mid\mathrm{SM}_{\mu\mu\ \mathrm{or}\ ee}).
  \label{eq:score}
\end{equation}
Large positive values indicate kinematics more characteristic of the trained DM sample
than of the corresponding SM CR model.

Before interpreting the SR, we check the SM-only flow in a 50--100~\GeV\ \MET\
validation band to verify stable channel-specific SM behavior. The primary search
constructs six SR score arrays, $\score_h^{\mu\mu}$ and $\score_h^{ee}$ for the
three mediator hypotheses, and fits three combined binned $\mu\mu$+ee likelihoods,
one per signal hypothesis~\cite{Cowan2011}.

\subsection{Contributions and paper outline}
\label{sec:outline}

This paper documents:
\begin{enumerate}
  \item extraction of 40 event-level features from CMS open data and MonoZToLL MC;
  \item a cleaning stage with fixed bounds, quantile-based tail rules, and reference-aware
        caps for MC samples;
  \item CR/SR definitions, NSF training protocol, and likelihood-ratio search strategy;
  \item final SR profile-likelihood fits and CL$_s$ limits for the Run~2015D dataset.
\end{enumerate}

Section~\ref{sec:datasets} lists data and MC inputs.
Section~\ref{sec:preprocessing} describes feature extraction and cleaning.
Section~\ref{sec:regions} defines CR, validation, and SR event categories.
Section~\ref{sec:nsf} summarizes NSF architecture and training.
Section~\ref{sec:search} details the score, fit, and cross-checks.
Section~\ref{sec:results} presents distributions and limits.
Section~\ref{sec:summary} concludes.

\section{Literature Review}
\label{sec:literature}

\subsection{Collider constraints and mono-\texorpdfstring{$Z$}{Z} phenomenology}
\label{sec:lit-pheno}

Early collider studies showed that effective-field-theory (EFT) descriptions of DM
coupling to quarks and gluons can yield limits on interaction scales that complement,
and in some kinematic regimes surpass, direct-detection bounds---particularly for light
DM and spin-dependent scattering~\cite{Goodman2011}.
The mono-$Z$ channel was proposed as an electroweak counterpart to mono-jet and
mono-photon searches: a $Z$ boson recoils against a pair of stable DM particles, with
the $Z$ emitted from initial-state quarks or an internal line in a simplified
model~\cite{Bell2012}.
Ref.~\cite{Carpenter2013} extended mono-$X$ reinterpretations to events with a
reconstructed $Z\to\ell^+\ell^-$ system, deriving limits on EFT operators with quark
bilinear and electroweak-boson couplings from ATLAS $Z$+$\MET$ measurements.

Related signatures broaden the mono-$Z$ program beyond a single on-shell $Z$ recoiling
against DM alone.
Ref.~\cite{Autran2015} studied $\MET$ in association with a dilepton or dijet resonance
from a $Z'$ boson, highlighting sensitivity in mass regions where standard resonance
searches lose efficiency.
Ref.~\cite{Altmannshofer2015} pointed out that DM with scalar mediators can induce
loop-level box corrections to Drell--Yan production, producing a ``monocline'' feature
near $m_{\ell\ell}\sim 2m_\chi$ rather than a narrow resonance.
Ref.~\cite{Berlin2015} discussed pseudoscalar-portal fermion DM in two-Higgs-doublet
setups, where chiral couplings suppress direct detection while preserving thermal relic
targets.
Ref.~\cite{Liew2017} compared mono-$X$ searches with direct searches for the mediating
particle across several simplified models, finding that mono-$X$ is most competitive
when the spectrum is compressed or when electroweak bosons participate directly in the
mediator--DM coupling.
Ref.~\cite{Yang2017} showed that lepton angular distributions in the Collins--Soper frame
can distinguish spin-0, spin-1, and spin-2 mediator scenarios in mono-$Z$ production and
that shape information beyond inclusive $\MET$ can strengthen coupling limits.

\subsection{Simplified models, benchmarks, and experimental searches}
\label{sec:lit-experiment}

The ATLAS/CMS Dark Matter Forum report~\cite{DMForum2015} codified simplified models
---$s$-channel vector, axial-vector, scalar, and pseudoscalar mediators, $t$-channel
colored mediators, and EFT benchmarks---together with recommended parameter scans and
implementation guidance for Run~2 $\MET+X$ analyses.
These models underpin the MonoZToLL open-data samples used in our signal studies; vector,
axial-vector, and scalar mediator samples are each extracted and assigned a separate DM
flow for hypothesis-specific scoring~\cite{CERNOpenDataMonoZVector,CERNOpenDataMonoZAxialMx10Mv20,CERNOpenDataMonoZAxialMx50Mv200,CERNOpenDataMonoZScalar,DMForum2015}.

On the experimental side, CMS published mono-$Z$ searches at $\sqrt{s}=13$~TeV interpreting
spin-0 and spin-1 mediators, invisible Higgs decays, and other $\MET$+$\ell^+\ell^-$
scenarios~\cite{CMS2017MonoZ,CMS2021MonoZ}.
The most recent CMS result uses the full Run~2 dataset ($137~\mathrm{fb}^{-1}$) and sets
limits on mediator and DM masses for vector, axial-vector, scalar, and pseudoscalar
mediators, with comparisons to direct-detection cross sections~\cite{CMS2021MonoZ}.
Ref.~\cite{Alves2015} projected the mono-$Z$ reach at $13$~TeV with a multivariate
likelihood discriminant, reporting roughly a factor-of-two improvement over cut-and-count
methods once background normalization systematics exceed a few percent.
High-mass dilepton searches from ATLAS~\cite{ATLAS2016Dilepton,ATLAS2017Dilepton} bound
$Z'$ resonances and four-fermion contact interactions; while not mono-$Z$ analyses, they
constrain overlapping $Z$-associated new-physics scenarios probed in
Refs.~\cite{Autran2015,Elgammal2024}.
Ref.~\cite{Elgammal2024} applied CMS open data from Run~1 to dimuon angular distributions
at high $m_{\mu\mu}$, interpreting the data in a mono-$Z'$ simplified model and finding
agreement with SM Drell--Yan expectations.

\subsection{Density estimation and learning-based search methods}
\label{sec:lit-ml}

Normalizing flows map a simple base density to a target distribution through invertible
transformations, enabling both sampling and exact likelihood evaluation~\cite{Papamakarios2021}.
Neural Spline Flows replace elementwise affine couplings with monotonic rational-quadratic
splines, increasing expressivity while retaining analytic inversion~\cite{Durkan2019}.
Ref.~\cite{Reyes2022} examined how several flow architectures scale with input
dimensionality on toy HEP-like datasets, a relevant consideration for multivariate collider
features.
Ref.~\cite{Verheyen2022} introduced surjective and stochastic flow layers to handle
permutation symmetry, variable particle multiplicities, and discrete quantum numbers in
matrix-element-level and detector-level event generation, as well as anomaly-detection
studies.
Ref.~\cite{Andreassen2020} proposed simulation-assisted likelihood-free anomaly detection
(SALAD), reweighting background simulation to data in sidebands before supervised
learning in a signal-sensitive region---a hybrid of data-driven and simulation-based
approaches distinct from our explicit two-hypothesis density ratio.
Closely related anomaly-detection methods that use sideband density estimates include
ANODE~\cite{Nachman2020} and CATHODE~\cite{Hallin2021}; both construct signal-region
classifiers from control-region density ratios, a strategy adjacent to our direct NSF
log-ratio score.
Ref.~\cite{Alves2015} remains the closest prior mono-$Z$ reference for multivariate
discrimination, though it used engineered kinematic inputs and a likelihood function rather
than normalizing-flow densities trained separately on control-region data and signal MC.

Our analysis sits at the intersection of these threads: we adopt the mono-$Z$ simplified-model
context and CR/SR philosophy of the experimental and Forum literature~\cite{DMForum2015,CMS2021MonoZ},
but estimate SM and DM event densities with NSFs and use their log-ratio as the test
statistic, following asymptotic inference prescriptions~\cite{Cowan2011} in the signal region.
To our knowledge, a published mono-$Z$ DM search combining NSF-based likelihood ratios with
CMS open Run~2015D data in both $\mu\mu$ and $ee$ channels has not been reported; the
present work documents that pipeline through the final SR likelihood fit.

\section{Datasets and Signal Model}
\label{sec:datasets}

SM events come from CMS Run~2015D MINIAOD open data: the DoubleMuon primary dataset
(record~24127)~\cite{CERNOpenDataDoubleMuon} and the DoubleEG primary dataset
(record~24132)~\cite{CERNOpenDataDoubleEG}, both at $\sqrt{s}=13$~TeV.
These samples provide opposite-sign dilepton events after analyzer-level selection;
they are used as SM-dominated inputs for CR training and for the SR search.
The Run~2015D dataset corresponds to an integrated luminosity of approximately
$2.32~\mathrm{fb}^{-1}$.
The extracted local TTrees contain 1,738,984 selected DoubleMuon events split across four
output files and 1,418,467 selected DoubleEG events split across eight output files.
The stored SM data schema contains the 40 physics features plus audit-only provenance and
quality-control branches used to validate trigger and $b$-tag behavior; the row counts are
summarized in Appendix~\ref{app:rows}, and the complete feature schema with formulas is
given in Appendix~\ref{app:features}.

The signal inputs are MonoZToLL MINIAODSIM samples generated in the same
$\sqrt{s}=13$~TeV era for vector, axial-vector, and scalar mediators~\cite{DMForum2015}.
The vector sample is record~16630~\cite{CERNOpenDataMonoZVector}; the axial-vector inputs
combine records~16575 and~16597~\cite{CERNOpenDataMonoZAxialMx10Mv20,CERNOpenDataMonoZAxialMx50Mv200};
the scalar sample is record~16601~\cite{CERNOpenDataMonoZScalar}.
The extracted selected samples contain 27,602 vector-mediator events, 45,710 axial-vector
events, and 32,792 scalar-mediator events.
The raw MC inputs are organized as ROOT source files; the axial-vector
training set combines four source files (one for \(M_\chi=10\,\GeV\),
\(M_V=20\,\GeV\) and three for \(M_\chi=50\,\GeV\), \(M_V=200\,\GeV\)),
totaling about 100,000 raw entries before the final mono-$Z$ selection. The same
40 physics branches listed in Appendix~\ref{app:features} are written for data and MC; the
MC output adds only the auxiliary \texttt{gen\_weight} branch.
Although no explicit upper \MET\ bound is imposed in the selection,
the data-driven tail clipping applied during feature cleaning caps
\MET\ at approximately 200\,\GeV\ in the extracted dataset; events
above this value are absent from the SR score distribution.
The axial-vector signal sample combines two physically distinct
benchmark points --- $M_\chi=10\,\GeV$, $M_V=20\,\GeV$ ($\sigma=1.856\,\mathrm{pb}$)
and $M_\chi=50\,\GeV$, $M_V=200\,\GeV$ ($\sigma=0.158\,\mathrm{pb}$) ---
stored across four ROOT files. Because the two points differ in cross-section
by a factor of $\sim$12, a single effective cross-section would not correspond
to any real benchmark; cross-section limits are therefore reported separately
for each axial-vector point using the same fitted $\mu$.
Generator-level cross-sections were extracted directly from the
\texttt{GenRunInfoProduct} branch of each MINIAODSIM ROOT file.
\begin{table*}[!tbp]
\centering
\caption{Event and object selections applied during feature extraction.}
\label{tab:extraction-cuts}

\begin{tabular}{p{0.18\textwidth} p{0.39\textwidth} p{0.39\textwidth}}
\toprule
Stage & DoubleMuon / DoubleEG data & MonoZToLL MC \\
\midrule
Trigger & Dataset-level HLT audit; retained events pass the extractor mask &
Dataset-level trigger audit; retained events pass the extractor mask \\
Lepton multiplicity & Leading same-flavor pair & Best valid same-flavor pair \\
Lepton $p_{\mathrm{T}}$ & $p_{\mathrm{T}}(\ell_1)>25\,\GeV$, $p_{\mathrm{T}}(\ell_2)>20\,\GeV$ &
Same \\
Charge & Opposite sign & Opposite sign \\
Muon acceptance & $|\eta_{\mu}|<2.4$ & $|\eta_{\mu}|<2.4$ \\
Electron acceptance & $|\eta_e|<2.4$, ECAL transition region excluded ($1.4442<|\eta_e|<1.566$) & $|\eta_e|<2.4$, ECAL transition region excluded \\
Isolation & Relative isolation $<0.15$ for both leptons & Same \\
Electron ID & Spring15 non-triggering MVA working point & Not applied \\
Impact parameter & Audited as \texttt{lep1\_dxy\_sig} feature; not applied as an event selection cut in either data or MC & Audited as \texttt{lep1\_dxy\_sig} feature; not applied as an event selection cut in either data or MC \\
Dilepton mass & $60<m_{\ell\ell}<120\,\GeV$ & Same \\
Jets & $p_{\mathrm{T}}>30\,\GeV$, $|\eta|<2.4$, lepton overlap removed & Same \\
$b$ veto & Medium CSVv2 veto, $n_b=0$, audited in output & \texttt{n\_bjets} retained for schema parity \\
\bottomrule
\end{tabular}

\end{table*}

\section{Feature Extraction and Cleaning}
\label{sec:preprocessing}

Raw MINIAOD and MINIAODSIM files are read directly with \texttt{uproot}, with jagged
collections handled in \texttt{awkward} and converted to fixed per-event NumPy arrays
after object selection. The analysis products are ROOT \texttt{Events} TTrees written with
\texttt{uproot.recreate}, \texttt{mktree}, and \texttt{extend}; CSV files are used only for
cut-flow, file-audit, and EDA summary outputs. Per-event quantities include \MET,
dilepton and $Z$ kinematics, jet and $b$-tag summaries, hadronic recoil variables, and
angular correlations (40 physics features in the extracted schema). The complete list of
stored branches and formulas for derived observables is given in
Appendix~\ref{app:features}; Table~\ref{tab:extraction-cuts} lists the selection cuts
implemented in the extraction notebooks.

The DoubleMuon and DoubleEG extractors use a custom trigger-object decoder because the
2015 MiniAOD trigger information is not exposed as simple one-branch-per-HLT-path
booleans. The decoder reads the \texttt{selectedPatTrigger} path-name,
\texttt{pathLastFilterAccepted}, and filter-label collections. When path names are
available, events are matched to the channel-specific HLT prefixes; when path names are
empty, the extractor falls back to accepted final-filter labels such as the double-muon
or double-electron filter tokens. Trigger decisions are accumulated in
\texttt{TRIGGER\_AUDIT} and the retained event mask is applied before physics-feature
writing. For the data $b$ veto, the notebooks also decode the MiniAOD
\texttt{pat::Jet::pairDiscriVector\_} payload using \texttt{uproot}'s \texttt{AsBinary}
interpretation. A slow decoder discovers the discriminator labels, a cached fixed-layout
fast path extracts the CSVv2 scores, and a validation step checks the fast decoder against
the slow decoder on sampled events. The medium 2015/76X CSVv2 working point
(\(0.800\)) is then used to compute \texttt{n\_bjets}, \texttt{max\_btag\_csvv2}, and
the $b$-veto audit branches.
\begin{figure*}[htbp]
\centering

\includegraphics[width=0.48\textwidth]{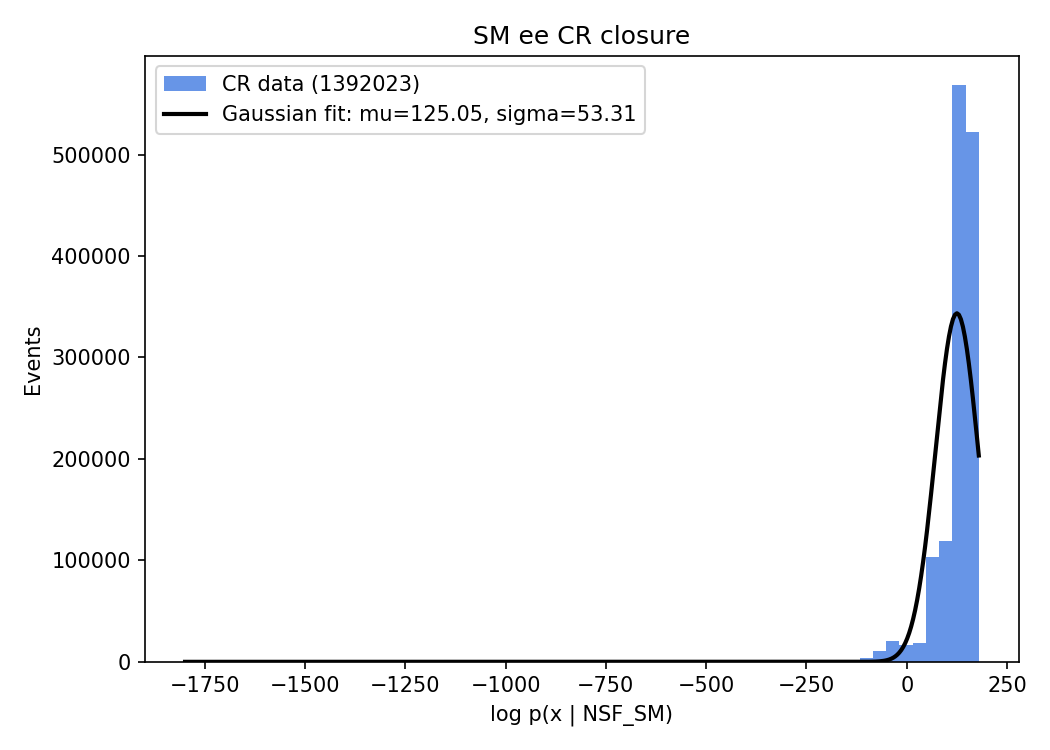}
\hfill
\includegraphics[width=0.48\textwidth]{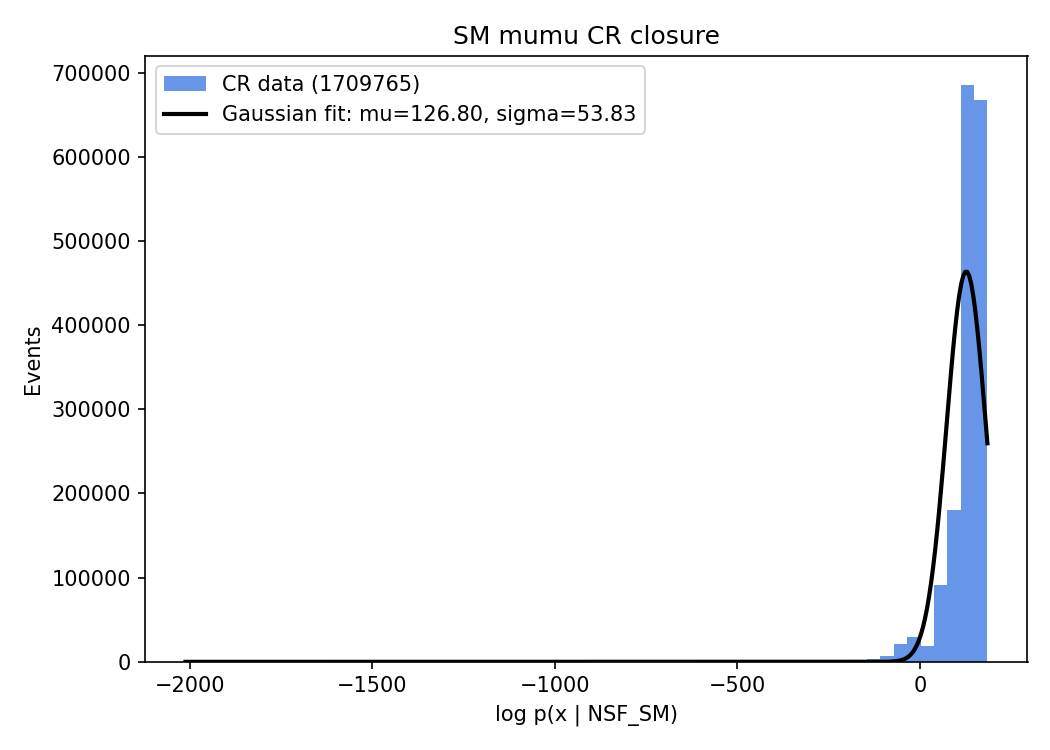}

\caption{
Control-region distributions of $\log p(\mathbf{x}\mid\mathrm{NSF\_SM})$ for
the electron (left) and muon (right) channels. Overlaid Gaussian fits
summarise the location and width of the dominant peak.
}
\label{fig:cr_closure}
\end{figure*}
Cleaning is applied only to SM data TTrees. The SM cleaner clips fixed-domain variables,
estimates high-quantile caps for heavy-tailed variables, and uses the opposite SM lepton
channel as the reference sample when deriving channel-consistent caps. The DoubleMuon
cleaner therefore references DoubleEG parts, and the DoubleEG cleaner references
DoubleMuon parts. The merged SM bounds are exported to
\texttt{Output/sm\_feature\_bounds.json}; DM MC ROOT files are not cleaned by
\texttt{cleaner.py}. Instead, the exported SM bounds are applied later at DataLoader time,
so genuine high-\MET\ or high-$p_{\mathrm{T}}$ DM tails are preserved in the extracted MC
files until the training pipeline applies the common SM feature domain. The fixed bounds
and tail-rule families used by the cleaning script are summarized in
Appendix~\ref{app:cleaning}.

The cleaned NSF feature manifest retains 35 extracted physics features and excludes
\texttt{lepton\_flavor}, \texttt{n\_bjets}, \texttt{gen\_weight}, \texttt{rho},
\texttt{n\_vertices}, and raw \texttt{HT}; \texttt{lepton\_flavor} is retained only for
channel routing and is not passed to any flow.
Two binary jet-presence indicators computed at DataLoader time
(\texttt{has\_jet1}$=\mathbb{1}[n_\mathrm{jets}\ge 1]$,
\texttt{has\_jet2}$=\mathbb{1}[n_\mathrm{jets}\ge 2]$) are appended to form the final
37-dimensional NSF input.

\section{Control and Signal Regions}
\label{sec:regions}

Region masks are built from \MET, $|\Delta\phi(\MET,Z)|$, and $n_{\mathrm{jets}}$ after
cleaning:
\begin{itemize}
  \item CR: $\MET < 50\,\GeV$.
  \item Validation: $50 \le \MET < 100\,\GeV$, with the same
        $|\Delta\phi(\MET,Z)|$ and jet cuts as the SR.
  \item SR: $\MET \ge 50\,\GeV$, $|\Delta\phi(\MET,Z)| > 2.5$, and
        $n_{\mathrm{jets}}\le 1$.
\end{itemize}
The CR ($\MET<50\,\GeV$) and SR ($\MET\ge 50\,\GeV$) are disjoint by construction at the
$\MET=50\,\GeV$ boundary.
The validation interval lies inside the SR \MET\ range and is reserved for SM closure
checks before interpreting the full SR scores in data.
The SR imposes no explicit upper \MET\ requirement; in practice the
cleaned dataset caps \MET\ at approximately 200\,\GeV\
(Section~\ref{sec:datasets}).

\section{Neural Spline Flow Models}
\label{sec:nsf}

Each NSF maps a feature vector $\mathbf{x}\in\mathbb{R}^d$ to a base density through a
composition of coupling layers with rational-quadratic spline transforms~\cite{Durkan2019}.
We implement flows in PyTorch using zuko , with hyperparameters
(layer count, hidden width, bin count) fixed before training and recorded in run manifests.

Training protocol:
\begin{itemize}
  \item \texttt{NSF\_SM\_mumu}: CR double-muon data, 70\% train / 30\% validation.
  \item \texttt{NSF\_SM\_ee}: CR double-electron data, 70\% train / 30\% validation.
  \item \texttt{NSF\_DM\_vector}: full vector-mediator MC, 70\% train / 30\% validation,
        all flavors.
  \item \texttt{NSF\_DM\_axial}: full axial-vector MC, 70\% train / 30\% validation,
        all flavors.
  \item \texttt{NSF\_DM\_scalar}: full scalar-mediator MC, 70\% train / 30\% validation,
        all flavors.
\end{itemize}
\begin{figure*}[htbp]
\centering

\includegraphics[width=0.48\textwidth]{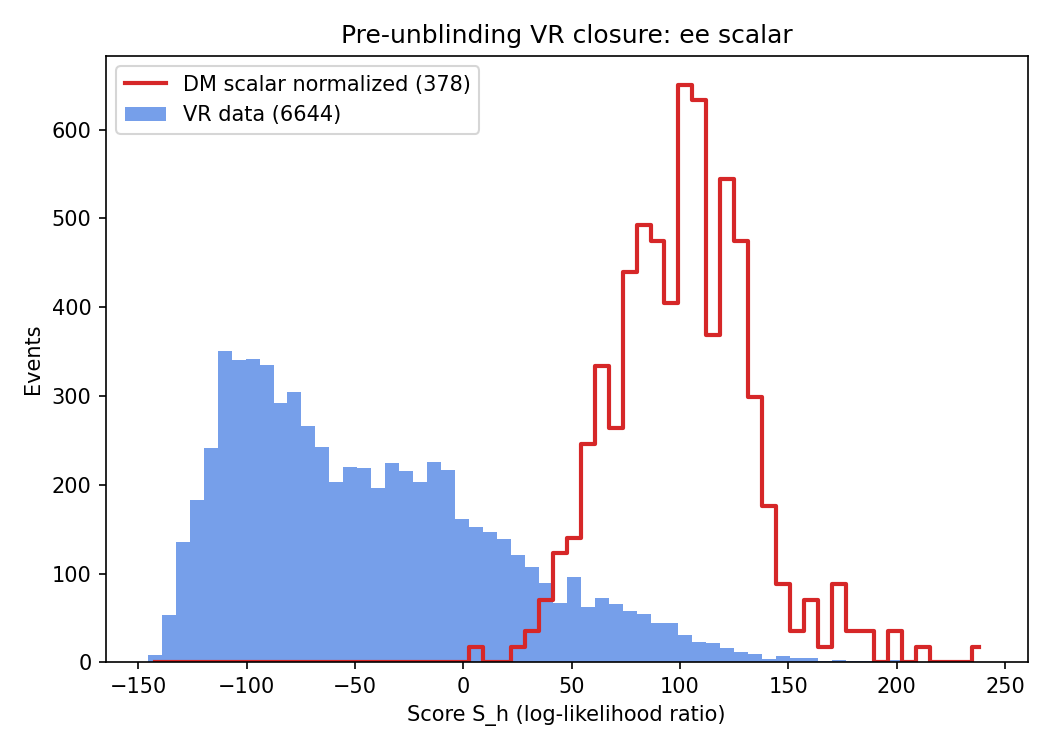}
\includegraphics[width=0.48\textwidth]{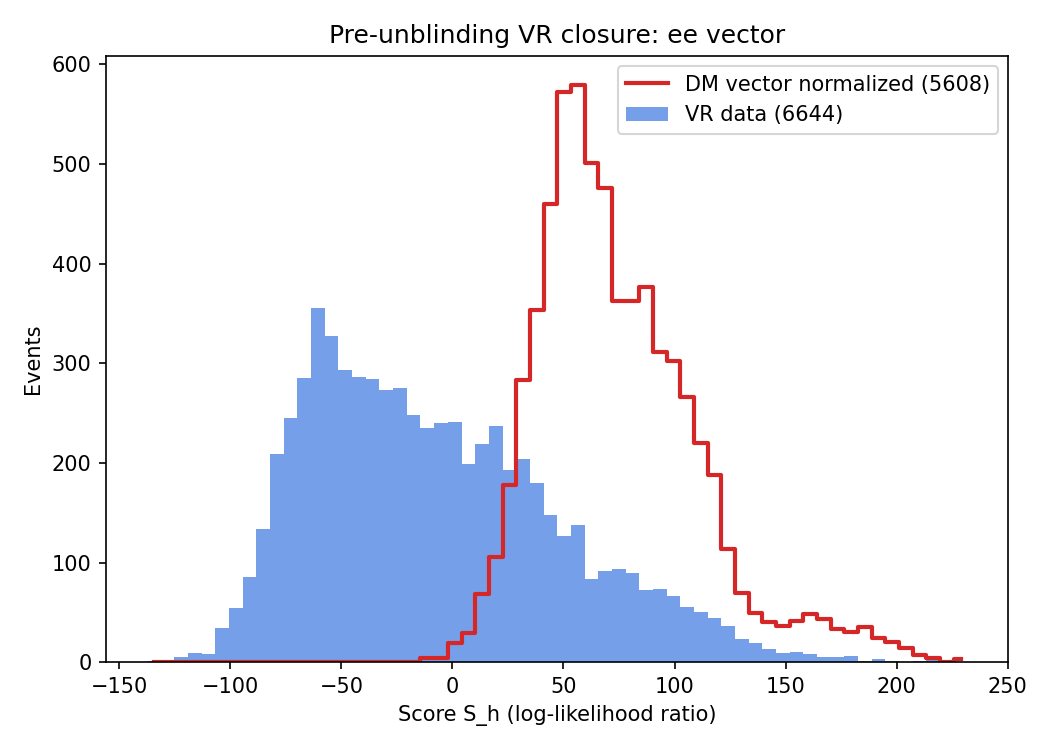}
\includegraphics[width=0.48\textwidth]{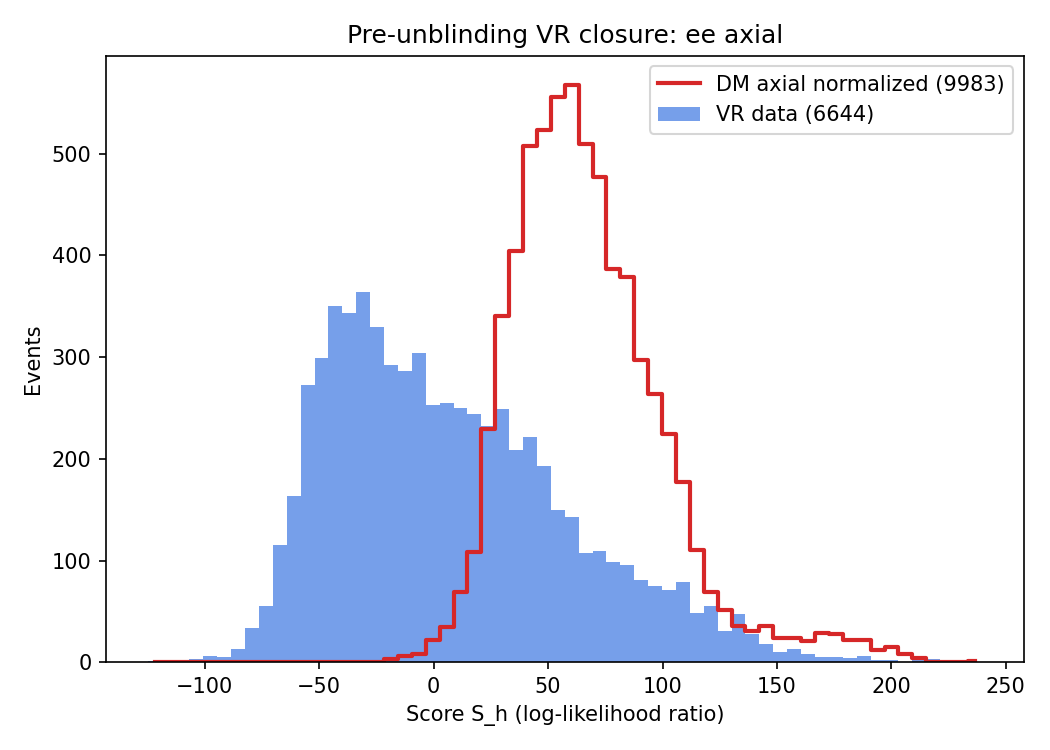}
\includegraphics[width=0.48\textwidth]{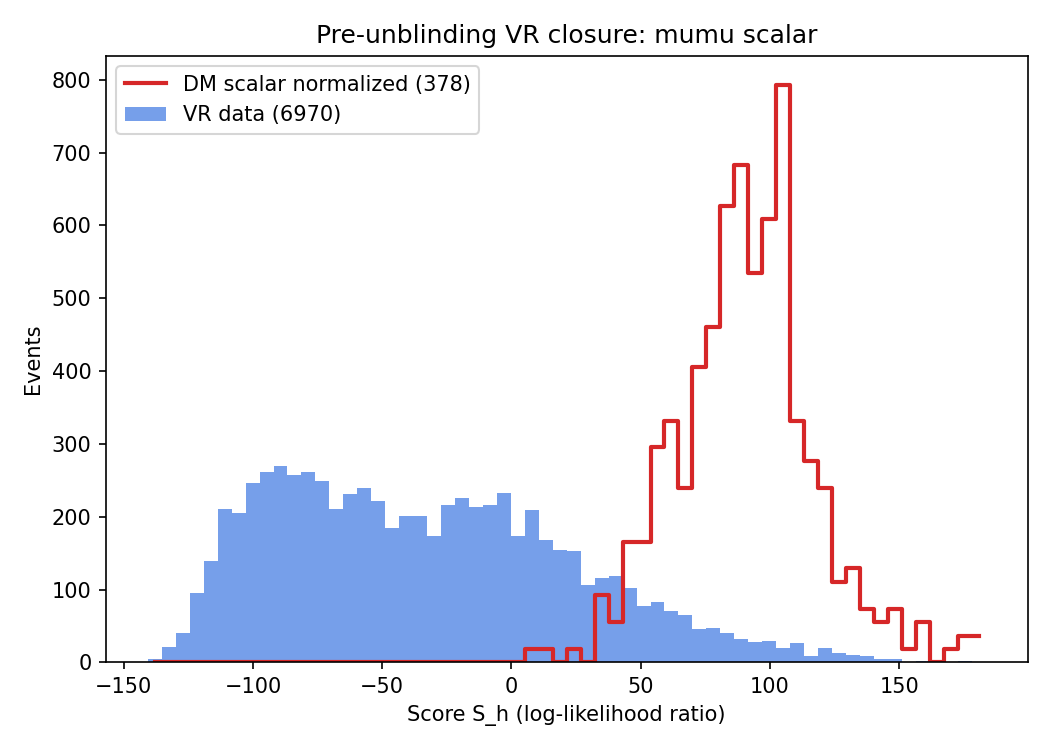}
\includegraphics[width=0.48\textwidth]{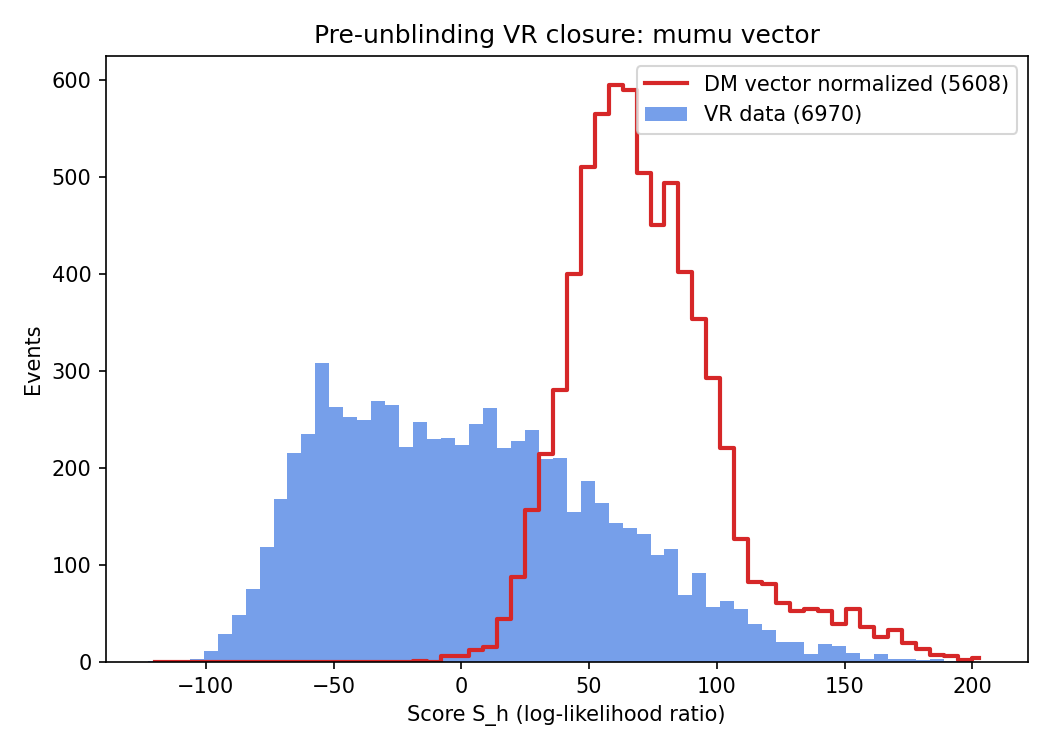}
\includegraphics[width=0.48\textwidth]{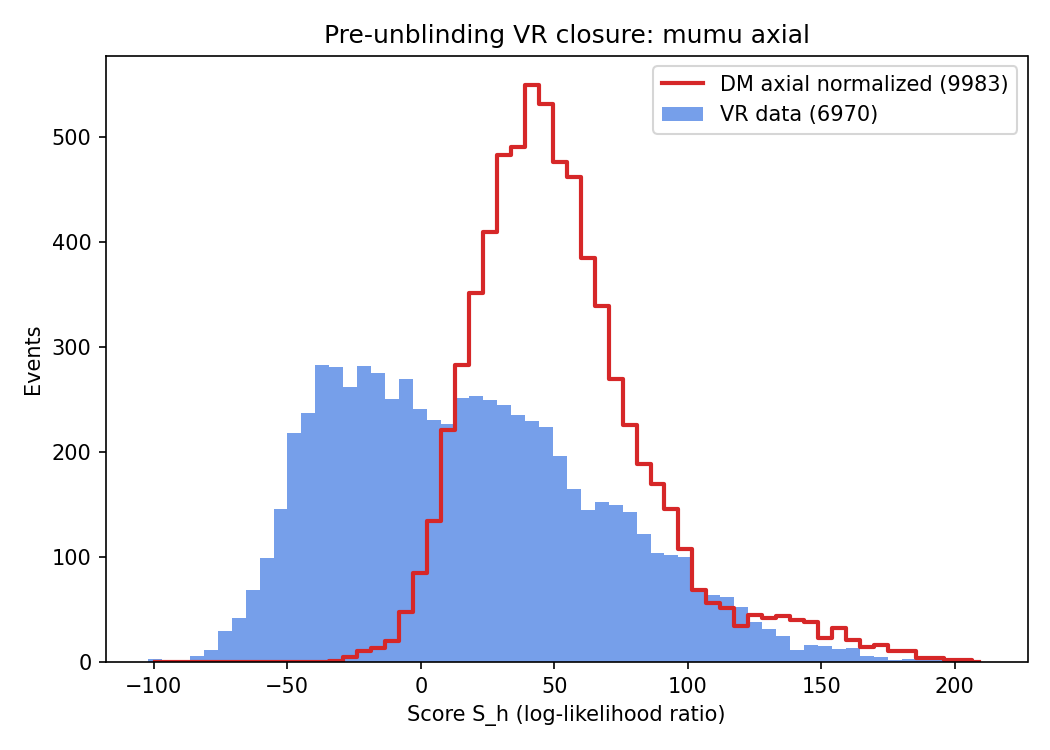}

\caption{
Pre-unblinding validation-region score distributions comparing observed data
with normalised dark-matter Monte Carlo for the electron (top row) and muon
(bottom row) channels. These plots test signal sensitivity in the
$50\le\MET<100\,\GeV$ band rather than background-modelling closure.
}
\label{fig:vr_validation}
\end{figure*}
\begin{figure*}[htbp]
\centering

\includegraphics[width=0.48\textwidth,height=0.23\textheight]{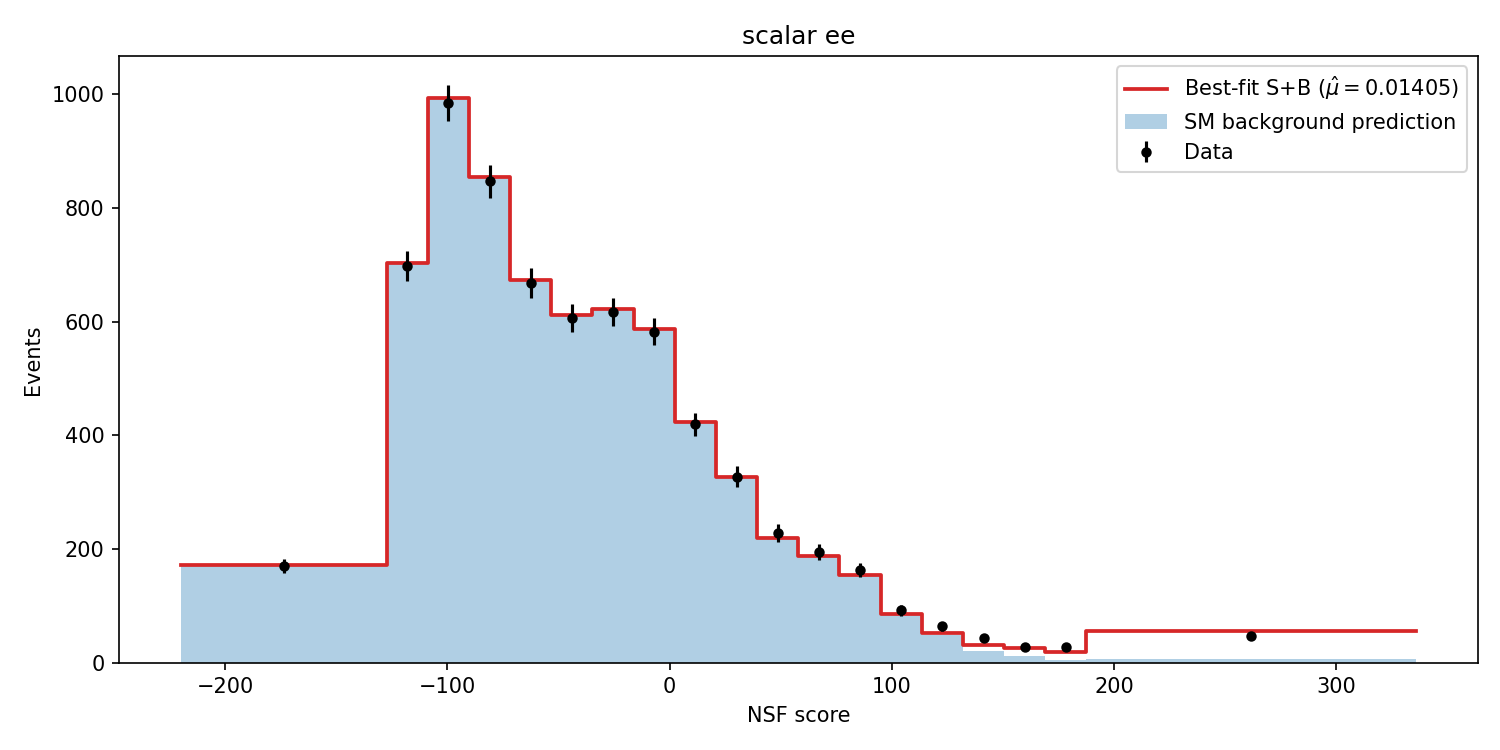}
\includegraphics[width=0.48\textwidth,height=0.23\textheight]{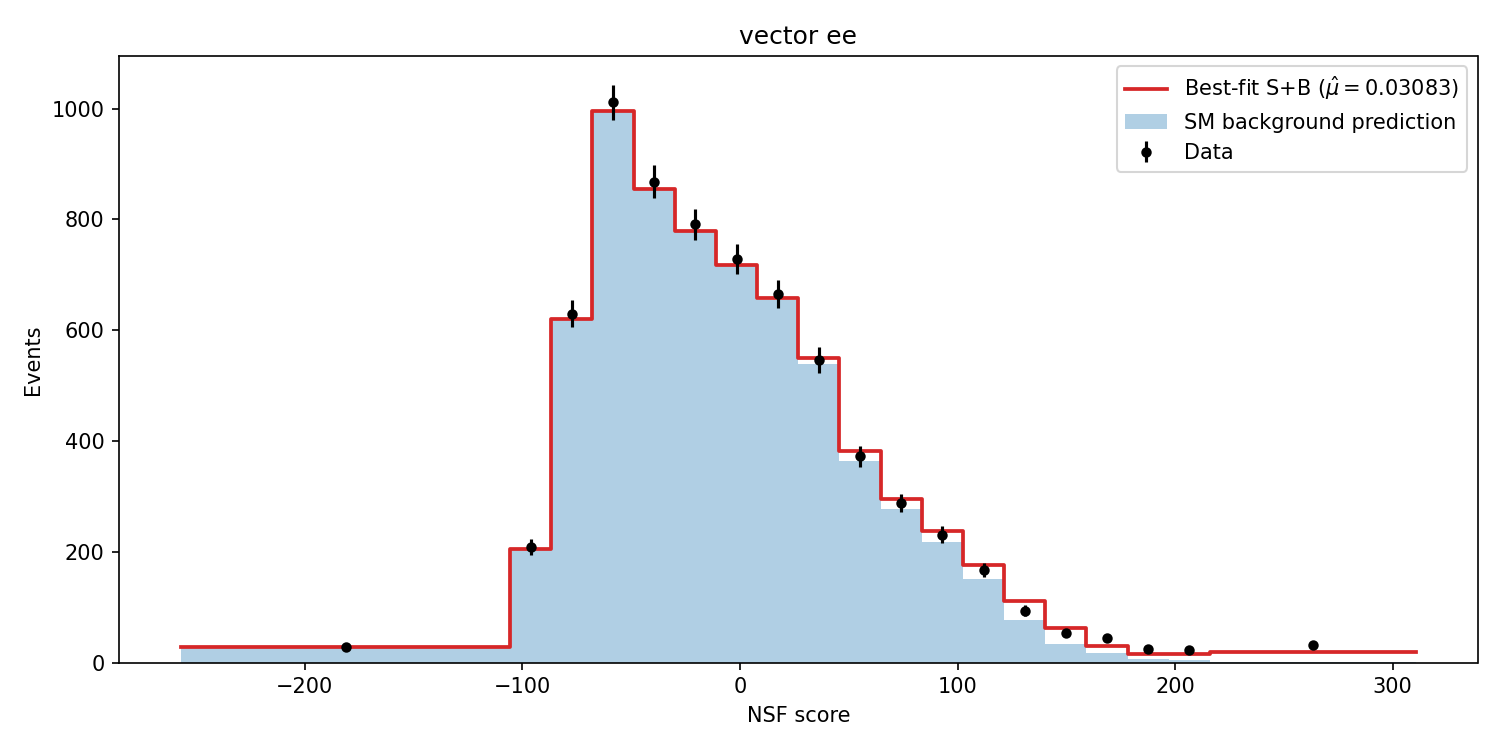}
\includegraphics[width=0.48\textwidth,height=0.23\textheight]{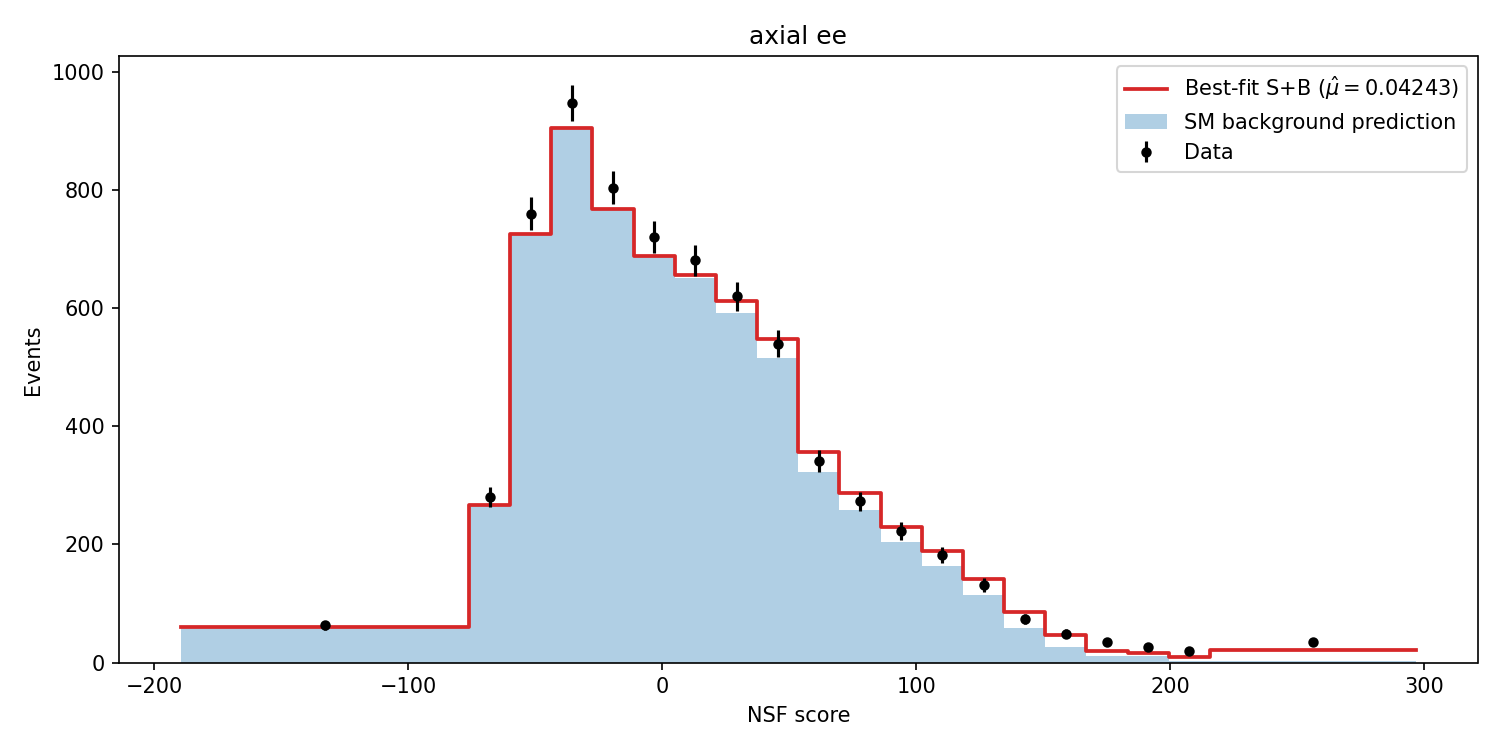}
\includegraphics[width=0.48\textwidth,height=0.23\textheight]{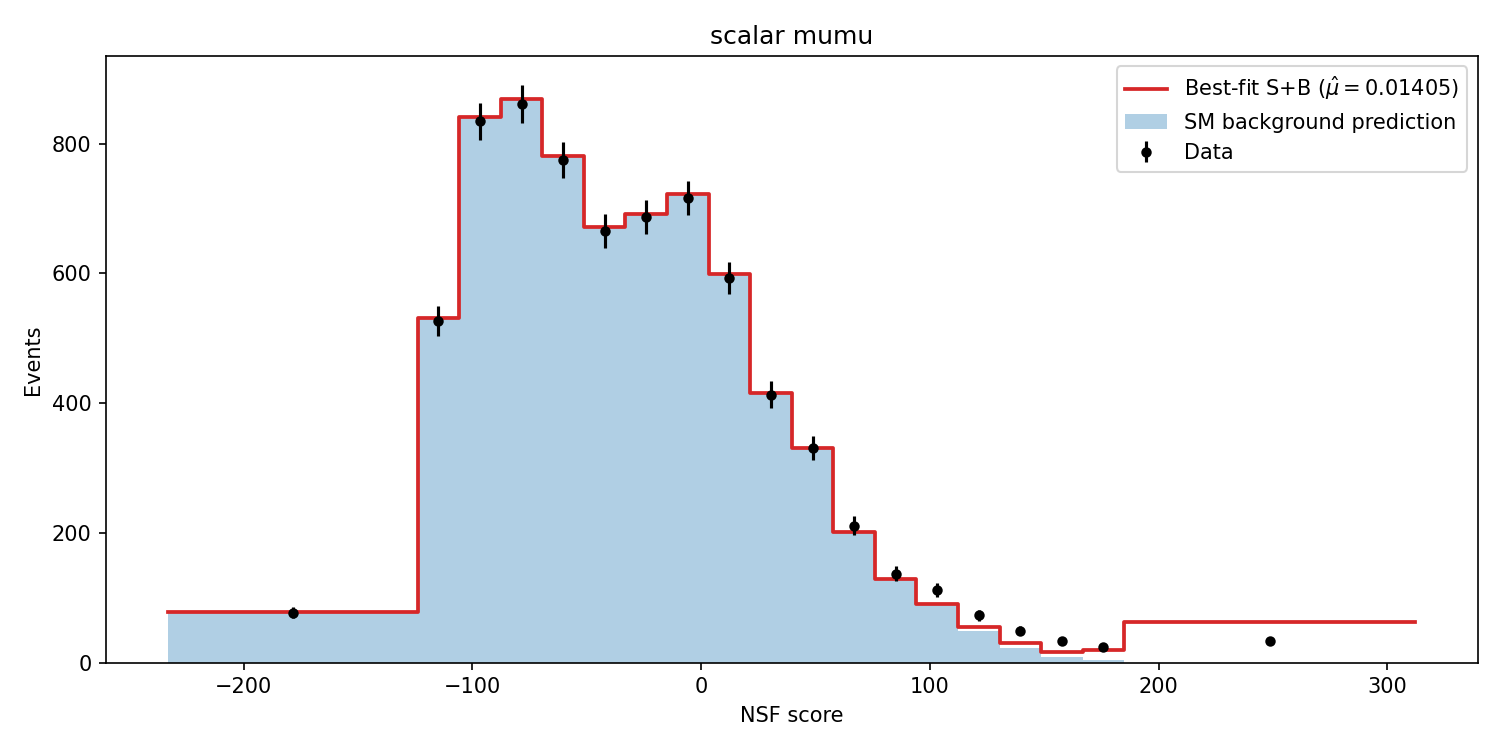}
\includegraphics[width=0.48\textwidth,height=0.23\textheight]{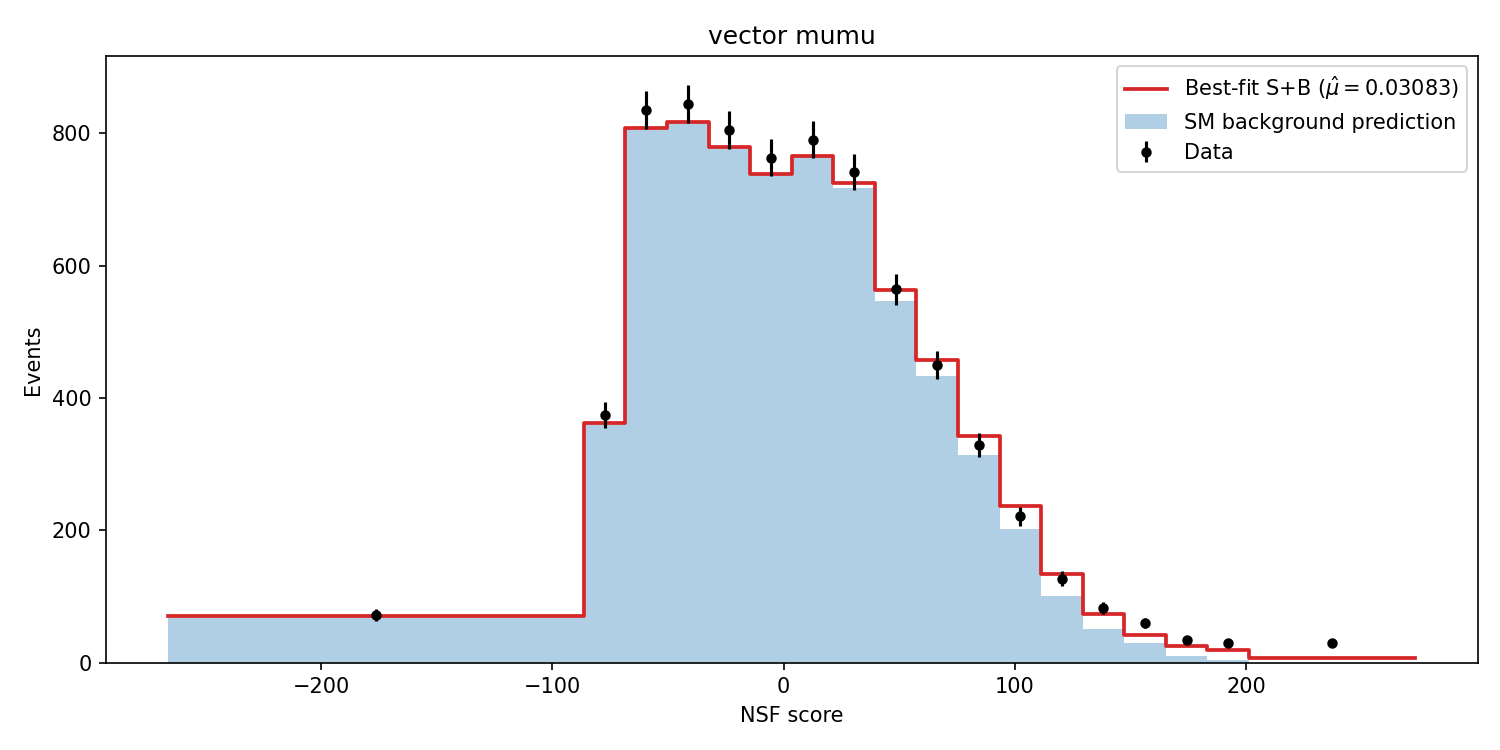}
\includegraphics[width=0.48\textwidth,height=0.23\textheight]{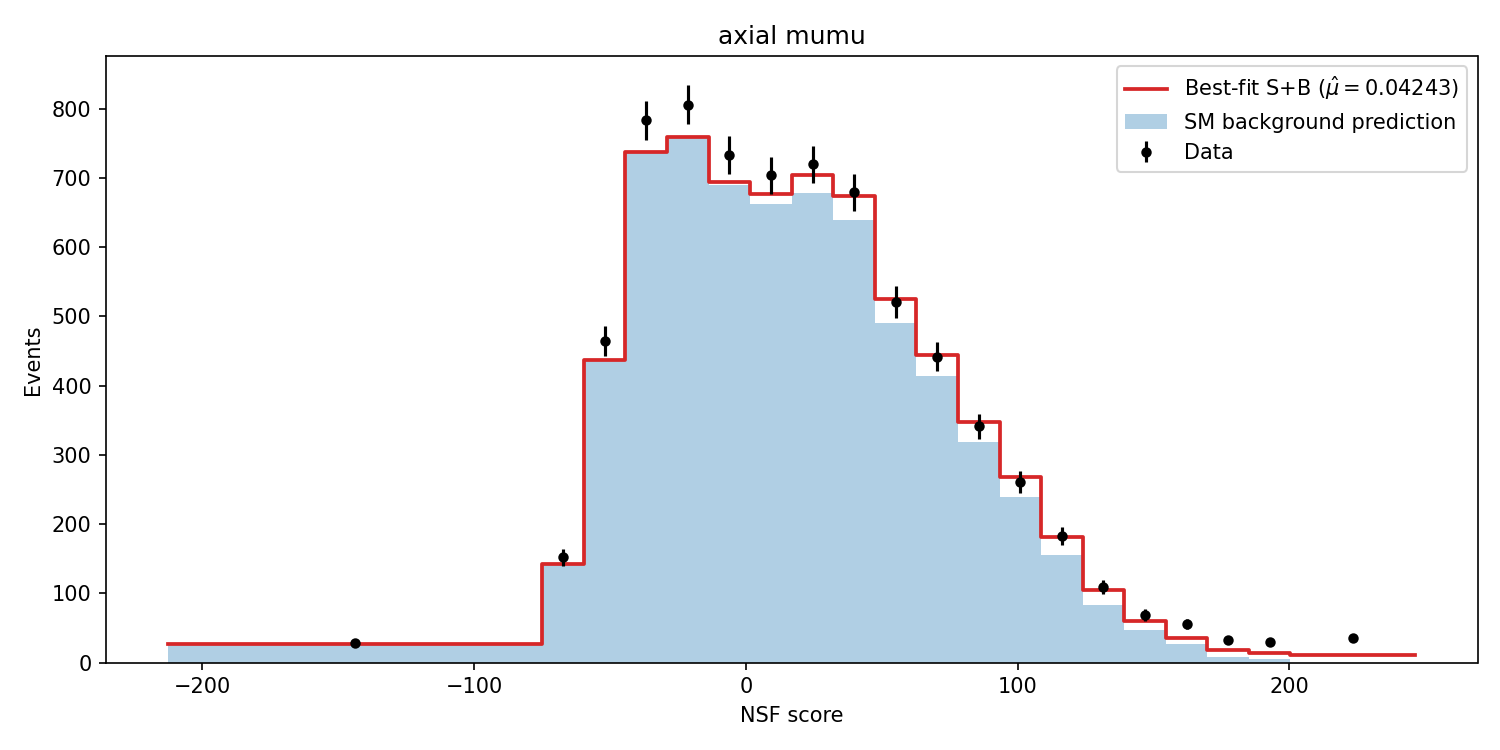}

\caption{
Post-fit SR score distributions in the electron (top row) and muon (bottom
row) channels for each mediator hypothesis. Black points are observed SR
data; blue histograms are best-fit SM background prediction obtained from the simultaneous SR+VR profile-likelihood fit; red curves
show the best-fit $S+B$ model.
}
\label{fig:fit_results}
\end{figure*}

\begin{table*}[t]
\centering
\caption{Summary of the final simultaneous SR+VR profile-likelihood results for the three dark matter mediator hypotheses. The VR constrains the background normalisation independently; the SR is used for signal search. The 68\% and 95\% expected bands are quoted for the expected 95\%~CL upper limit on $\mu$. Cross-section limits use the generator cross-sections; the two axial-vector values correspond to the two benchmark points.}
\label{tab:fit_summary}
\resizebox{\textwidth}{!}{%
\begin{tabular}{lrrrrrrrr}
\toprule
Mediator & $\hat{\mu}$ & $\sigma_{\hat{\mu}}$ & $p_0$ & $Z$ &
$\mu^{95}_{\mathrm{obs}}$ & $\mu^{95}_{\mathrm{exp}}$ &
68\% exp. band / 95\% exp. band & $\sigma^{95}_{\mathrm{obs}}$ [pb] \\
\midrule
Scalar & $1.41\times10^{-2}$ & 0.00136 & $\simeq0$ & $8.0^\dagger$ &
0.0177 & 0.0018 & [0.00156, 0.00163] / [0.00154, 0.00166] &
$1.76\times10^{-9}$ \\
Vector & $3.08\times10^{-2}$ & 0.00245 & $\simeq0$ & $8.0^\dagger$ &
0.0362 & 0.0039 & [0.00326, 0.00364] / [0.00309, 0.00382] &
$9.90\times10^{-4}$ \\
Axial-vector & $4.24\times10^{-2}$ & 0.00357 & $\simeq0$ & $8.0^\dagger$ &
0.0498 & 0.0069 & [0.00560, 0.00659] / [0.00517, 0.00715] &
$9.24\times10^{-2}$ / $7.89\times10^{-3}$ \\
\bottomrule
\end{tabular}
}
\end{table*}
\noindent\(\dagger\) The significance is capped at 8.0 due to floating-point
underflow of the chi-square survival function at large \(q_0\). The underlying
\(q_0\) values (233--327) indicate a background-modelling residual
(Section~\ref{sec:met-tail}), not a genuine \(8\sigma\) discovery.
Early stopping uses validation negative log-likelihood.

The five NSFs share the following hyperparameters: 8 masked autoregressive coupling
transforms with rational-quadratic splines (8 bins per spline, randomised feature
permutation per transform), conditioner networks with two hidden layers of 256 units each,
and a standard-normal base distribution.
Training uses the Adam optimiser ($\eta=3\times10^{-4}$, $\lambda=10^{-5}$ weight decay)
with a 500-step linear warm-up followed by cosine annealing to
$\eta_\mathrm{min}=10^{-6}$ over 200 epochs, a mini-batch size of 2048, and gradient
clipping at $\|\nabla\|_2\le 1$.
Early stopping monitors validation NLL with a patience of 20 epochs and restores the
best-epoch weights.
In the present training runs, validation NLL continued to decrease at
epoch~200 for all five flows; early stopping was not triggered for any
flow before reaching \texttt{MAX\_EPOCHS}. Every \texttt{\_best.pt}
checkpoint therefore corresponds to the final epoch rather than a
converged minimum, and we flag potential residual undertraining as a
limitation affecting the learned densities and reported limits.
The discrete feature \texttt{n\_jets} is dequantized during training by adding uniform
noise $\mathcal{U}(-0.5,0.5)$; raw integer values are used at inference.

All five flows operate in the same standardized feature space.
SM flows fit a per-feature $z$-score scaler on their respective training splits and save
the parameters to a JSON manifest.
DM flows reuse the \texttt{NSF\_SM\_mumu} scaler rather than fitting new scale parameters,
ensuring that DM event densities are evaluated in the SM feature domain.
Before standardization, DM MC features are clipped to the SM bounds exported in
\texttt{Output/sm\_feature\_bounds.json}; this preserves genuine high-\MET\ tails while
preventing out-of-domain extrapolation of the learned SM density.

The total training campaign therefore consists of five independent NSF runs: two
channel-specific SM flows and three mediator-specific DM flows.
Closure checks use the 50--100~\GeV\ validation band to verify the channel-specific SM
models before SR score arrays are interpreted.
The axial-vector $\mu\mu$ score distribution shows a localised bin-wobble
pattern near $\mathcal{S}_h\approx0$--50, consistent with residual NSF
spline instability from non-convergence; this feature appears in all three
fit variants and is therefore a property of the trained flow rather than an
artifact of any single background model.

\section{Likelihood-Ratio Search}
\label{sec:search}

The search uses $\score_h$ as the per-event test statistic for each mediator hypothesis
(Eq.~\eqref{eq:score}).
For $\mu\mu$ SR events, vector, axial-vector, and scalar DM log densities are compared
with \texttt{NSF\_SM\_mumu}; for $ee$ SR events the same three DM log densities are
compared with \texttt{NSF\_SM\_ee}.
This yields six score arrays in total.
We perform three simultaneous SR+VR binned profile-likelihood fits for signal
strength $\mu$, one for each DM hypothesis. Each fit combines the $\mu\mu$ and
$ee$ score histograms from both the SR and VR simultaneously:
\begin{equation}
  \ln\mathcal{L} = \ln\mathcal{L}_{\mathrm{SR}} + \ln\mathcal{L}_{\mathrm{VR}}
  - \tfrac{1}{2}\theta_{\mu\mu}^2 - \tfrac{1}{2}\theta_{ee}^2,
\end{equation}
where $\mathcal{L}_{\mathrm{SR}}$ and $\mathcal{L}_{\mathrm{VR}}$ are binned
Poisson likelihoods over the score histograms in each region, the signal
strength $\mu$ and the per-channel normalisation nuisances
$\theta_{\mu\mu}$, $\theta_{ee}$ are shared between regions, and the
background templates in both regions are derived from a single-step
VR$\to$SR shape transfer: the SM VR score histogram is renormalised to
the respective region yield and used as the nominal background
prediction for that region. The observed data histograms are the raw SR
and VR score counts, which differ from the renormalised VR background
template by real Poisson fluctuations and by genuine shape differences
between the VR and SR score distributions.
The VR component constrains the background normalisation using
$\sim$6,970 ($\mu\mu$) and $\sim$6,644 ($ee$) VR events independently
of the SR, providing a genuine sideband-based background prediction.
Signal MC contributes to both SR and VR components with the same $\mu$.
Asymptotic formulae~\cite{Cowan2011} are used for the discovery test statistic and CL$_s$ upper limits.
\begin{figure*}[t]
\centering

\includegraphics[width=0.48\textwidth]{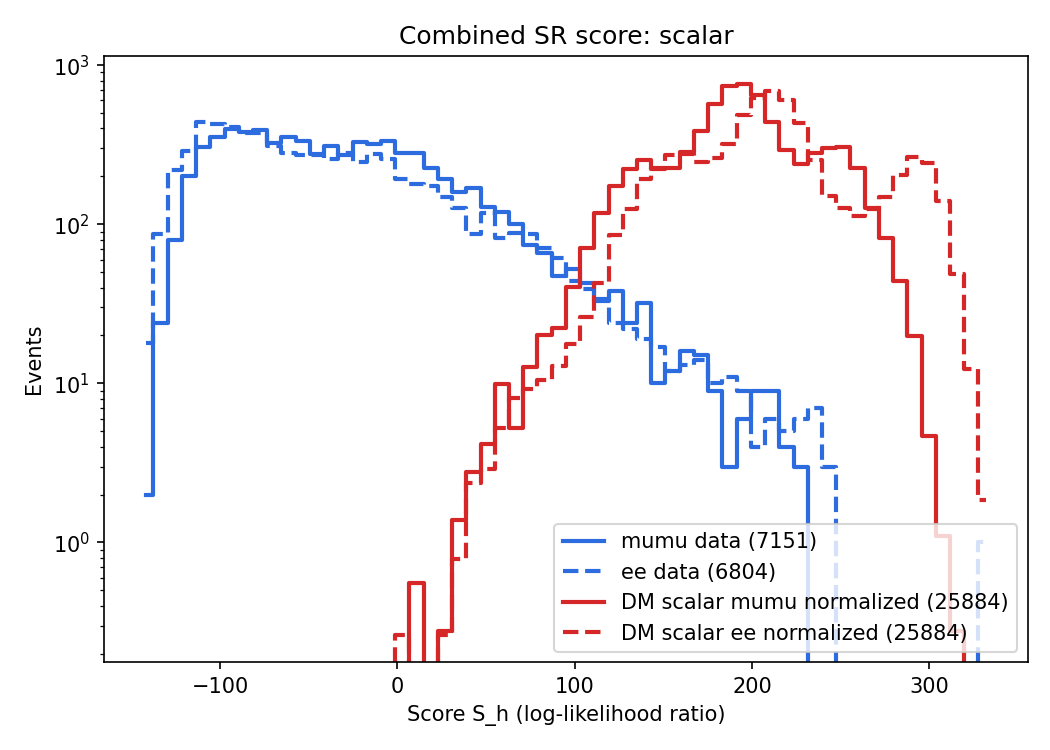}
\includegraphics[width=0.48\textwidth]{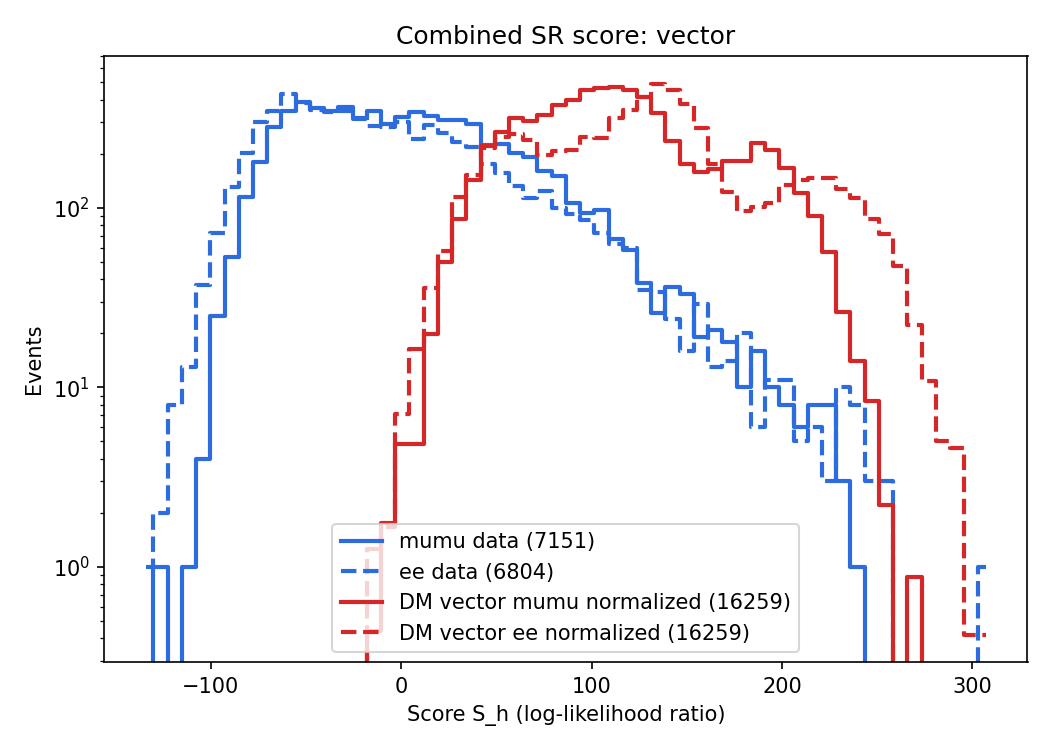}
\includegraphics[width=0.48\textwidth]{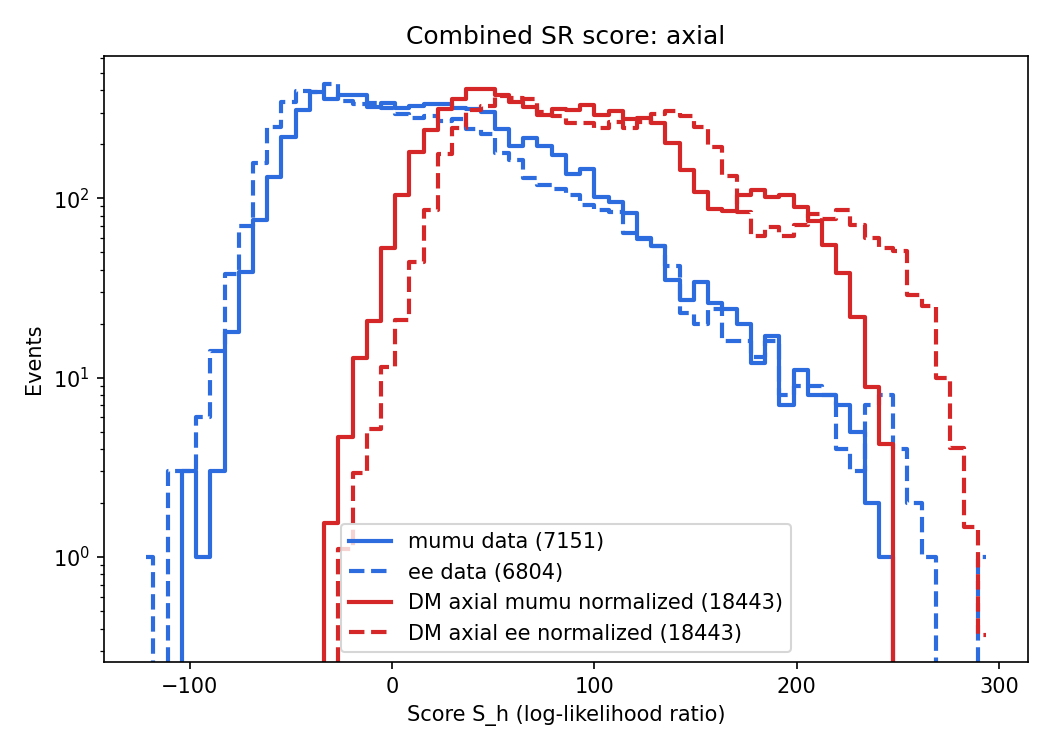}

\caption{
Signal-region NSF score distributions for scalar, vector, and axial-vector mediator hypotheses. Data from the electron and muon channels are shown together with the corresponding dark matter signal templates.
}
\label{fig:combined_sr_scores}
\end{figure*}
\subsection{Systematic uncertainties}
\label{sec:systematics}
The profile-likelihood fit includes independent Gaussian-constrained
normalisation nuisance parameters for the $\mu\mu$ and $ee$ channels
with a prior width of 5\% ($\sigma=0.05$).
In the simultaneous SR+VR fit the normalisation nuisances are
constrained additionally by the VR data; the posterior uncertainty on
$\theta_{\mathrm{norm}}$ is approximately 0.17 (compared to the prior
of 1.0 in sigma units), corresponding to an effective normalisation
constraint of $\sim$0.85\% from the VR statistics alone.

\paragraph{Integrated luminosity (propagated).}
The Run~2015D dataset corresponds to
$\mathcal{L}=2.32\pm0.037\,\mathrm{fb}^{-1}$ (uncertainty 1.6\%),
taken from the CMS precision luminosity measurement~\cite{CMS-LUM-2021}.
A correlated log-normal luminosity nuisance is applied across both channels.

\paragraph{Lepton efficiency (propagated).}
Combined identification, reconstruction, and trigger efficiency
uncertainties of approximately 3\% per channel are applied as
independent per-channel log-normal nuisances, following the treatment
in the CMS Run~2 mono-$Z$ search at the same centre-of-mass
energy~\cite{CMS2017MonoZ}. Lepton momentum-scale uncertainties of 1\%
($\mu$) and 2\%/5\% (electrons, barrel/endcap) are included as additional
shape nuisances on the lepton $p_{\mathrm{T}}$ distributions.

\paragraph{MET resolution and pileup (not propagated).}
Evaluating \MET\ resolution and pileup-reweighting uncertainties
requires re-scoring SR, VR, and CR events with shifted inputs, which
constitutes new analysis work not performed in this study. These
sources are acknowledged as a gap in the systematic budget.

\paragraph{Combined systematic budget.}
Treating luminosity as fully correlated across channels and lepton
efficiency as per-channel independent, the combined propagated external
systematic is approximately $\sqrt{1.6^2+3^2+3^2}\approx4.5\%$ per
channel, on top of the 0.85\% data-driven normalisation constraint.
The dominant uncertainty is the unresolved high-\MET\ tail modelling
residual (Section~\ref{sec:met-tail}).

\subsection{High-\MET\ tail modelling residual}
\label{sec:met-tail}

Events with $\MET\ge100\,\GeV$ constitute 2.4--2.5\% of the SR
($\sim$160--181 events across both channels) but carry mean NSF scores
140--195 units higher than the VR bulk --- a shift exceeding the score
distribution's own standard deviation of 40--60 units.
The single-step VR background has negligible support at these scores,
so the tail population appears as a systematic upward fluctuation in
the fit.

To characterise this effect, we tested linear and quadratic
\MET-dependent models of the CR/VR score trend, validating each against
held-out VR bins before extrapolating to the SR tail
(Section~\ref{sec:extrapolation-validation}).
Neither functional form reliably predicts the observed tail behaviour:
the linear model underpredicts the true SR-tail mean score by 90--115
units, while the quadratic model overpredicts by 300--330 units. The
limited tail statistics (160--181 events across 100\,\GeV) do not
support more flexible extrapolation.

The VR-shape background is therefore used as nominal, and the resulting
inflation of $\hat\mu$, $q_0$, and the observed-to-expected limit ratios
reflects this unresolved modelling residual rather than evidence of DM
production.

\subsection{NSF CR$\to$SR extrapolation validation}
\label{sec:extrapolation-validation}

To validate that the NSF SM density, trained only on the control region
($\MET<50\,\GeV$), extrapolates in a characterised way into the signal
region, we measured the mean NSF score as a function of \MET\ across CR
and VR bins for all three mediator hypotheses and both lepton channels
(six combinations), then tested two extrapolation models against
held-out data.

\paragraph{Procedure.}
Events in the CR and VR were binned by \MET\ in 10\,GeV intervals.
For each bin the mean NSF score $\langle\mathcal{S}_h\rangle$ was
computed. The two highest-\MET\ VR bins (80--100\,GeV) were held out
as a validation set. Linear and quadratic functions of \MET\ were fit
to the remaining bins and extrapolated to the SR tail
($\MET\ge100\,\GeV$), where the true mean score was measured
independently.

\paragraph{Results.}
The linear fit undershot the held-out VR bins by 0.3--0.6$\sigma$
(systematic across all six combinations). When extrapolated to
$\MET\sim150\,\GeV$ it overshot the true SR-tail mean score by 90--115
units. The quadratic fit overshot the held-out bins by 1.1--3.3$\sigma$
(in the opposite direction from the linear case) and overshot the true
SR-tail mean by 300--330 units; the positive curvature in all six fits
indicates the quadratic form was dominated by the steep low-\MET\ rise
rather than the true behaviour at higher \MET.

\paragraph{Conclusion.}
Neither functional form reliably extrapolates the CR-trained NSF
density into the high-\MET\ SR tail. Combined with the limited tail
statistics (160--181 events across 100\,\GeV), this demonstrates that
the NSF's CR-trained density does not extrapolate smoothly or
predictably into the sparse high-\MET\ signal region. This validates
and motivates the treatment of the resulting background modelling
residual as an explicit, quantified limitation
(Section~\ref{sec:met-tail}) rather than an uncorrected prediction.

\section{Results}
\label{sec:results}
\begin{figure*}[t]
\centering

\includegraphics[width=0.48\textwidth]{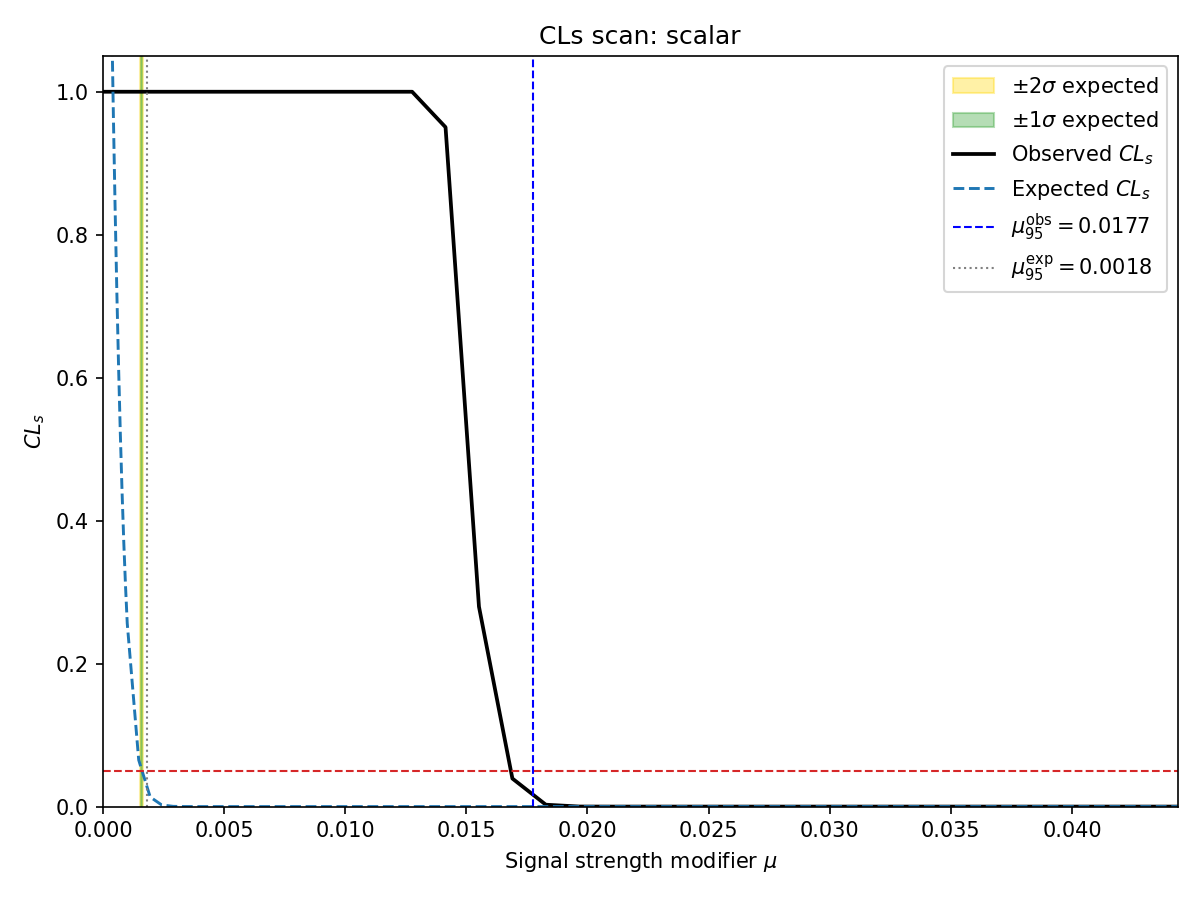}
\includegraphics[width=0.48\textwidth]{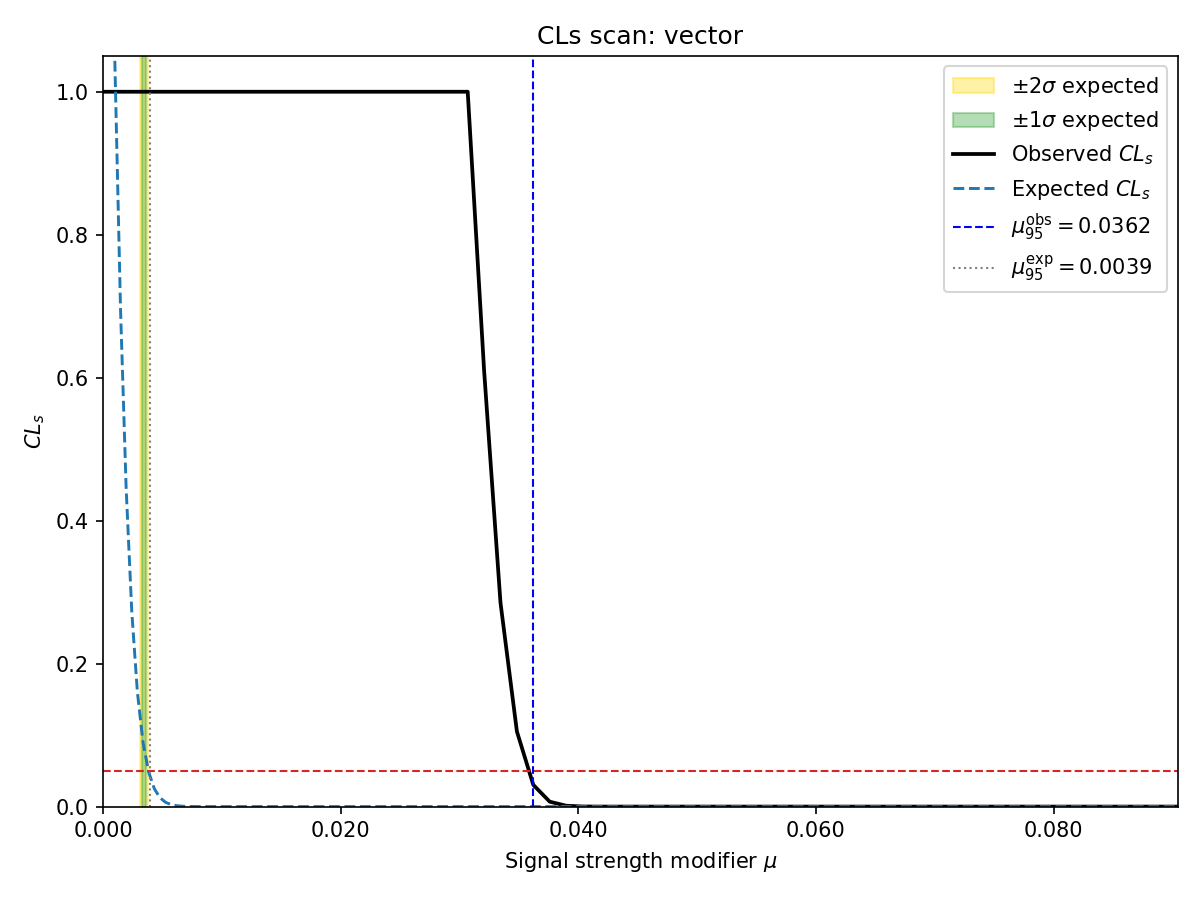}
\includegraphics[width=0.48\textwidth]{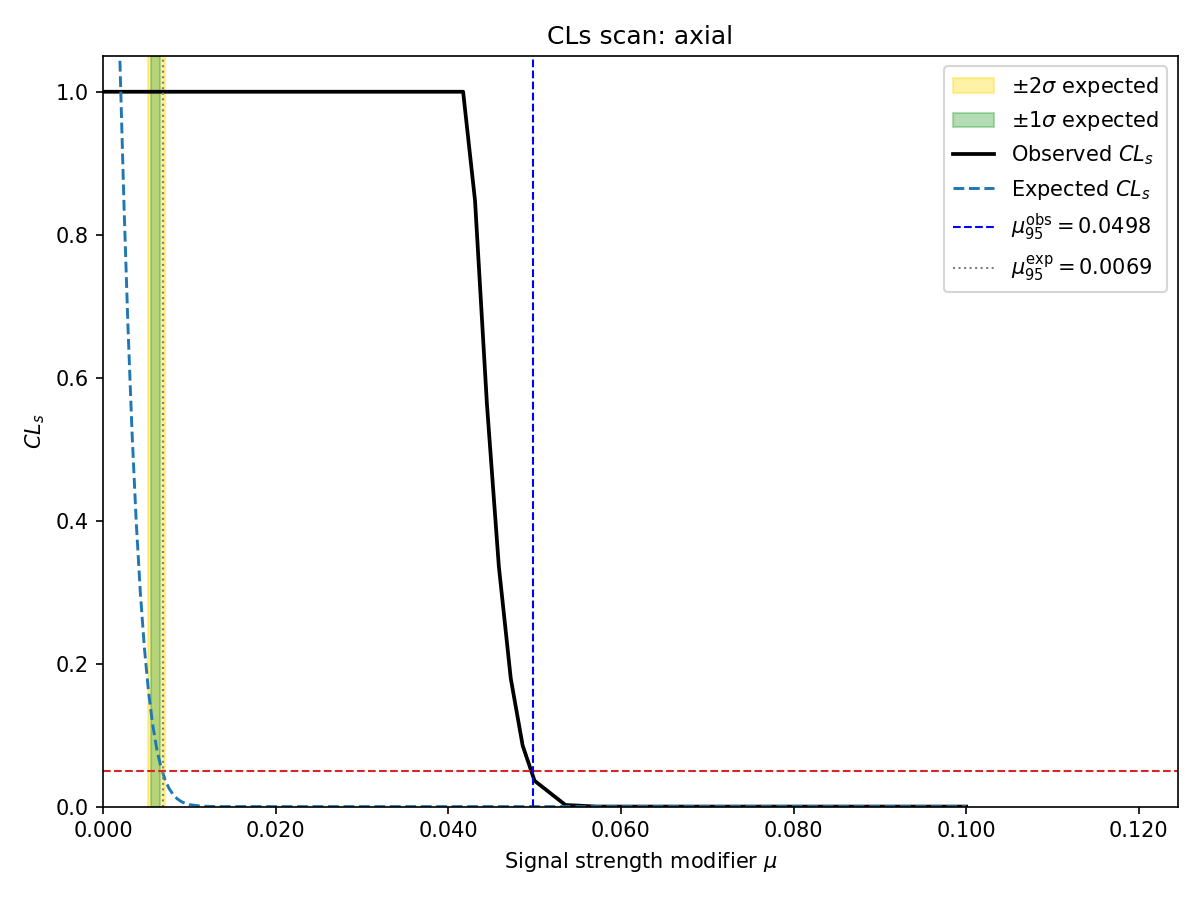}

\caption{
Asymptotic CL$_s$ scans for the scalar, vector, and axial-vector mediator
hypotheses from the simultaneous SR+VR fit. The VR constrains the background
normalisation; the SR provides the signal search sensitivity. The horizontal dashed line marks the
95\%~CL exclusion threshold; vertical lines indicate the observed and expected
upper limits on $\mu$.
}
\label{fig:cls_scans}
\end{figure*}
The normalizing-flow likelihood-ratio discriminant was evaluated using disjoint
control-region (CR), validation-region (VR), and signal-region (SR) samples.
The Standard Model density flows were trained only on CR events with
$\MET<50\,\GeV$. The final statistical interpretation uses the full SR
selection, $\MET\ge50\,\GeV$, $|\Delta\phi(\MET,Z)|>2.5$, and
$n_{\mathrm{jets}}\le1$. Although no explicit upper \MET\ bound is
imposed in the selection, the cleaned dataset caps \MET\ at
approximately 200\,\GeV, and the VR interval ($50\le\MET<100\,\GeV$) is
retained as a validation and stress-test sample rather than as the nominal
background template for the final limits.

For each mediator hypothesis the NSF score
\begin{equation}
S_h(\mathbf{x}) =
\log p(\mathbf{x}\mid\mathrm{DM}_h)
-
\log p(\mathbf{x}\mid\mathrm{SM}_{\ell\ell})
\end{equation}

was constructed separately for the vector, axial-vector, and scalar mediator hypotheses. Large positive values correspond to signal-like events, while negative values indicate SM-like kinematics.

Control-region closure tests show stable channel-specific behaviour of the
trained SM flows. The CR distributions of $\log p(\mathbf{x}\mid\mathrm{SM})$
peak near $\sim 125$ with similar widths in the $\mu\mu$ and $ee$ channels,
consistent with a well-converged density model on Drell--Yan-dominated events.

Validation-region score distributions show clear separation between observed
data and simulated dark matter events for all mediator hypotheses. The
strongest discrimination is observed for the scalar mediator, whose signal
distribution occupies a largely distinct region of score space. Consistent
behaviour is observed in both lepton channels, demonstrating that the NSF
score generalises across detector signatures and event topologies.

The SR score distributions are shown in Fig.~\ref{fig:combined_sr_scores}. In all cases, the observed data are concentrated at lower score values, whereas simulated dark matter events populate the high-score tail. The scalar hypothesis produces the largest signal-background separation, followed by the vector and axial-vector models.

A binned profile-likelihood fit was performed using the NSF score distributions.
Initial histograms were constructed using 30 score bins and subsequently merged
to ensure a minimum nominal background yield of 20 events per bin. The nominal
background shapes are derived from a single-step VR$\to$SR transfer, while the
final statistical interpretation uses a simultaneous SR+VR profile-likelihood
fit with independent 5\% normalization nuisance parameters in the $\mu\mu$
and $ee$ channels. The validation region acts as a normalisation sideband and
constrains the background prediction through nuisance-parameter profiling,
providing an independent cross-check of the SR-only result shown in
Appendix~\ref{app:sr-only-fit}.
A pure VR-extrapolated shape alternative was also investigated but failed closure in the high-score tail; the corresponding VR-only post-fit distributions are provided in Appendix~\ref{app:vr-extrapolation} as a diagnostic study and are not used for the final statistical interpretation.

The fitted score distributions for all mediator hypotheses and both lepton
channels are shown in Fig.~\ref{fig:fit_results}. In every case the observed
data show a pull relative to the VR-based background prediction, most
pronounced in the high-score tail (Section~\ref{sec:met-tail}).
The discovery test statistic \(q_0\) is large (233--327 across the three
hypotheses; Table~\ref{tab:fit_summary}), formally corresponding to a
significance capped at \(Z=8.0\) by floating-point underflow; we attribute
this to the characterized high-\MET\ background-modelling residual rather
than a dark matter signal (Sections~\ref{sec:met-tail}--\ref{sec:extrapolation-validation}).

Upper limits on the signal strength parameter $\mu$ were derived using the
asymptotic CL$_s$ method. The resulting limits are summarized in
Table~\ref{tab:fit_summary}, while the corresponding confidence-level scans
are shown in Fig.~\ref{fig:cls_scans}. The scalar mediator provides the
strongest exclusion, yielding an observed 95\% confidence-level upper limit
of $\mu < 0.0177$. The vector and axial-vector hypotheses yield observed
limits of $\mu < 0.0362$ and $\mu < 0.0498$, respectively.

For all three mediator hypotheses the observed limits exceed the median
expected limits by factors of roughly 7--12 across the nominal and cross-check
background constructions, reflecting the unresolved high-\MET\ tail
modelling residual described in Section~\ref{sec:met-tail}. The superior
performance of the scalar model is consistent with its significantly
stronger separation in NSF score space. The axial-vector hypothesis
exhibits the weakest sensitivity owing to the greater overlap between
signal and background score distributions.

\section{Summary and Outlook}
\label{sec:summary}

We presented a mono-$Z$ DM search that replaces hand-crafted cut flows with
likelihood-ratio scores from Neural Spline Flows trained on CR data and DM MC.
The workflow---open-data extraction, audited cleaning, disjoint CR/SR design, and
three combined $\mu\mu$+ee fits from five trained flows---provides a reproducible
template for density-based searches at the LHC. The final simultaneous SR+VR
fits yield observed 95\%~CL limits of $\mu<0.0177$ (scalar),
$\mu<0.0362$ (vector), and $\mu<0.0498$
(axial-vector), corresponding to the cross-section limits reported in
Table~\ref{tab:fit_summary}. The observed-to-expected gaps are driven by the
high-\MET\ tail modelling residual rather than by a dark matter signal.
A pure VR-extrapolated shape alternative fails closure in the high-score
tail (Appendix~\ref{app:vr-extrapolation}) and is not used for the final limits.

\clearpage
\onecolumn
\appendix

\section*{Appendix}
\addcontentsline{toc}{section}{Appendix}

\section{Dataset and Extraction Summary}
\label{sec:appendix}

\subsection{Rows read and extracted}
\label{app:rows}

Table~\ref{tab:appendix-rows} summarizes the row counts used after extraction.
For Run~2015D data, raw rows read are aggregated from the per-file audit CSVs under
\texttt{Data/Extracted}. For DM MC, the source files are ROOT files; the
Rows read column reports the total raw entries processed before extraction, and
the extracted rows are the selected event counts in the output TTrees and EDA summaries.

\begin{table}[h]
\centering
\scriptsize
\caption{Dataset-level rows read and rows extracted for the analysis inputs.}
\label{tab:appendix-rows}
\resizebox{0.6\columnwidth}{!}{%
\begin{tabular}{lrrr}
\toprule
Dataset source & Source files & Rows read & Rows extracted \\
\midrule
DoubleMuon data & 1068 & 51,342,919 & 1,738,984 \\
DoubleEG data & 3568 & 92,865,644 & 1,418,467 \\
DM vector MC & 1 & $\sim$50,000 & 27,602 \\
DM axial-vector MC & 4 & $\sim$100,000 & 45,710 \\
DM scalar MC & 1 & $\sim$50,000 & 32,792 \\
\bottomrule
\end{tabular}
}
\end{table}

\subsection{HLT trigger paths and b-tag configuration}
\label{app:trigger-btag}

Table~\ref{tab:appendix-trigger-btag} summarizes the trigger and $b$-tag settings
recorded by the extraction notebooks. Trigger decisions are read from
\texttt{selectedPatTrigger} path-name and final-filter collections when available; the
filter tokens are used as the fallback audit path for 2015 MiniAOD files with empty
path-name payloads. The data extractors apply the medium CSVv2 $b$-veto, while the DM MC
extractor retains \texttt{n\_bjets} only for schema compatibility.

\begin{table}[h]
\centering
\scriptsize
\caption{HLT trigger prefixes, fallback filter tokens, and $b$-tag configuration used during event extraction.}
\label{tab:appendix-trigger-btag}

\begin{tabular}{p{0.12\textwidth}p{0.30\textwidth}p{0.34\textwidth}p{0.14\textwidth}}
\toprule
\textbf{Channel} &
\textbf{HLT prefixes} &
\textbf{Fallback accepted-filter tokens} &
\textbf{$b$-tag treatment} \\
\midrule

DoubleMuon data &
\path{HLT_Mu17_TrkIsoVVL_Mu8_TrkIsoVVL_v},
\path{HLT_Mu17_TrkIsoVVL_TkMu8_TrkIsoVVL_v},
\path{HLT_IsoMu20_v},
\path{HLT_IsoTkMu20_v}
&
\path{hltDiMuonGlb17Glb8RelTrkIsoFiltered0p4},
\path{hltDiMuonGlb17Trk8RelTrkIsoFiltered0p4},
\path{hltDiMuonGlb17Glb8RelTrkIsoFiltered0p4DzFiltered0p2},
\path{hltDiMuonGlb17Trk8RelTrkIsoFiltered0p4DzFiltered0p2}
&
CSVv2 medium veto, WP = 0.800
\\

\addlinespace

DoubleEG data &
\path{HLT_Ele17_Ele12_CaloIdL_TrackIdL_IsoVL_DZ_v},
\path{HLT_Ele23_Ele12_CaloIdL_TrackIdL_IsoVL_DZ_v},
\path{HLT_Ele27_WPLoose_Gsf_v}
&
\path{hltEle17Ele12CaloIdLTrackIdLIsoVLDZFilter},
\path{hltEle23Ele12CaloIdLTrackIdLIsoVLDZFilter},
\path{hltEle27WPLooseGsfTrackIsoFilter},
\path{hltEle27noerWPLooseGsfTrackIsoFilter}
&
CSVv2 medium veto, WP = 0.800
\\

\addlinespace

MonoZToLL MC &
Dataset-level trigger definition; named HLT paths are audited when
\path{TriggerResults} are present.
&
All simulated events retained by the uproot-only trigger decoder.
&
\path{n_bjets} retained; no veto applied.
\\

\bottomrule
\end{tabular}
\end{table}
\clearpage
\subsection{Signal MC parameter points}
\label{app:signal-points}

Table~\ref{tab:appendix-signal-points} lists the signal samples used for the three DM
density models. The parameter values are taken from the CERN Open Data sample names in
the local bibliography.

\begin{table}[h]
\centering
\caption{MonoZToLL signal MC parameter points and extracted event counts.}
\label{tab:appendix-signal-points}
\resizebox{\textwidth}{!}{%
\begin{tabular}{llrrr}
\toprule
Mediator sample & CERN record / DOI & \(m_{\chi}\) [GeV] & Mediator scale [GeV] & Extracted events \\
\midrule
Vector, \texttt{V\_Mx-1\_Mv-500\_gDMgQ-1} & Record 16630 / 10.7483/OPENDATA.CMS.YTLH.E0N7 & 1 & \(m_V=500\) & 27,602 \\
Axial-vector, \texttt{A\_Mx-10\_Mv-20\_gDMgQ-1} & Record 16575 / 10.7483/OPENDATA.CMS.YB3Q.XTXY & 10 & \(m_V=20\) & part of combined 45,710 \\
Axial-vector, \texttt{A\_Mx-50\_Mv-200\_gDMgQ-1} & Record 16597 / 10.7483/OPENDATA.CMS.34IE.KN6I & 50 & \(m_V=200\) & part of combined 45,710 \\
Scalar, \texttt{EWK\_Scalar\_Mx-100\_Lambda-3000} & Record 16601 / 10.7483/OPENDATA.CMS.NW7F.NFGG & 100 & \(\Lambda=3000\) & 32,792 \\
\bottomrule
\end{tabular}
}
\end{table}
\clearpage 
\section{Feature Definitions}
\label{app:feature-definitions}

\subsection{Physics feature branches}
\label{app:features}

The extraction notebooks write the 40 physics branches listed in
Tables~\ref{tab:appendix-features} and~\ref{tab:appendix-features-cont} to the output \texttt{Events} TTrees. The DoubleMuon,
DoubleEG, and MonoZToLL DM extractors use the same branch set; data outputs add
audit/provenance branches, while DM MC adds \texttt{gen\_weight}. We use
\(\Delta\phi(a,b)=|((a-b+\pi)\bmod 2\pi)-\pi|\), \(m_Z=91.1876\,\GeV\), and
\(\epsilon=10^{-6}\) for protected divisions.

\begin{table*}[h]
\centering
\scriptsize
\caption{Physics feature branches written by the extraction notebooks, part 1 of 2. Derived features include the formula used in the analysis code.}
\label{tab:appendix-features}
\begin{tabular}{L{0.18\textwidth}L{0.16\textwidth}L{0.58\textwidth}}
\toprule
Branch & Category & Definition or formula \\
\midrule
\texttt{met\_pt} & MET & Missing transverse momentum magnitude read from the MiniAOD MET object. \\
\texttt{met\_phi} & MET & Azimuthal angle of the MiniAOD MET object. \\
\texttt{min\_dphi\_met\_jets} & Jet/MET angle & \(\min_j \Delta\phi(\phi_{\mathrm{MET}},\phi_j)\) over selected jets; filled with \(\pi\) when no selected jet is present. \\
\texttt{dphi\_met\_jet1} & Jet/MET angle & \(\Delta\phi(\phi_{\mathrm{MET}},\phi_{j1})\) for the leading selected jet; absent slot filled with the extraction sentinel, imputed at DataLoader time to \(0\), and masked with \texttt{has\_jet1}. \\
\texttt{dphi\_met\_jet2} & Jet/MET angle & \(\Delta\phi(\phi_{\mathrm{MET}},\phi_{j2})\) for the subleading selected jet; absent slot filled with the extraction sentinel, imputed at DataLoader time to \(0\), and masked with \texttt{has\_jet2}. \\
\texttt{met\_over\_sqrtHT} & MET/recoil & \(\MET/\sqrt{H_{\mathrm{T}}}\) for \(H_{\mathrm{T}}>5\,\GeV\), otherwise 0. \\
\texttt{hadronic\_recoil\_pt} & Recoil & With \(h_x=-(\MET\cos\phi_{\mathrm{MET}}+p_{\mathrm{T}}^Z\cos\phi_Z)\) and \(h_y=-(\MET\sin\phi_{\mathrm{MET}}+p_{\mathrm{T}}^Z\sin\phi_Z)\), \(\sqrt{h_x^2+h_y^2}\). \\
\texttt{mll} & Dilepton & \(\sqrt{\max[(E_1+E_2)^2-(p_x^Z{}^2+p_y^Z{}^2+p_z^Z{}^2),0]}\). \\
\texttt{pt\_Z} & Dilepton/Z & \(\sqrt{(p_{x1}+p_{x2})^2+(p_{y1}+p_{y2})^2}\), with \(p_{xi}=p_{\mathrm{T}i}\cos\phi_i\), \(p_{yi}=p_{\mathrm{T}i}\sin\phi_i\). \\
\texttt{eta\_Z\_pseudo} & Dilepton/Z & \(\operatorname{asinh}(p_z^Z/p_{\mathrm{T}}^Z)\) for \(p_{\mathrm{T}}^Z>1\,\GeV\), otherwise 0. \\
\texttt{phi\_Z} & Dilepton/Z & \(\operatorname{atan2}(p_y^Z,p_x^Z)\). \\
\texttt{dphi\_met\_Z} & MET/Z angle & \(\Delta\phi(\phi_{\mathrm{MET}},\phi_Z)\). \\
\texttt{met\_over\_ptZ} & MET/Z & \(\MET/\max(|p_{\mathrm{T}}^Z|,\epsilon)\), clipped to \([0,10]\) at extraction. \\
\texttt{abs\_met\_minus\_ptZ} & MET/Z & \(|\MET-p_{\mathrm{T}}^Z|\). \\
\texttt{abs\_mll\_minus\_mZ} & Dilepton & \(|m_{\ell\ell}-m_Z|\). \\
\texttt{u\_parallel} & Recoil & \(h_x\cos\phi_Z+h_y\sin\phi_Z\). \\
\texttt{u\_perp} & Recoil & \(-h_x\sin\phi_Z+h_y\cos\phi_Z\). \\
\texttt{u\_parallel\_over\_ptZ} & Recoil/Z & \(u_{\parallel}/\max(|p_{\mathrm{T}}^Z|,\epsilon)\), clipped to \([-10,10]\). \\
\texttt{u\_perp\_over\_ptZ} & Recoil/Z & \(u_{\perp}/\max(|p_{\mathrm{T}}^Z|,\epsilon)\), clipped to \([-10,10]\). \\
\texttt{lep1\_pt} & Lepton & Leading selected lepton transverse momentum. \\
\texttt{lep1\_eta} & Lepton & Leading selected lepton pseudorapidity. \\
\texttt{lep2\_pt} & Lepton & Subleading selected lepton transverse momentum. \\
\texttt{lep2\_eta} & Lepton & Subleading selected lepton pseudorapidity. \\
\bottomrule
\end{tabular}
\end{table*}

\begin{table*}[t]
\centering
\scriptsize
\caption{Physics feature branches written by the extraction notebooks, part 2 of 2.}
\label{tab:appendix-features-cont}
\begin{tabular}{L{0.18\textwidth}L{0.16\textwidth}L{0.58\textwidth}}
\toprule
Branch & Category & Definition or formula \\
\midrule
\texttt{deltaR\_ll} & Dilepton angle & \(\sqrt{(\eta_1-\eta_2)^2+\Delta\phi(\phi_1,\phi_2)^2}\). \\
\texttt{deltaphi\_ll} & Dilepton angle & \(\Delta\phi(\phi_1,\phi_2)\). \\
\texttt{lepton\_flavor} & Channel & Absolute PDG lepton ID: 13 for muon-pair events and 11 for electron-pair events. \\
\texttt{lep1\_reliso} & Lepton ID & Leading lepton relative isolation from MiniAOD isolation components. \\
\texttt{lep2\_reliso} & Lepton ID & Subleading lepton relative isolation from MiniAOD isolation components. \\
\texttt{cos\_theta\_star} & Dilepton angle & Cosine of the negative lepton angle in the reconstructed dilepton rest frame, computed by boosting the negative lepton into the \(Z\)-candidate rest frame and projecting onto the reconstructed \(Z\) direction. \\
\texttt{lep1\_dxy\_sig} & Lepton ID & Leading lepton transverse impact-parameter significance, \(|d_{xy}|/\sigma(d_{xy})\). \\
\texttt{dphi\_met\_lep1} & MET/lepton angle & \(\Delta\phi(\phi_{\mathrm{MET}},\phi_{\ell1})\). \\
\texttt{n\_jets} & Jets & Number of selected jets with \(p_{\mathrm{T}}>30\,\GeV\), \(|\eta|<2.4\), and lepton-overlap removal. \\
\texttt{n\_bjets} & Jets & Number of selected jets passing the medium CSVv2 working point in data; retained in DM MC for schema parity. \\
\texttt{jet1\_pt} & Jets & Leading selected-jet transverse momentum; absent slot filled with \(0\) and masked with \texttt{has\_jet1}. \\
\texttt{jet1\_eta} & Jets & Leading selected-jet pseudorapidity; absent slot filled with the extraction sentinel, imputed at DataLoader time to \(0\), and masked with \texttt{has\_jet1}. \\
\texttt{jet2\_pt} & Jets & Subleading selected-jet transverse momentum; absent slot filled with \(0\) and masked with \texttt{has\_jet2}. \\
\texttt{jet2\_eta} & Jets & Subleading selected-jet pseudorapidity; absent slot filled with the extraction sentinel, imputed at DataLoader time to \(0\), and masked with \texttt{has\_jet2}. \\
\texttt{HT} & Jets & Scalar sum of selected-jet transverse momenta. \\
\texttt{n\_vertices} & Event & Number of reconstructed primary vertices. \\
\texttt{rho} & Event & Event energy-density variable read from the MiniAOD fixed-grid \(\rho\) branch. \\
\midrule
\multicolumn{3}{l}{DataLoader-computed features:} \\
\texttt{has\_jet1} & Jet indicator & \(\mathbb{1}[n_\mathrm{jets}\ge 1]\); computed at DataLoader time, not stored in the extracted ROOT TTree. Appended to the NSF input vector to allow the flow to condition on jet presence. \\
\texttt{has\_jet2} & Jet indicator & \(\mathbb{1}[n_\mathrm{jets}\ge 2]\); computed at DataLoader time. Appended to the NSF input vector to allow the flow to condition on second-jet presence. \\
\bottomrule
\end{tabular}
\end{table*}

\subsection{NSF input feature order}
\label{app:nsf-feature-order}

Table~\ref{tab:appendix-nsf-feature-order} gives the 37-dimensional input order used when
loading the trained NSF checkpoints and scoring events. This order is read from
\texttt{Output/results/scoring\_audit.json}; \texttt{has\_jet1} and \texttt{has\_jet2}
are computed from \texttt{n\_jets} before clipping.

\begin{table*}[t]
\centering
\scriptsize
\caption{NSF input feature order used for SM and DM scoring.}
\label{tab:appendix-nsf-feature-order}
\resizebox{\textwidth}{!}{%
\begin{tabular}{rlrlrl}
\toprule
Index & Feature & Index & Feature & Index & Feature \\
\midrule
1 & \texttt{met\_pt} & 14 & \texttt{abs\_met\_minus\_ptZ} & 27 & \texttt{lep2\_reliso} \\
2 & \texttt{met\_phi} & 15 & \texttt{abs\_mll\_minus\_mZ} & 28 & \texttt{cos\_theta\_star} \\
3 & \texttt{min\_dphi\_met\_jets} & 16 & \texttt{u\_parallel} & 29 & \texttt{lep1\_dxy\_sig} \\
4 & \texttt{dphi\_met\_jet1} & 17 & \texttt{u\_perp} & 30 & \texttt{dphi\_met\_lep1} \\
5 & \texttt{dphi\_met\_jet2} & 18 & \texttt{u\_parallel\_over\_ptZ} & 31 & \texttt{jet1\_pt} \\
6 & \texttt{met\_over\_sqrtHT} & 19 & \texttt{u\_perp\_over\_ptZ} & 32 & \texttt{jet1\_eta} \\
7 & \texttt{hadronic\_recoil\_pt} & 20 & \texttt{lep1\_pt} & 33 & \texttt{jet2\_pt} \\
8 & \texttt{mll} & 21 & \texttt{lep1\_eta} & 34 & \texttt{jet2\_eta} \\
9 & \texttt{pt\_Z} & 22 & \texttt{lep2\_pt} & 35 & \texttt{has\_jet1} \\
10 & \texttt{eta\_Z\_pseudo} & 23 & \texttt{lep2\_eta} & 36 & \texttt{has\_jet2} \\
11 & \texttt{phi\_Z} & 24 & \texttt{deltaR\_ll} & 37 & \texttt{n\_jets} \\
12 & \texttt{dphi\_met\_Z} & 25 & \texttt{deltaphi\_ll} & & \\
13 & \texttt{met\_over\_ptZ} & 26 & \texttt{lep1\_reliso} & & \\
\bottomrule
\end{tabular}
}
\end{table*}

\clearpage
\section{Data Cleaning}
\label{app:data-cleaning}

\subsection{Fixed domain bounds}
\label{app:cleaning}

Table~\ref{tab:appendix-cleaning-fixed} lists the fixed-domain bounds applied by the SM
cleaner. Features not shown in the fixed-bound table are either pass-through variables or
receive only the tail-rule treatment in Table~\ref{tab:appendix-cleaning-tail}.

\begin{table*}[h]
\centering
\caption{Fixed bounds applied during SM cleaning.}
\label{tab:appendix-cleaning-fixed}
\resizebox{\textwidth}{!}{%
\begin{tabular}{ll}
\toprule
Feature(s) & Bound \\
\midrule
\texttt{met\_phi}, \texttt{phi\_Z} & \([-\pi,\pi]\) \\
\texttt{min\_dphi\_met\_jets}, \texttt{dphi\_met\_jet1}, \texttt{dphi\_met\_jet2}, \texttt{dphi\_met\_Z}, \texttt{dphi\_met\_lep1}, \texttt{deltaphi\_ll} & \([0,\pi]\) \\
\texttt{mll} & \([60,120]\,\GeV\) \\
\texttt{eta\_Z\_pseudo} & \([-5,5]\) \\
\texttt{lep1\_eta}, \texttt{lep2\_eta}, \texttt{jet1\_eta}, \texttt{jet2\_eta} & \([-2.4,2.4]\) \\
\texttt{met\_over\_ptZ} & \([0,10]\) \\
\texttt{u\_parallel\_over\_ptZ}, \texttt{u\_perp\_over\_ptZ} & \([-10,10]\) \\
\texttt{abs\_mll\_minus\_mZ}, \texttt{deltaR\_ll}, \texttt{n\_jets}, \texttt{n\_bjets}, \texttt{HT}, \texttt{n\_vertices}, \texttt{rho} & lower bound 0 \\
\texttt{lep1\_reliso}, \texttt{lep2\_reliso} & \([0,0.15]\) \\
\texttt{cos\_theta\_star} & \([-1,1]\) \\
\texttt{lepton\_flavor}, \texttt{gen\_weight} & pass-through \\
\bottomrule
\end{tabular}
}
\end{table*}

\subsection{Data-derived tail caps}
\label{app:tail-caps}

Table~\ref{tab:appendix-cleaning-caps} lists the numerical caps stored in
\texttt{Output/sm\_feature\_bounds.json} for variables governed by data-derived tail
rules. Positive rules clip to \([0,\mathrm{cap}]\), while symmetric rules clip to
\([-\mathrm{cap},+\mathrm{cap}]\).

\begin{table}[h]
\centering
\small
\caption{Data-derived tail caps used by the SM feature-domain cleaner.}
\label{tab:appendix-cleaning-caps}
\begin{tabular}{lrrl}
\toprule
Feature & Lower bound & Upper bound & Rule family \\
\midrule
\texttt{met\_pt} & 0.000 & 200.000 & positive \\
\texttt{met\_over\_sqrtHT} & 0.000 & 20.516 & positive \\
\texttt{hadronic\_recoil\_pt} & 0.000 & 564.334 & positive \\
\texttt{pt\_Z} & 0.000 & 556.310 & positive \\
\texttt{abs\_met\_minus\_ptZ} & 0.000 & 517.988 & positive \\
\texttt{u\_parallel} & -563.434 & 563.434 & symmetric \\
\texttt{u\_perp} & -158.807 & 158.807 & symmetric \\
\texttt{lep1\_pt} & 0.000 & 433.151 & positive \\
\texttt{lep2\_pt} & 0.000 & 182.855 & positive \\
\texttt{lep1\_dxy\_sig} & 0.000 & 10.000 & positive, reference-min \\
\texttt{jet1\_pt} & 0.000 & 629.442 & positive \\
\texttt{jet2\_pt} & 0.000 & 422.450 & positive \\
\texttt{HT} & 0.000 & 1071.170 & positive \\
\bottomrule
\end{tabular}

\end{table}

\subsection{Tail-rule families}
\label{app:tail-rule-families}

\begin{table}[H]
\centering
\caption{High-quantile tail rules used by the SM cleaner. A positive rule bounds the feature below by zero and above by the estimated cap; a symmetric rule applies \([-\mathrm{cap},+\mathrm{cap}]\).}
\label{tab:appendix-cleaning-tail}
\resizebox{\textwidth}{!}{%
\begin{tabular}{lll}
\toprule
Feature(s) & Rule type & Minimum cap \\
\midrule
\texttt{met\_pt}, \texttt{abs\_met\_minus\_ptZ} & positive & 200 \\
\texttt{met\_over\_sqrtHT} & positive & 10 \\
\texttt{hadronic\_recoil\_pt}, \texttt{pt\_Z} & positive & 300 \\
\texttt{u\_parallel}, \texttt{u\_perp} & symmetric & 80 \\
\texttt{lep1\_pt}, \texttt{jet1\_pt} & positive & 150, 100 \\
\texttt{lep2\_pt}, \texttt{jet2\_pt} & positive & 80, 50 \\
\texttt{lep1\_dxy\_sig} & positive, cross-channel reference-min & 10 \\
\texttt{HT} & positive & 150 \\
\bottomrule
\end{tabular}
}
\end{table}

\clearpage
\section{Region Definitions and Event Yields}
\label{app:regions-yields}

\subsection{Region selection criteria}
\label{app:region-criteria}

Table~\ref{tab:appendix-region-criteria} records the CR, VR, and SR masks from
\texttt{Output/results/scoring\_audit.json}. The VR is nested in the SR kinematic
definition and is used as a 50--100~\GeV\ closure and stress-test region.

\begin{table}[h]
\centering
\caption{Analysis region definitions used during scoring.}
\label{tab:appendix-region-criteria}

\begin{tabular}{ll}
\toprule
Region & Selection \\
\midrule
CR & \(\MET < 50\,\GeV\) \\
VR & \(50 \le \MET < 100\,\GeV\), \(|\Delta\phi(\MET,Z)|>2.5\), \(n_{\mathrm{jets}}\le 1\) \\
SR & \(\MET \ge 50\,\GeV\), \(|\Delta\phi(\MET,Z)|>2.5\), \(n_{\mathrm{jets}}\le 1\) \\
\bottomrule
\end{tabular}

\end{table}

\subsection{Event counts per region and channel}
\label{app:region-yields}

Table~\ref{tab:appendix-region-yields} summarizes the region counts stored in the scoring
audit. DM MC is scored in both SM channel scaler spaces, so the mediator-specific VR and
SR counts are identical for the \(\mu\mu\) and \(ee\) scoring passes.

\begin{table}[h]
\centering
\small
\caption{Event counts after cleaning and region assignment.}
\label{tab:appendix-region-yields}

\begin{tabular}{lrrrr}
\toprule
Sample & Total & CR & VR & SR \\
\midrule
SM \(\mu\mu\) data & 1,738,984 & 1,709,765 & 6,970 & 7,151 \\
SM \(ee\) data & 1,418,467 & 1,392,023 & 6,644 & 6,804 \\
DM vector MC & 27,602 & -- & 5,608 & 16,259 \\
DM axial-vector MC & 45,710 & -- & 9,983 & 18,443 \\
DM scalar MC & 32,792 & -- & 378 & 25,884 \\
\bottomrule
\end{tabular}

\end{table}

\clearpage
\section{Fit Configuration and Numerical Results}
\label{app:fit-results}

\subsection{Fit configuration}
\label{app:fit-configuration}

Table~\ref{tab:appendix-fit-config} lists the final simultaneous SR+VR fit settings
recorded in \texttt{paper/json/sr+vr/fit\_summary.json} and associated output files.
The fit combines the \(\mu\mu\) and \(ee\) score histograms for each mediator hypothesis.

\begin{table}[h]
\centering
\caption{Profile-likelihood fit configuration.}
\label{tab:appendix-fit-config}
\resizebox{\columnwidth}{!}{%
\begin{tabular}{ll}
\toprule
Setting & Value \\
\midrule
Statistical method & \(\chi^2\) asymptotic CL\(_s\) approximation \\
Initial score bins & 30 \\
Minimum background per merged bin & 20.0 \\
Signal-strength scan points & 67 \\
Fit regions & SR + VR simultaneous \\
VR normalisation constraint & Per-channel $\theta_{\rm norm}$, prior width 0.05 \\
Normalization nuisance width & 0.05 per channel \\
Random seed & 42 \\
Fit packages & \texttt{numpy 1.26.4}, \texttt{scipy}, \texttt{iminuit 2.32.0} \\
\bottomrule
\end{tabular}
}
\end{table}

\subsection{Full fit results}
\label{app:full-fit-results}

Table~\ref{tab:appendix-fit-results} reports the numerical results from
\texttt{paper/json/sr+vr/fit\_summary.json}. All three signal-plus-background
fits converged; the large \(q_0\) values are attributed to the
high-\MET\ background-modelling residual discussed in
Section~\ref{sec:met-tail}.

\begin{table*}[h]
\centering
\caption{Observed and expected fit results for the three mediator hypotheses from the simultaneous SR+VR profile-likelihood fit.}
\label{tab:appendix-fit-results}
\resizebox{\textwidth}{!}{%
\begin{tabular}{lrrrrrr}
\toprule
Mediator & \(\hat{\mu}\) & \(\sigma_{\mu}\) & \(\mu^{95}_{\mathrm{obs}}\) & \(\mu^{95}_{\mathrm{exp}}\) & 68\% exp. band & 95\% exp. band \\
\midrule
Vector      & 0.0308 & 0.00245 & 0.0362 & 0.0039 &
  [0.00326, 0.00364] & [0.00309, 0.00382] \\
Axial-vector & 0.0424 & 0.00357 & 0.0498 & 0.0069 &
  [0.00560, 0.00659] & [0.00517, 0.00715] \\
Scalar      & 0.0141 & 0.00136 & 0.0177 & 0.0018 &
  [0.00156, 0.00163] & [0.00154, 0.00166] \\
\bottomrule
\end{tabular}
}
\end{table*}

\subsection{Background estimation method note}
\label{app:background-method}

The simultaneous SR+VR fit uses a single-step VR$\to$SR shape transfer:
the SM VR score histogram is renormalised to the region yield and used
as the nominal background template for that region. The VR component of
the likelihood constrains the per-channel normalisation nuisances
$\theta_{\rm norm}$ using $\sim$6,970 ($\mu\mu$) and $\sim$6,644 ($ee$)
VR events; the posterior uncertainty on $\theta_{\rm norm}$ is
$\sim$0.17 in sigma units, significantly tighter than the 5\% Gaussian
prior alone. Signal MC contributes to both SR and VR with the same
signal strength $\mu$, so the VR signal contamination is accounted for.
This approach avoids a pure SR-direct circularity while keeping the SR
score shape as the primary discriminant.
Bins with zero nominal background are excluded from the likelihood.
\clearpage
\subsection{VR-extrapolated validation fit}
\label{app:vr-extrapolation}

As a sideband stress test, an alternative fit used the SM VR
(\(50\le\MET<100\,\GeV\)) score histograms, normalized to the SR yield, as the
nominal background template. This procedure does not close in the high-score tail:
the VR template underpredicts the SR tail, and the signal-plus-background fit absorbs
the resulting shape difference as a positive signal strength. Table~\ref{tab:appendix-vr-fit}
therefore records the VR-extrapolated fit as a diagnostic rather than as the final
limit result.

\begin{table*}[t]
\centering
\caption{VR-extrapolated validation fit. The large positive fitted signal
strengths and weakened observed limits indicate VR-to-SR shape bias; these
numbers are not used for the final interpretation.}
\label{tab:appendix-vr-fit}
\resizebox{\textwidth}{!}{%
\begin{tabular}{lrrrrrr}
\toprule
Mediator & \(\hat{\mu}\) & \(\sigma_{\mu}\) & \(Z\) &
\(\mu^{95}_{\mathrm{obs}}\) & \(\mu^{95}_{\mathrm{exp}}\) & 68\% exp. band / 95\% exp. band \\
\midrule
Vector & 0.0392 & 0.00286 & 8.0 & 0.0454 & 0.0043 &
[0.00359, 0.00399] / [0.00339, 0.00421] \\
Axial-vector & 0.0587 & 0.00434 & 8.0 & 0.0694 & 0.0073 &
[0.00597, 0.00701] / [0.00554, 0.00761] \\
Scalar & 0.0160 & 0.00132 & 8.0 & 0.0195 & 0.0018 &
[0.00151, 0.00163] / [0.00145, 0.00168] \\
\bottomrule
\end{tabular}
}
\end{table*}

With a signal-strength scan extended to bracket \(\hat{\mu}\) for all three
mediators, the VR-extrapolated fit yields observed limits 9.5--10.8\(\times\)
weaker than expected, consistent in direction and magnitude with the nominal
fit's 7.2--9.8\(\times\) gap (Table~\ref{tab:fit_summary}). This corroborates
that the high-\MET\ tail residual, not a background-construction artifact
specific to one method, drives the observed/expected discrepancy.
\subsection{SR-only validation fit}
\label{app:sr-only-fit}
\begin{figure}[h]
\centering

\includegraphics[width=0.48\textwidth]{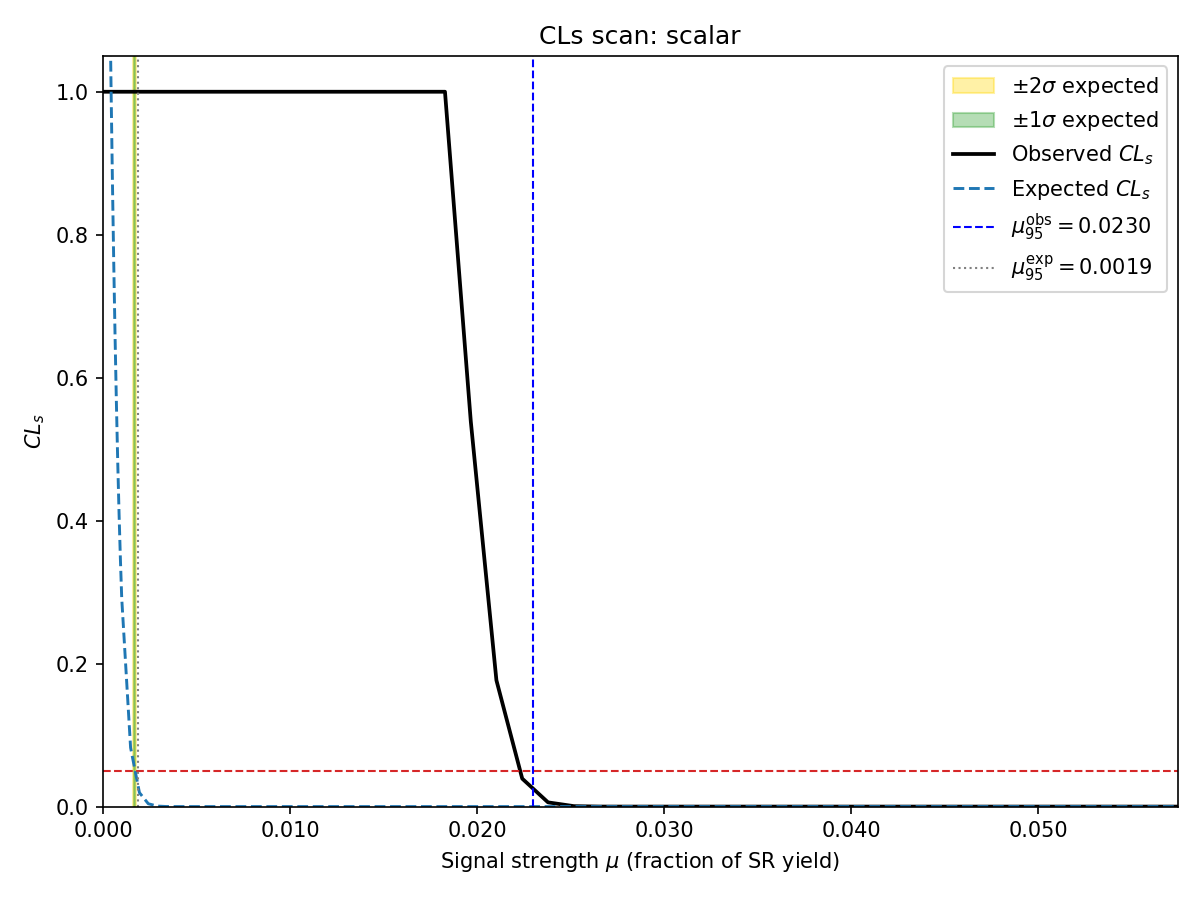}
\includegraphics[width=0.48\textwidth]{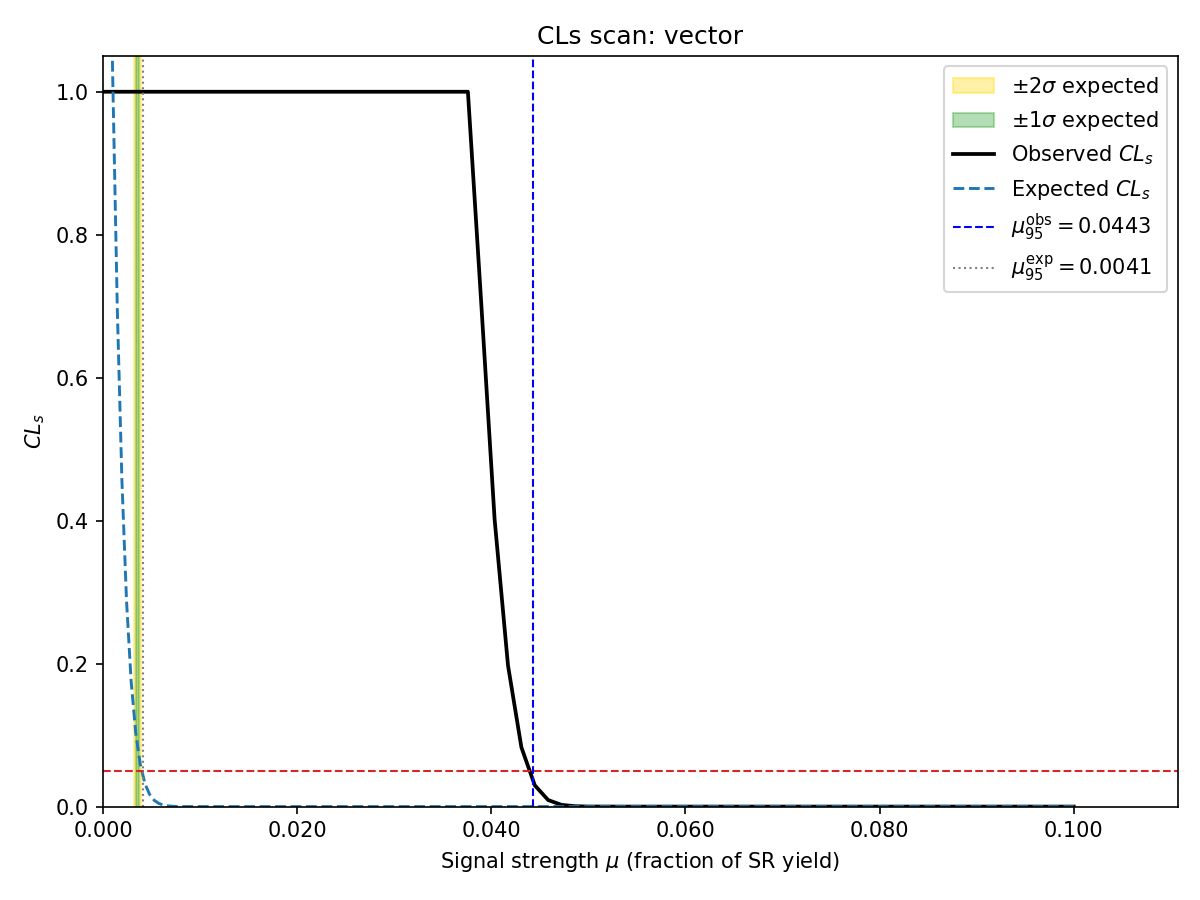}
\includegraphics[width=0.48\textwidth]{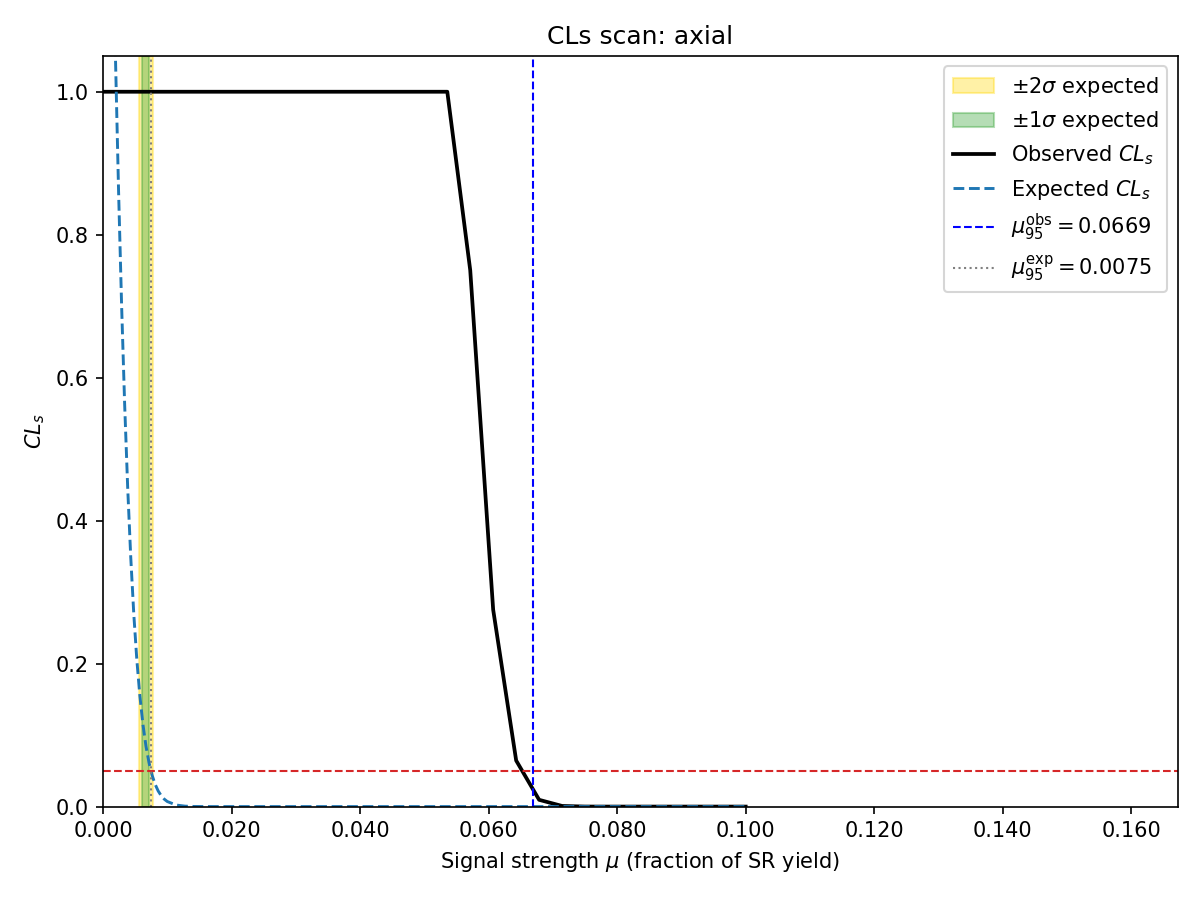}

\caption{
Asymptotic CL$_s$ scans for the scalar, vector, and axial-vector mediator
hypotheses from the SR-direct fit. The horizontal dashed line marks the
95\%~CL exclusion threshold; vertical lines indicate the observed and expected
upper limits on $\mu$.
}
\label{fig:sr_cls_scans}
\end{figure}

\begin{figure}[h]
\centering

\includegraphics[width=0.48\textwidth,height=0.23\textheight]{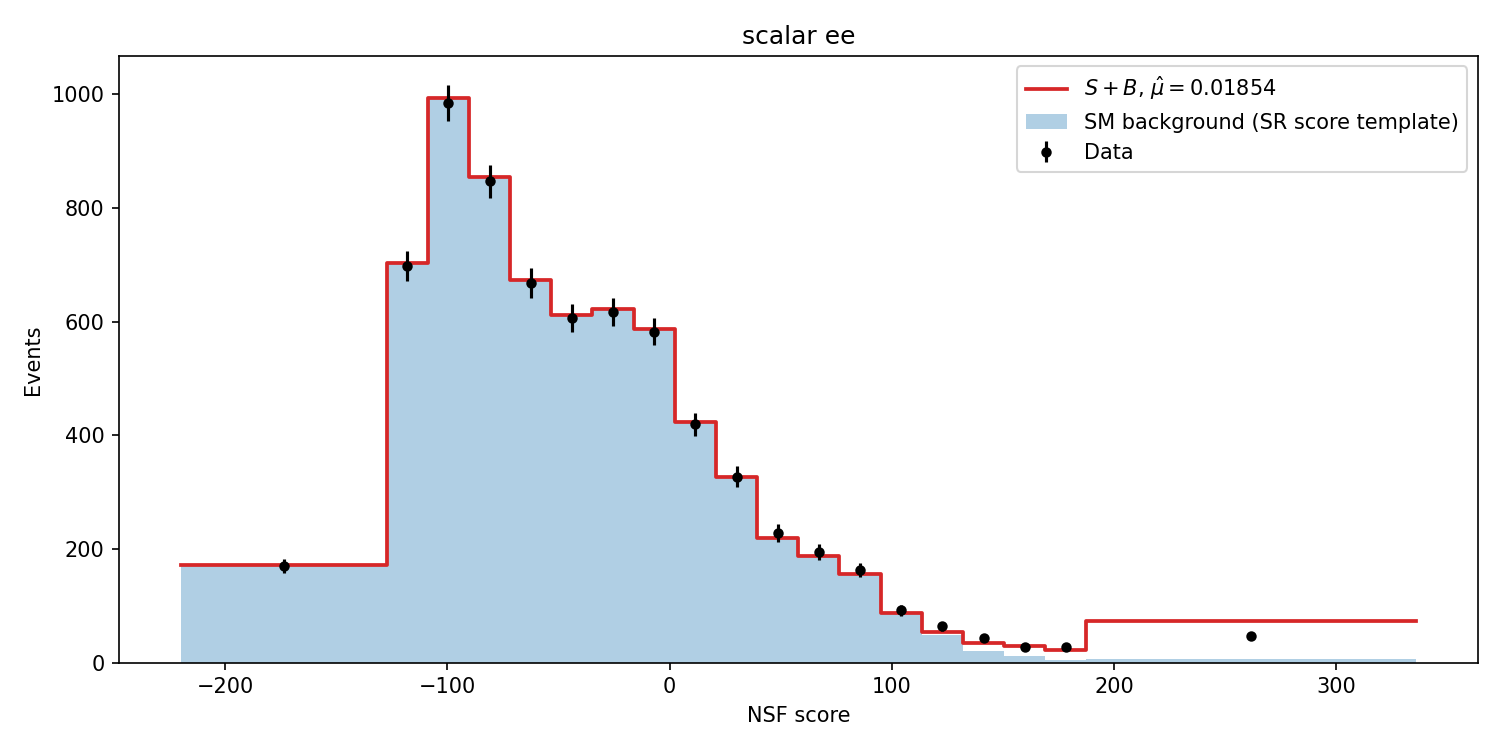}
\includegraphics[width=0.48\textwidth,height=0.23\textheight]{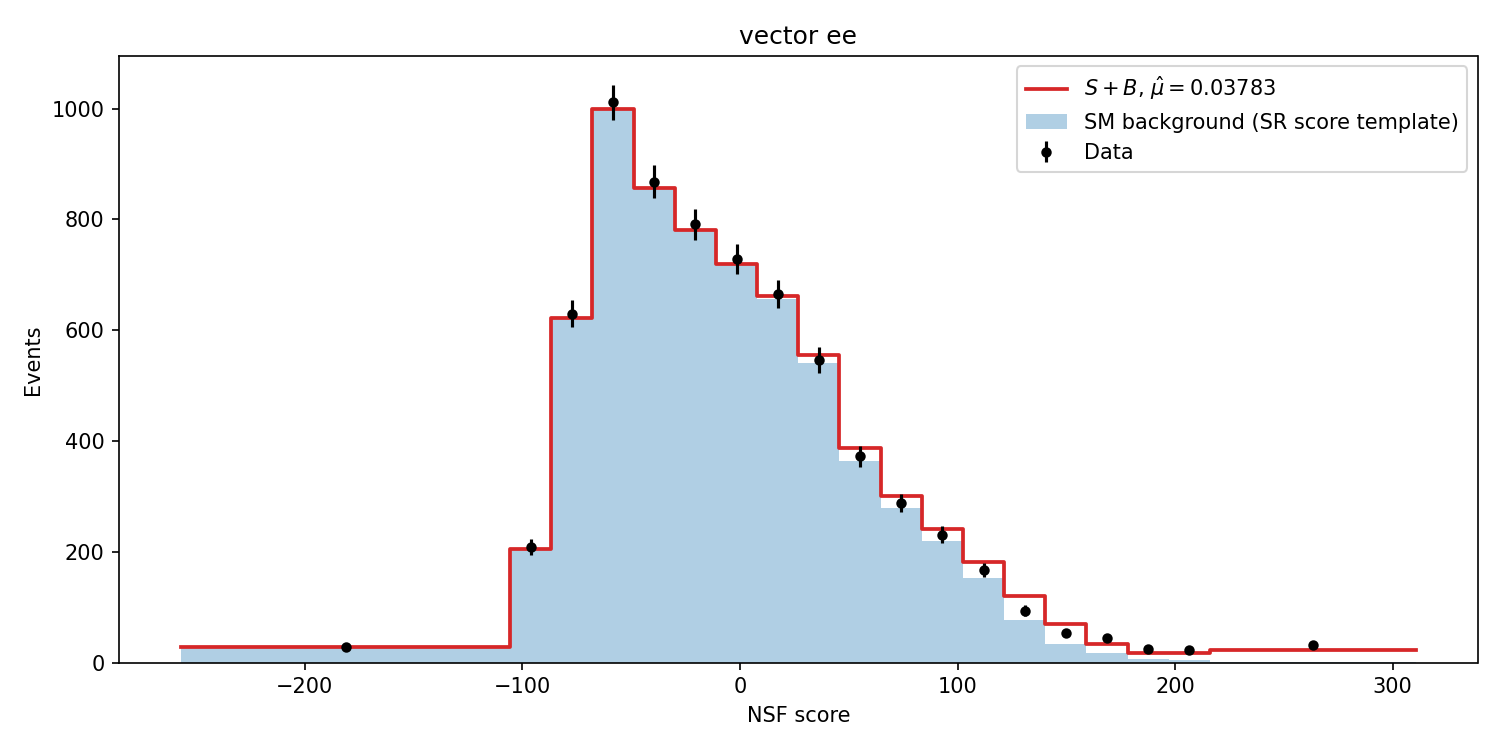}
\includegraphics[width=0.48\textwidth,height=0.23\textheight]{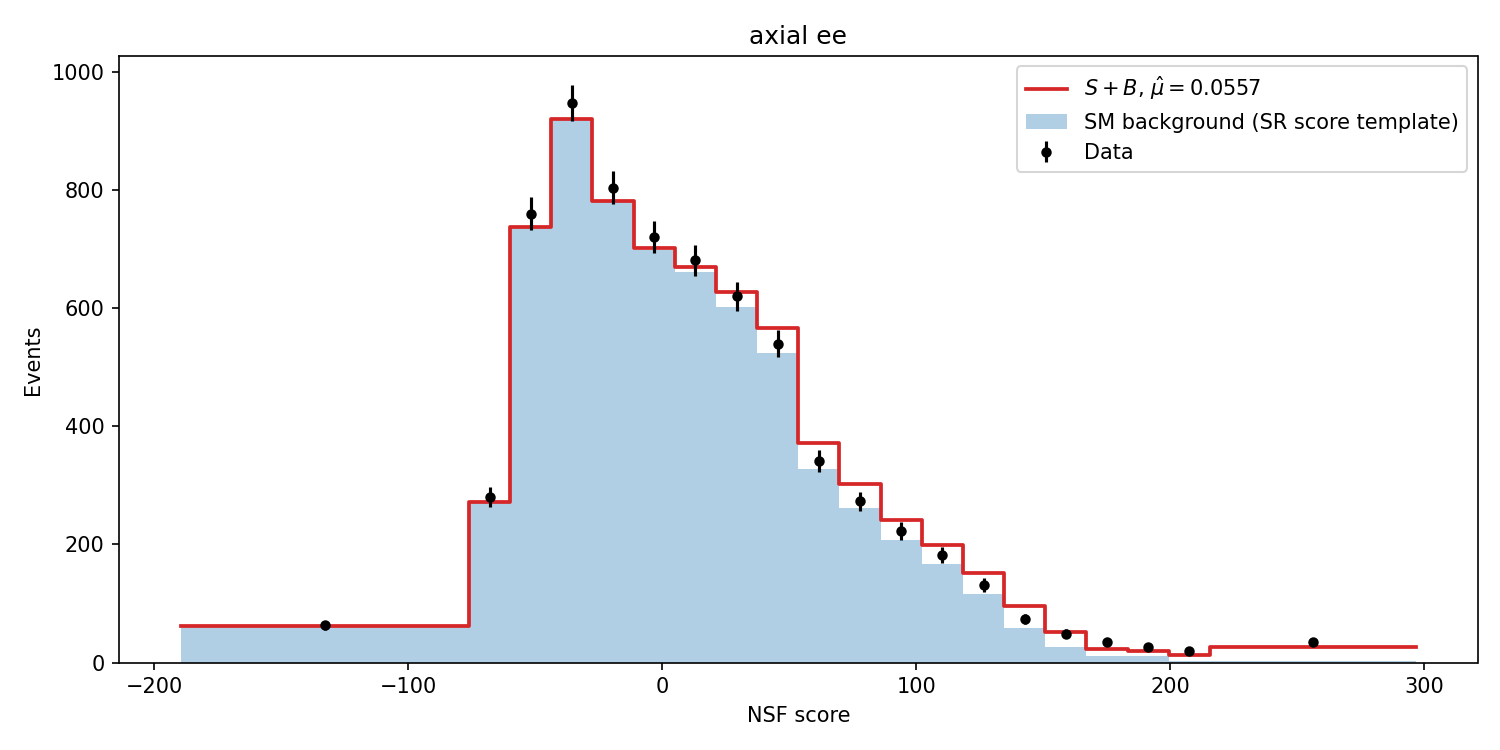}
\includegraphics[width=0.48\textwidth,height=0.23\textheight]{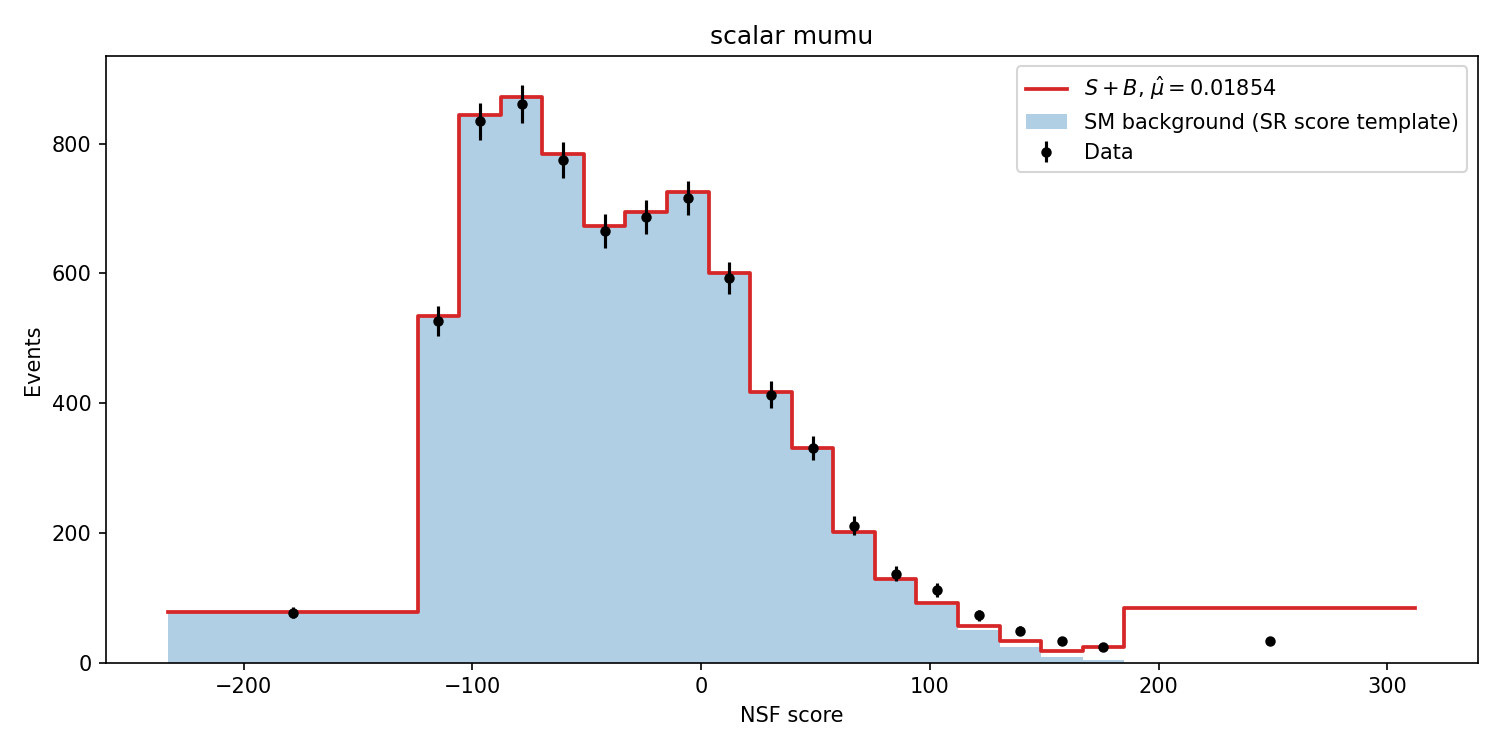}
\includegraphics[width=0.48\textwidth,height=0.23\textheight]{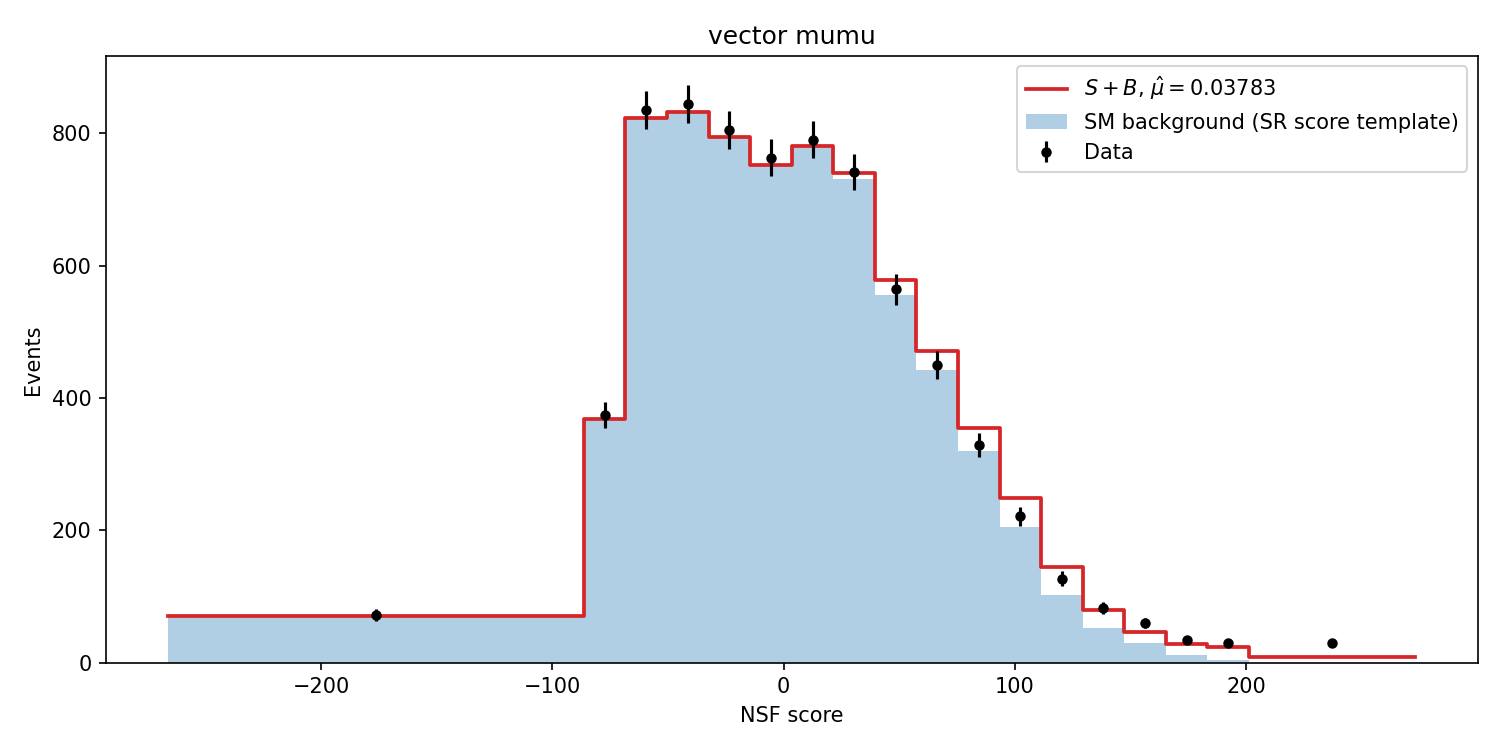}
\includegraphics[width=0.48\textwidth,height=0.23\textheight]{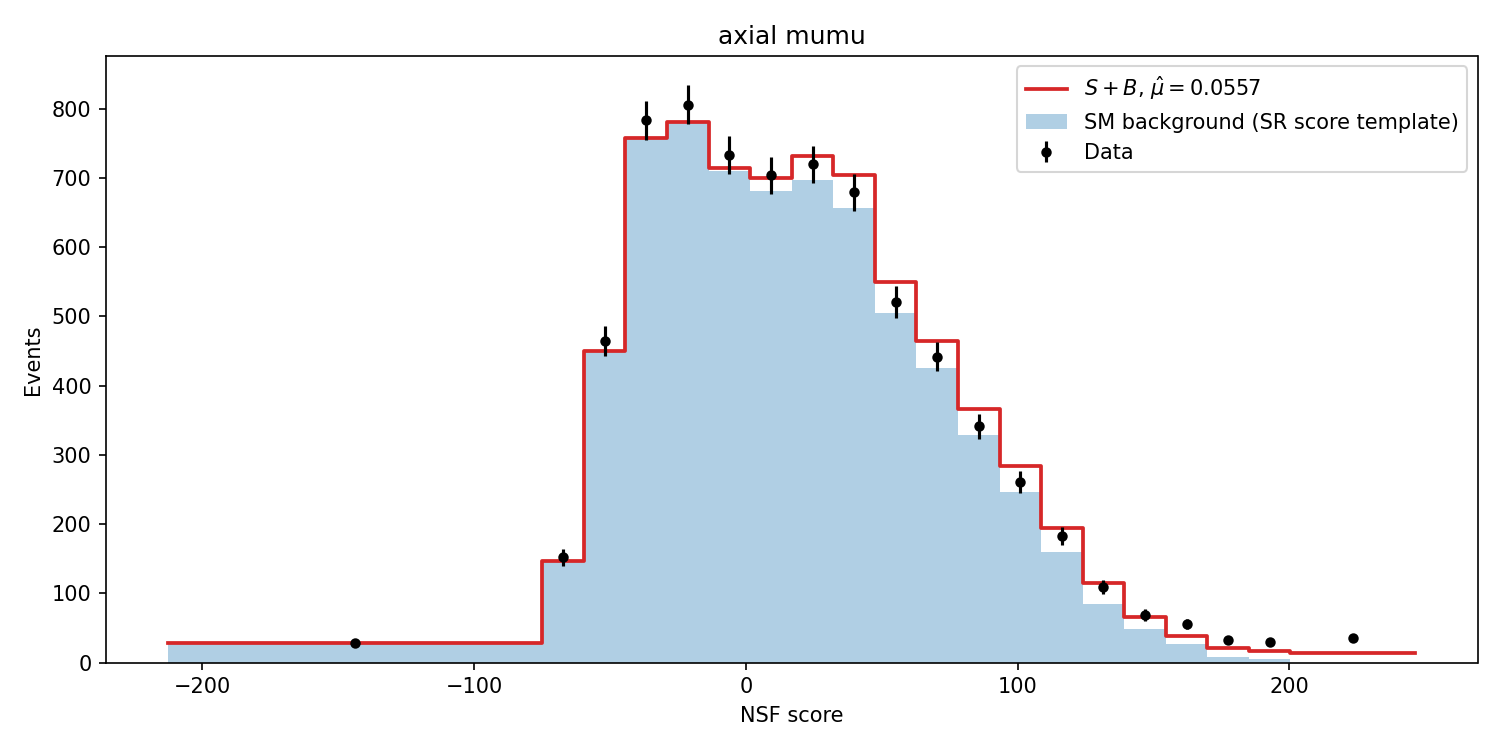}

\caption{
Post-fit SR score distributions in the electron (top row) and muon (bottom
row) channels for each mediator hypothesis. Black points are observed SR
data; blue histograms are the SR-direct SM background template; red curves
show the best-fit $S+B$ model.
}
\label{fig:sr_fit_results}
\end{figure}

As an additional robustness check, the likelihood fit was repeated using
only the SR score distributions without the VR normalisation constraint.
The fitted signal strengths remain positive with capped \(Z=8.0\), as expected
from the same high-\MET\ residual, while the resulting limits are compatible with those
obtained from the nominal simultaneous SR+VR fit.
The corresponding CL$_s$ scans and post-fit distributions are shown in
Figs.~\ref{fig:sr_cls_scans} and~\ref{fig:sr_fit_results}.
\begin{table}[h]
\centering
\caption{SR-only validation fit. The fitted signal strengths remain positive
with capped \(Z=8.0\), as expected from the same high-\MET\ residual, while
the resulting limits are compatible with those obtained from the nominal
simultaneous SR+VR fit. The \(Z\) values
are capped as in Table~\ref{tab:fit_summary}.}
\label{tab:appendix-sr-fit}
\resizebox{\textwidth}{!}{%
\begin{tabular}{lrrrrrr}
\toprule
Mediator & \(\hat{\mu}\) & \(\sigma_{\mu}\) & \(Z\) &
\(\mu^{95}_{\mathrm{obs}}\) & \(\mu^{95}_{\mathrm{exp}}\) &
68\% exp. band / 95\% exp. band \\
\midrule
Vector &
\(3.78\times10^{-2}\) &
0.00301 &
8.0 &
0.0443 &
0.0041 &
[0.00335, 0.00373] / [0.00322, 0.00393] \\
Axial-vector &
\(5.57\times10^{-2}\) &
0.00459 &
8.0 &
0.0669 &
0.0075 &
[0.00606, 0.00713] / [0.00562, 0.00779] \\
Scalar &
\(1.85\times10^{-2}\) &
0.00183 &
8.0 &
0.0230 &
0.0019 &
[0.00164, 0.00172] / [0.00159, 0.00176] \\
\bottomrule
\end{tabular}
}
\end{table}
\clearpage
\section{Additional Validation Figures}
\label{app:results-figures}
\subsection{\texorpdfstring{VR-extrapolated CL\(_s\) scans}{VR-extrapolated CLs scans}}
\label{app:figures-vr-closure}

\begin{figure*}[h]
\centering
\includegraphics[width=0.48\textwidth]{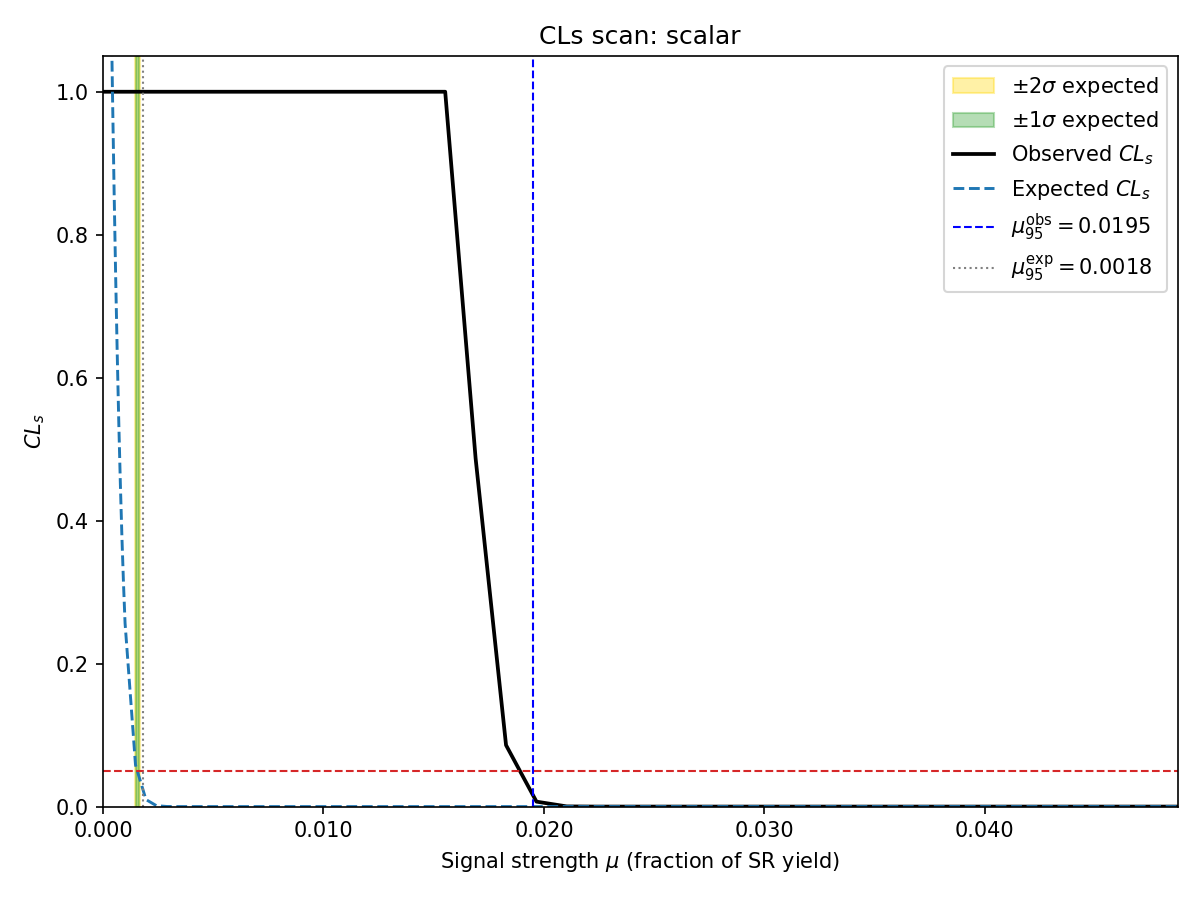}
\includegraphics[width=0.48\textwidth]{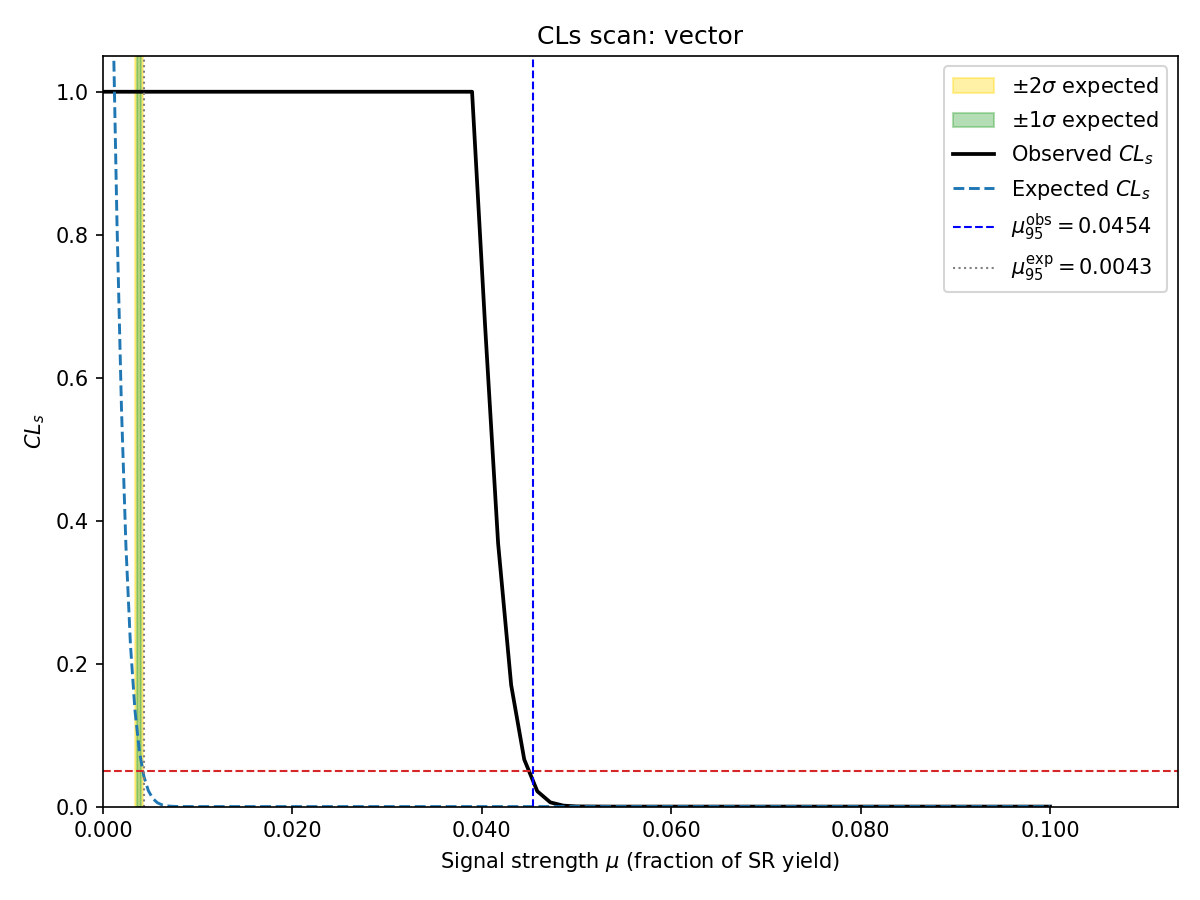}
\includegraphics[width=0.48\textwidth]{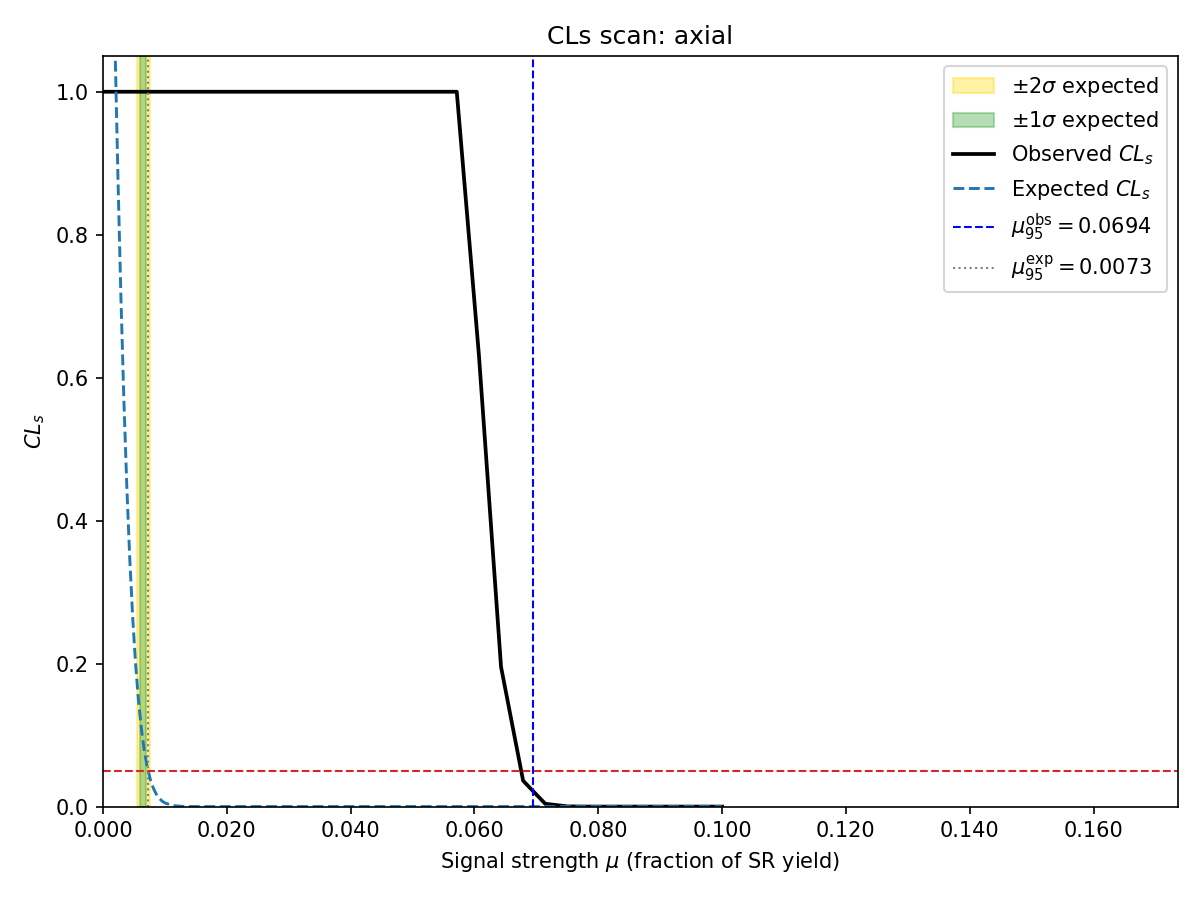}
\caption{CL$_s$ scans for the VR-extrapolated validation fit. Observed
CL$_s$ crosses the 95\% threshold for all three mediators once the scan
range brackets \(\hat{\mu}\); see Table~\ref{tab:appendix-vr-fit}.}
\label{fig:vr_cls_scans}
\end{figure*}
\clearpage
\subsection{VR-extrapolated post-fit distributions}
\label{app:figures-vr-postfit}

\begin{figure*}[h]
\centering
\includegraphics[width=0.48\textwidth]{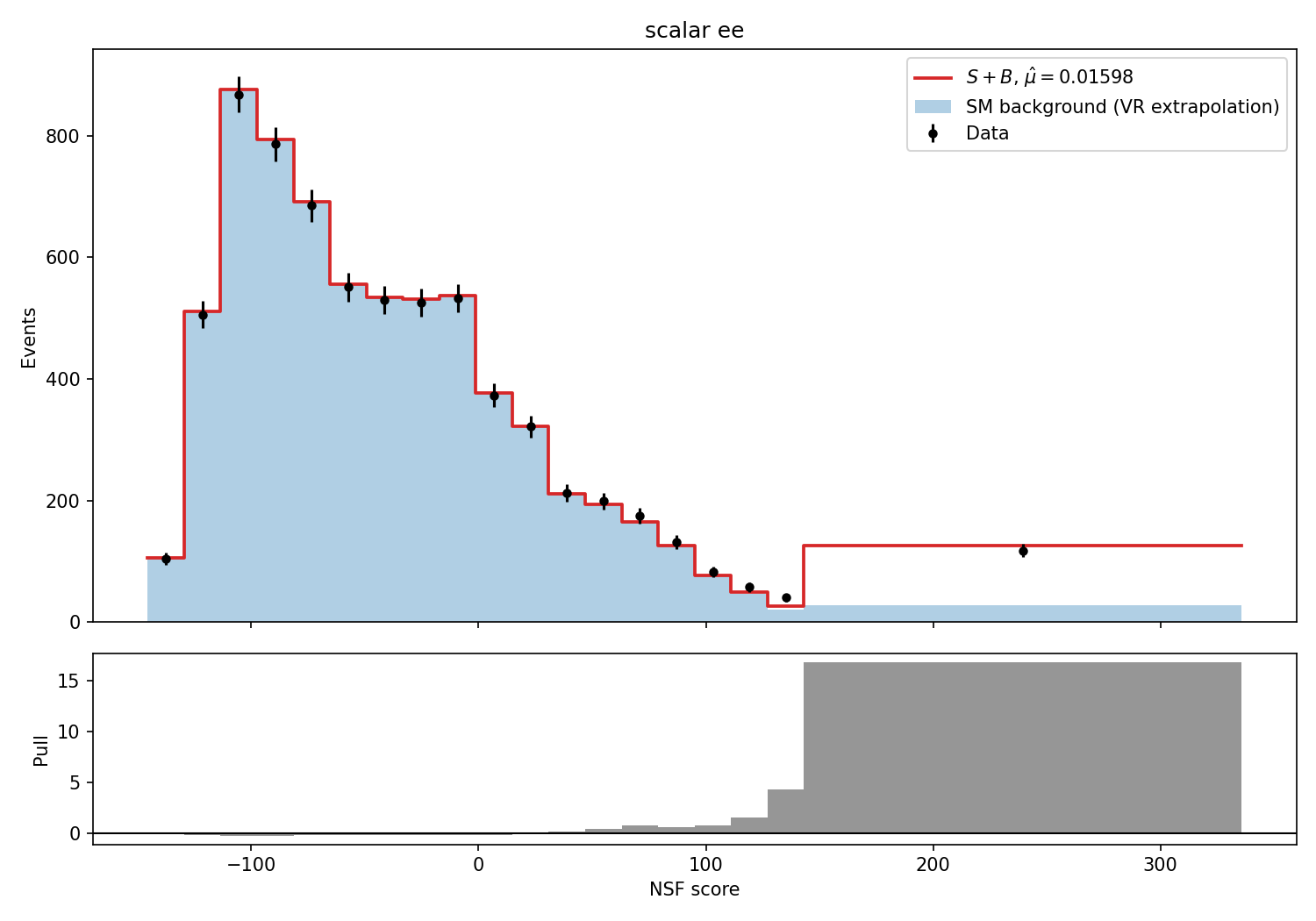}
\includegraphics[width=0.48\textwidth]{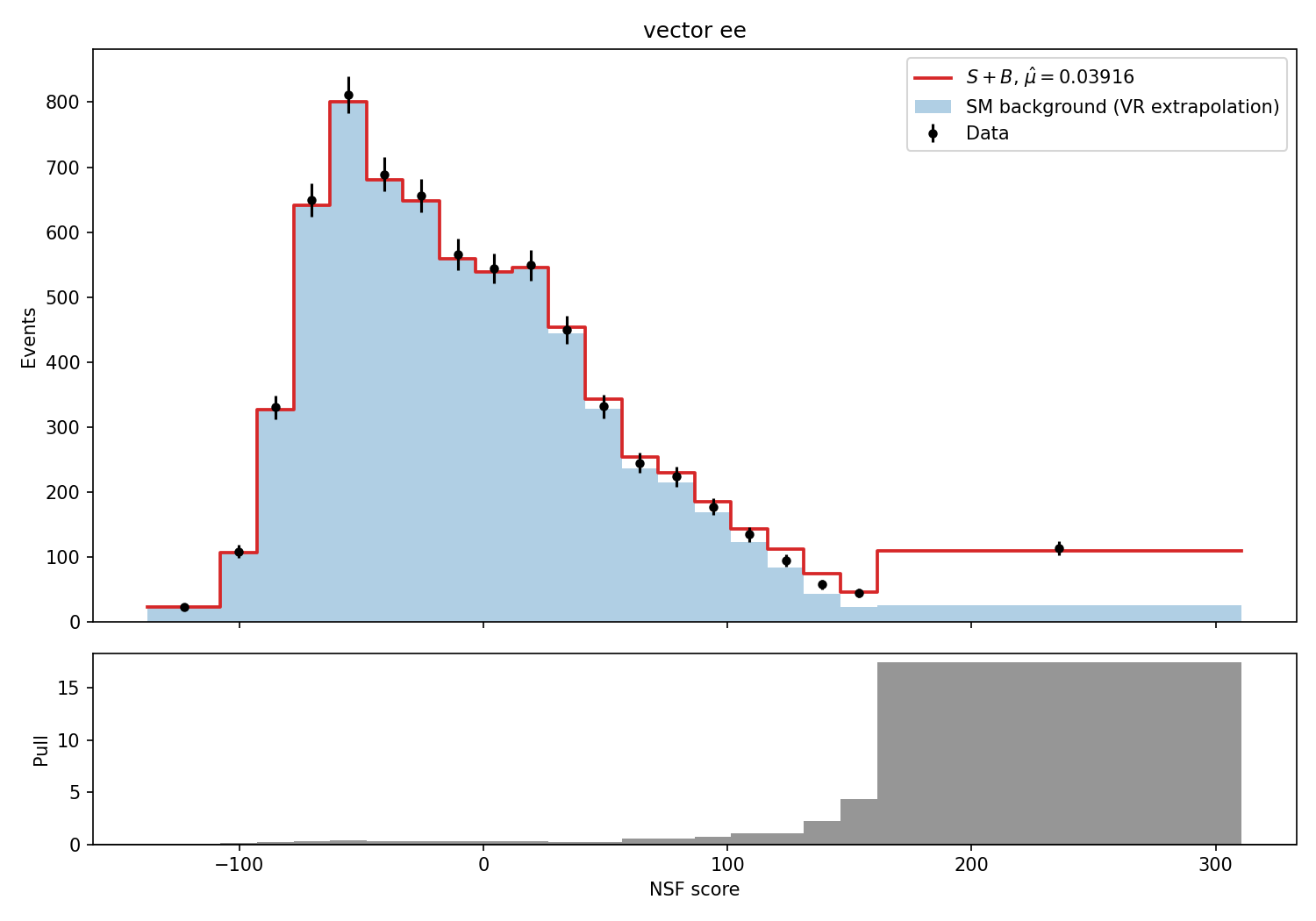}
\includegraphics[width=0.48\textwidth]{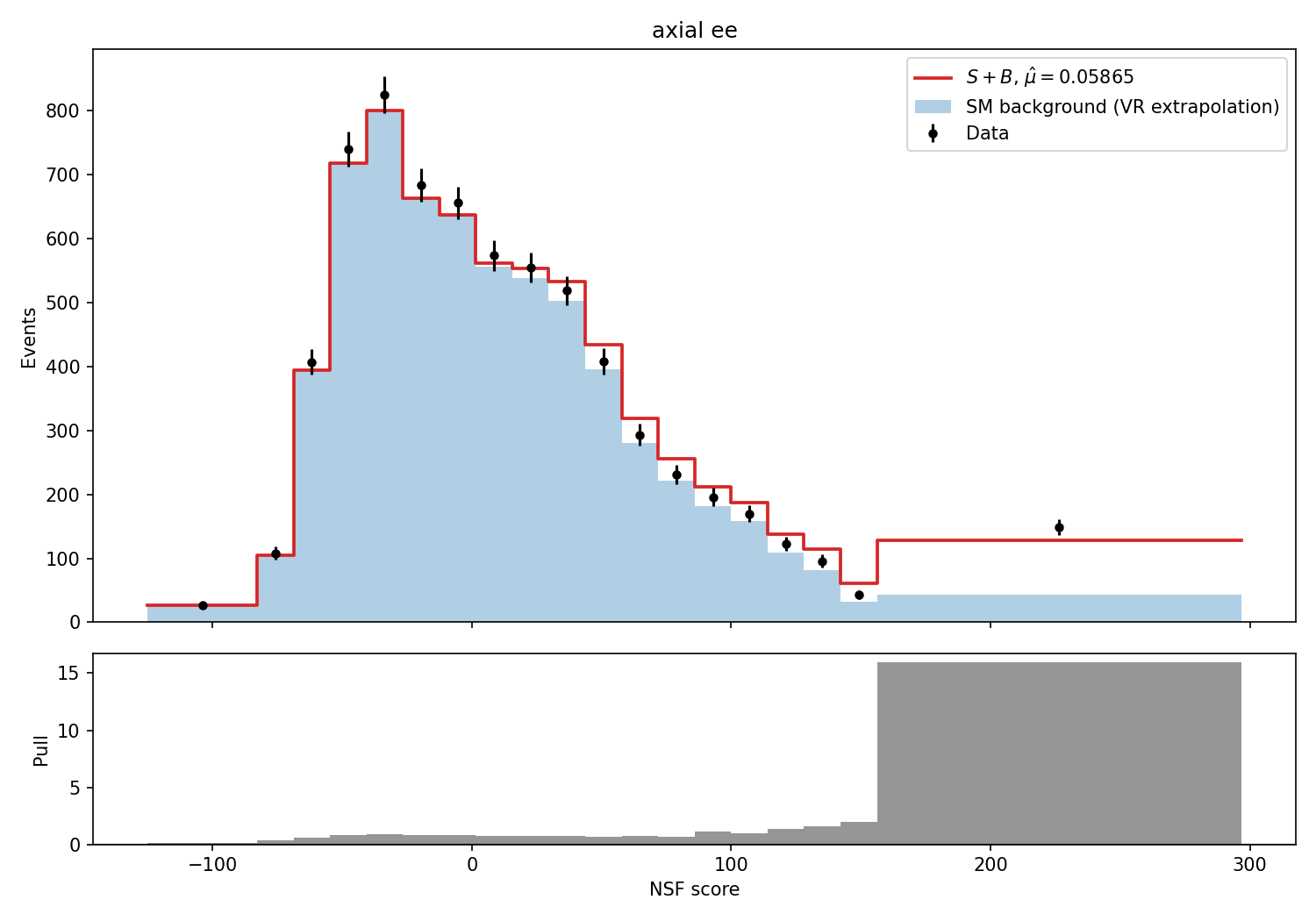}
\includegraphics[width=0.48\textwidth]{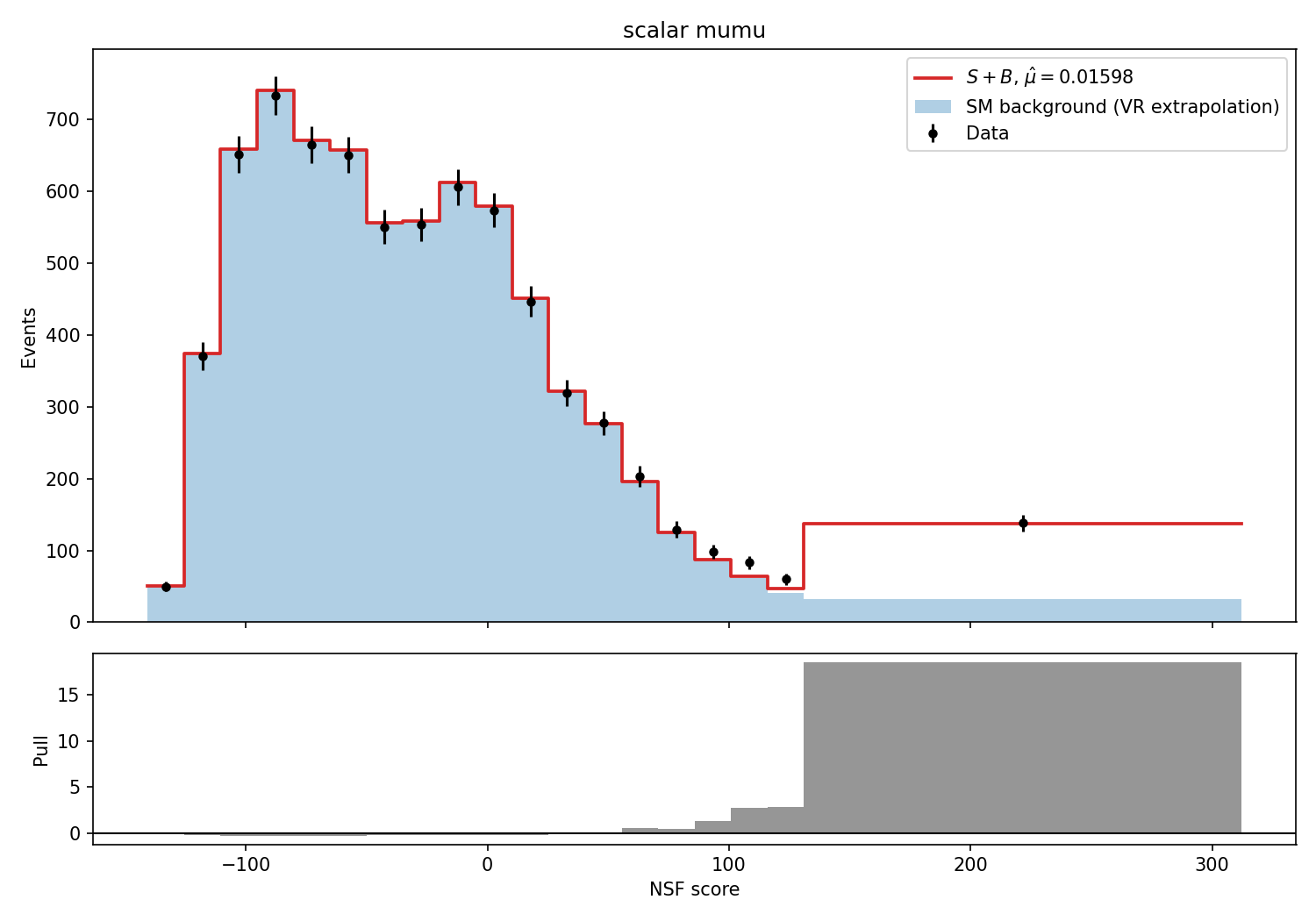}
\includegraphics[width=0.48\textwidth]{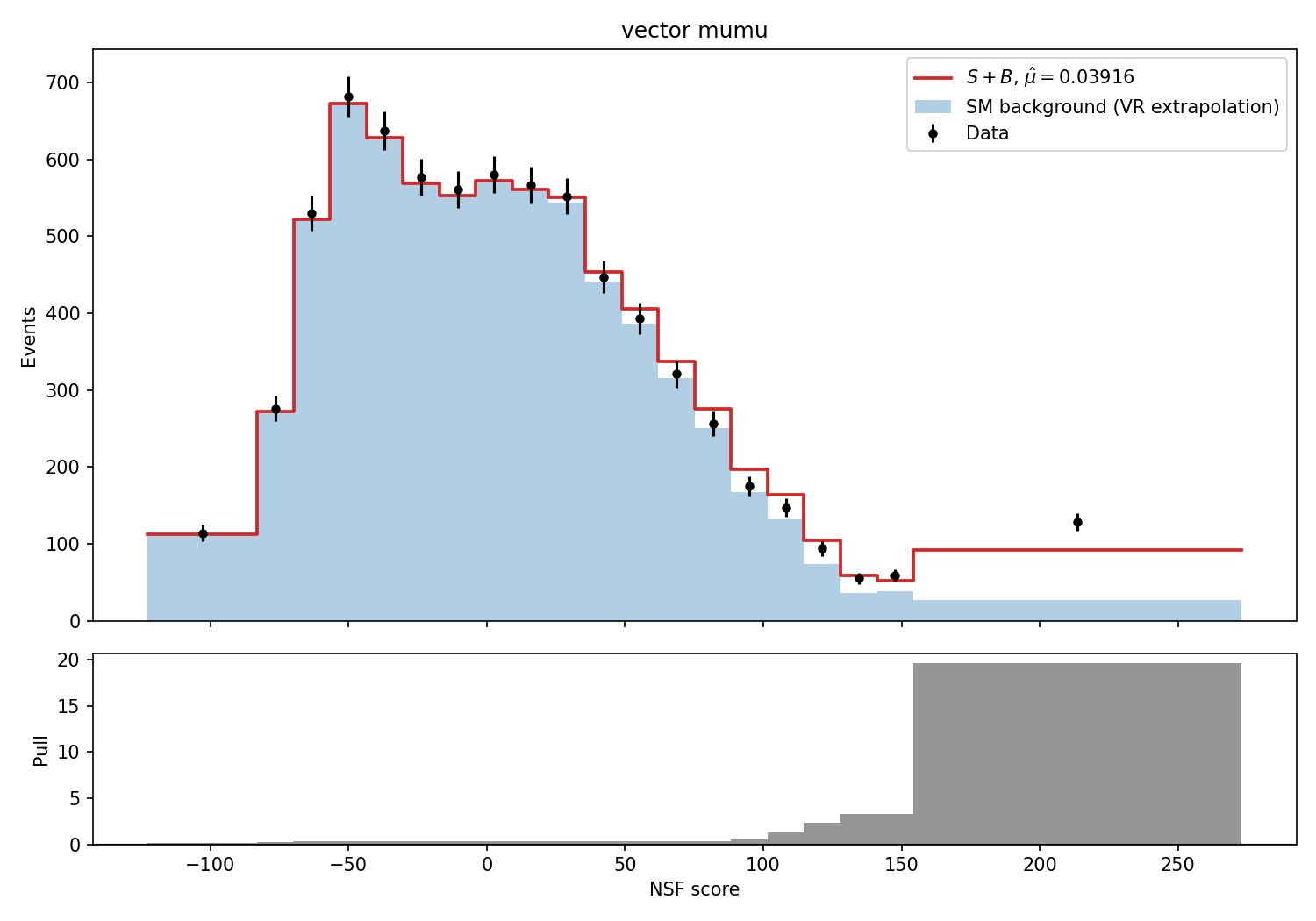}
\includegraphics[width=0.48\textwidth]{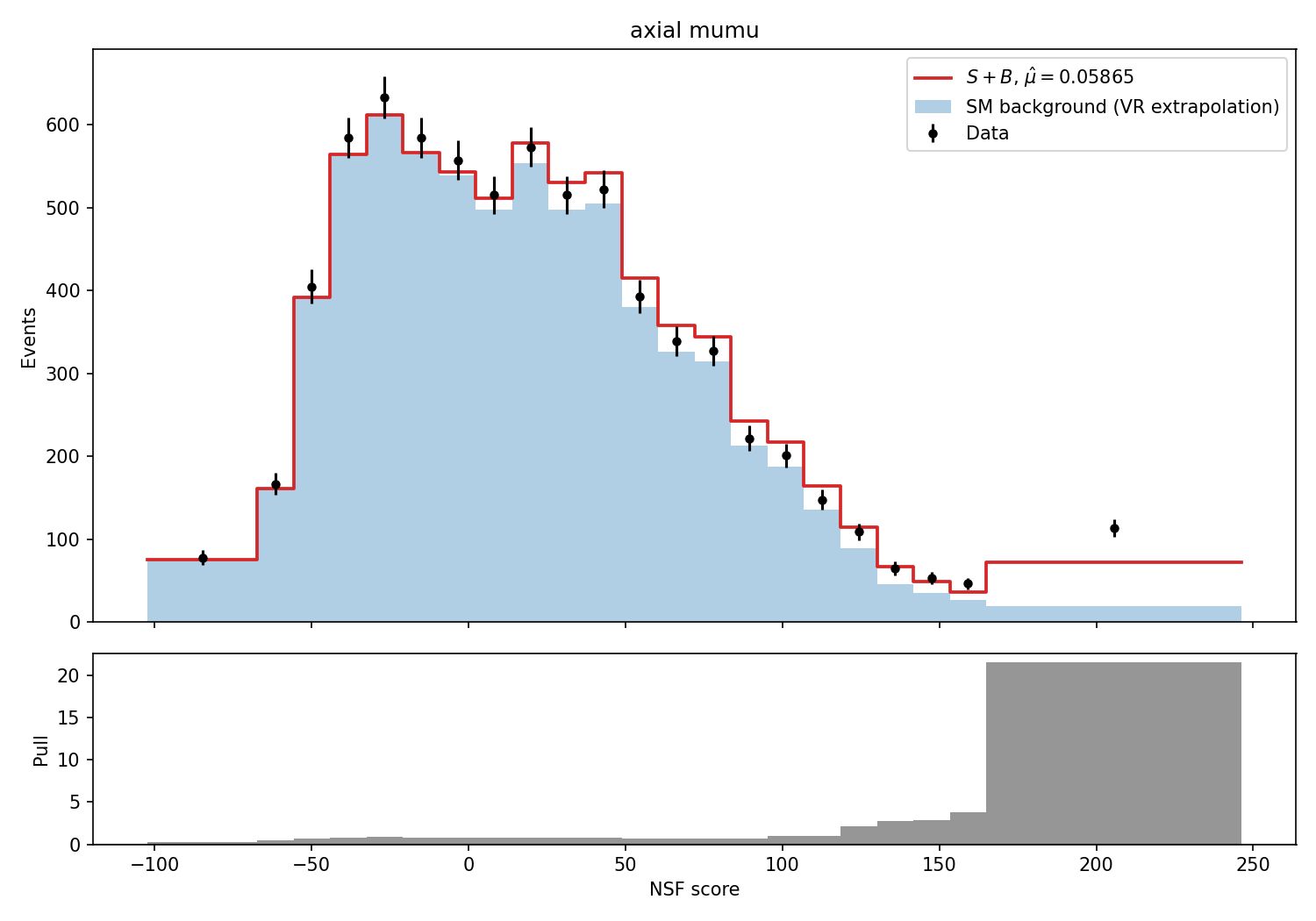}
\caption{Post-fit score distributions for the VR-extrapolated validation fit.
The signal-plus-background component compensates for the VR-to-SR high-score
tail mismatch, confirming that the fitted positive signal strength is a
background-modeling bias rather than evidence for DM in SM data.}
\label{fig:vr_postfit}
\end{figure*}
\clearpage
\subsection{\texorpdfstring{SR-only CL\(_s\) scans}{SR-only CLs scans}}
\clearpage
\subsection{SR-only post-fit distributions}
\clearpage
\twocolumn
\bibliographystyle{unsrt}
\bibliography{references}

@article{Goodman2011,
  author         = {Goodman, Jessica and Ibe, Masahiro and Rajaraman, Arvind and Shepherd, William and Tait, Tim M. P. and Yu, Hai-Bo},
  title          = {Constraints on Dark Matter from Colliders},
  journal        = {Phys. Rev. D},
  volume         = {82},
  pages          = {116010},
  year           = {2010},
  eprint         = {1008.1783},
  archivePrefix  = {arXiv},
  primaryClass   = {hep-ph},
  doi            = {10.1103/PhysRevD.82.116010}
}

@article{Bell2012,
  author         = {Bell, Nicole F. and Galea, Ahmad J. and Dent, James B. and Jacques, Thomas D. and Krauss, Lawrence M. and Weiler, Thomas J.},
  title          = {Searching for Dark Matter at the {LHC} with a Mono-{Z}},
  journal        = {Phys. Rev. D},
  volume         = {86},
  pages          = {096011},
  year           = {2012},
  eprint         = {1209.0231},
  archivePrefix  = {arXiv},
  primaryClass   = {hep-ph},
  doi            = {10.1103/PhysRevD.86.096011}
}

@article{Carpenter2013,
  author         = {Carpenter, Linda M. and Nelson, Andrew and Shimmin, Chase and Tait, Tim M. P. and Whiteson, Daniel},
  title          = {Collider searches for dark matter in events with a {Z} boson and missing energy},
  journal        = {Phys. Rev. D},
  volume         = {87},
  pages          = {074005},
  year           = {2013},
  eprint         = {1212.3352},
  archivePrefix  = {arXiv},
  primaryClass   = {hep-ph},
  doi            = {10.1103/PhysRevD.87.074005}
}

@article{Altmannshofer2015,
  author         = {Altmannshofer, Wolfgang and Fox, Patrick J. and Harnik, Roni and Kribs, Graham D. and Raj, Nirmal},
  title          = {Dark Matter Signals in Dilepton Production at Hadron Colliders},
  journal        = {Phys. Rev. D},
  volume         = {91},
  pages          = {015005},
  year           = {2015},
  eprint         = {1411.6743},
  archivePrefix  = {arXiv},
  primaryClass   = {hep-ph},
  doi            = {10.1103/PhysRevD.91.015005}
}

@article{Berlin2015,
  author         = {Berlin, Asher and Gori, Stefania and Lin, Tongyan and Wang, Lian-Tao},
  title          = {Pseudoscalar Portal Dark Matter},
  journal        = {Phys. Rev. D},
  volume         = {92},
  pages          = {015005},
  year           = {2015},
  eprint         = {1502.06000},
  archivePrefix  = {arXiv},
  primaryClass   = {hep-ph},
  doi            = {10.1103/PhysRevD.92.015005}
}

@article{Autran2015,
  author         = {Autran, Marcelo and Bauer, Kevin and Lin, Tongyan and Whiteson, Daniel},
  title          = {mono-{Z}$'$: searches for dark matter in events with a resonance and missing transverse energy},
  journal        = {Phys. Rev. D},
  volume         = {92},
  pages          = {115014},
  year           = {2015},
  eprint         = {1504.01386},
  archivePrefix  = {arXiv},
  primaryClass   = {hep-ph},
  doi            = {10.1103/PhysRevD.92.115014}
}

@article{DMForum2015,
  author         = {{ATLAS and CMS Collaborations}},
  title          = {Dark Matter Benchmark Models for Early {LHC} Run-2 Searches: Report of the {ATLAS/CMS} Dark Matter Forum},
  year           = {2015},
  journal        = {arXiv preprint},
  eprint         = {1507.00966},
  archivePrefix  = {arXiv},
  primaryClass   = {hep-ex},
  note           = {Editors: Antonio Boveia and Caterina Doglioni}
}

@article{Alves2015,
  author         = {Alves, Alexandre and Sinha, Kuver},
  title          = {Searches for Dark Matter at the {LHC}: A Multivariate Analysis in the Mono-{Z} Channel},
  journal        = {Phys. Rev. D},
  volume         = {92},
  pages          = {115005},
  year           = {2015},
  eprint         = {1507.08294},
  archivePrefix  = {arXiv},
  primaryClass   = {hep-ph},
  doi            = {10.1103/PhysRevD.92.115005}
}

@article{ATLAS2016Dilepton,
  author         = {{ATLAS Collaboration}},
  title          = {Search for high-mass new phenomena in the dilepton final state using proton--proton collisions at $\sqrt{s}=13$~{TeV} with the {ATLAS} detector},
  journal        = {Phys. Lett. B},
  volume         = {761},
  pages          = {372--392},
  year           = {2016},
  eprint         = {1607.03669},
  archivePrefix  = {arXiv},
  primaryClass   = {hep-ex},
  doi            = {10.1016/j.physletb.2016.08.052}
}

@article{Liew2017,
  author         = {Liew, Seng Pei and Papucci, Michele and Vichi, Alessandro and Zurek, Kathryn M.},
  title          = {Mono-{X} Versus Direct Searches: Simplified Models for Dark Matter at the {LHC}},
  journal        = {JHEP},
  volume         = {03},
  pages          = {100},
  year           = {2017},
  eprint         = {1612.00219},
  archivePrefix  = {arXiv},
  primaryClass   = {hep-ph},
  doi            = {10.1007/JHEP03(2017)100}
}

@article{ATLAS2017Dilepton,
  author         = {{ATLAS Collaboration}},
  title          = {Search for new high-mass phenomena in the dilepton final state using 36~{fb}$^{-1}$ of proton--proton collision data at $\sqrt{s}=13$~{TeV} with the {ATLAS} detector},
  journal        = {Phys. Lett. B},
  volume         = {776},
  pages          = {318--338},
  year           = {2018},
  eprint         = {1707.02424},
  archivePrefix  = {arXiv},
  primaryClass   = {hep-ex},
  doi            = {10.1016/j.physletb.2017.11.035}
}

@article{CMS2017MonoZ,
  author         = {{CMS Collaboration}},
  title          = {Search for new physics in events with a leptonically decaying {Z} boson and a large transverse momentum imbalance in proton--proton collisions at $\sqrt{s}=13$~{TeV}},
  journal        = {Phys. Rev. D},
  volume         = {97},
  pages          = {092005},
  year           = {2018},
  eprint         = {1711.00431},
  archivePrefix  = {arXiv},
  primaryClass   = {hep-ex},
  doi            = {10.1103/PhysRevD.97.092005}
}

@article{Yang2017,
  author         = {Yang, Daneng and Li, Qiang},
  title          = {Probing the Dark Sector through Mono-{Z} Boson Leptonic Decays},
  journal        = {Phys. Rev. D},
  volume         = {97},
  pages          = {015022},
  year           = {2018},
  eprint         = {1711.09845},
  archivePrefix  = {arXiv},
  primaryClass   = {hep-ph},
  doi            = {10.1103/PhysRevD.97.015022}
}

@article{Durkan2019,
  author         = {Durkan, Conor and Bekasov, Artur and Murray, Iain and Papamakarios, George},
  title          = {Neural Spline Flows},
  journal        = {Advances in Neural Information Processing Systems},
  volume         = {32},
  year           = {2019},
  eprint         = {1906.04032},
  archivePrefix  = {arXiv},
  primaryClass   = {stat.ML}
}

@article{Papamakarios2021,
  author         = {Papamakarios, George and Nalisnick, Eric and Rezende, Danilo Jimenez and Mohamed, Shakir and Lakshminarayanan, Balaji},
  title          = {Normalizing Flows for Probabilistic Modeling and Inference},
  journal        = {J. Mach. Learn. Res.},
  volume         = {22},
  pages          = {1--64},
  year           = {2021},
  eprint         = {1912.02762},
  archivePrefix  = {arXiv},
  primaryClass   = {stat.ML}
}

@article{Andreassen2020,
  author         = {Andreassen, Anders and Nachman, Benjamin and Shih, David},
  title          = {Simulation Assisted Likelihood-free Anomaly Detection},
  journal        = {Phys. Rev. D},
  volume         = {101},
  pages          = {055004},
  year           = {2020},
  eprint         = {2001.05001},
  archivePrefix  = {arXiv},
  primaryClass   = {hep-ph},
  doi            = {10.1103/PhysRevD.101.055004}
}

@article{CMS2021MonoZ,
  author         = {{CMS Collaboration}},
  title          = {Search for dark matter produced in association with a leptonically decaying {Z} boson in proton--proton collisions at $\sqrt{s}=13$~{TeV}},
  journal        = {Eur. Phys. J. C},
  volume         = {81},
  pages          = {13},
  year           = {2021},
  eprint         = {2008.04735},
  archivePrefix  = {arXiv},
  primaryClass   = {hep-ex},
  doi            = {10.1140/epjc/s10052-020-08739-5}
}

@article{Reyes2022,
  author         = {Reyes-Gonz{\'a}lez, Humberto and Torre, Riccardo},
  title          = {Testing the boundaries: Normalizing Flows for higher dimensional data sets},
  journal        = {SciPost Phys.},
  volume         = {13},
  pages          = {047},
  year           = {2022},
  eprint         = {2202.09188},
  archivePrefix  = {arXiv},
  primaryClass   = {hep-ph},
  doi            = {10.21468/SciPostPhys.13.2.047}
}

@article{Verheyen2022,
  author         = {Verheyen, Rob},
  title          = {Event Generation and Density Estimation with Surjective Normalizing Flows},
  journal        = {SciPost Phys.},
  volume         = {13},
  pages          = {047},
  year           = {2022},
  eprint         = {2205.01697},
  archivePrefix  = {arXiv},
  primaryClass   = {hep-ph},
  doi            = {10.21468/SciPostPhys.13.4.047},
  note           = {See arXiv:2205.01697 for publication details}
}

@article{Elgammal2024,
  author         = {Elgammal, S.},
  title          = {Angular distribution study for high mass dimuon pairs in {CMS} open 2012 data and for Mono-{Z}$'$ model},
  year           = {2024},
  eprint         = {2410.05755},
  archivePrefix  = {arXiv},
  primaryClass   = {hep-ex},
  journal        = {arXiv preprint}
}

@article{Cowan2011,
  author         = {Cowan, Glen and Cranmer, Kyle and Gross, Eilam and Vitells, Ofer},
  title          = {Asymptotic formulae for likelihood-based tests of new physics},
  journal        = {Eur. Phys. J. C},
  volume         = {71},
  pages          = {1554},
  year           = {2011},
  eprint         = {1007.1727},
  archivePrefix  = {arXiv},
  primaryClass   = {physics.data-an},
  doi            = {10.1140/epjc/s10052-011-1554-0}
}

@article{Nachman2020,
  author         = {Nachman, Benjamin and Shih, David},
  title          = {Anomaly Detection with Density Estimation},
  journal        = {Phys.\ Rev.\ D},
  volume         = {101},
  pages          = {075042},
  year           = {2020},
  eprint         = {2001.04990},
  archivePrefix  = {arXiv}
}

@article{CMS-LUM-2021,
  author         = {{CMS Collaboration}},
  title          = {Precision luminosity measurement in proton-proton collisions at $\sqrt{s}=13\,\mathrm{TeV}$ in 2015 and 2016 at CMS},
  journal        = {Eur.\ Phys.\ J.\ C},
  volume         = {81},
  pages          = {800},
  year           = {2021},
  eprint         = {2104.01927},
  archivePrefix  = {arXiv},
  doi            = {10.1140/epjc/s10052-021-09538-6}
}

@article{Hallin2021,
  author         = {Hallin, Anna and others},
  title          = {Classifying Anomalies Through Outer Density Estimation},
  journal        = {Phys.\ Rev.\ D},
  volume         = {106},
  pages          = {055006},
  year           = {2022},
  eprint         = {2109.00546},
  archivePrefix  = {arXiv}
}

@misc{CERNOpenDataDoubleMuon,
  author         = {{CMS Collaboration}},
  title          = {{DoubleMuon primary dataset in MINIAOD format from RunD of 2015 (/DoubleMuon/Run2015D-16Dec2015-v1/MINIAOD)}},
  year           = {2021},
  howpublished   = {CERN Open Data Portal, Record 24127},
  doi            = {10.7483/OPENDATA.CMS.H3TX.ZJZX},
  url            = {https://opendata.cern.ch/record/24127},
  note           = {DOI: 10.7483/OPENDATA.CMS.H3TX.ZJZX}
}

@misc{CERNOpenDataDoubleEG,
  author         = {{CMS Collaboration}},
  title          = {{DoubleEG primary dataset in MINIAOD format from RunD of 2015 (/DoubleEG/Run2015D-08Jun2016-v1/MINIAOD)}},
  year           = {2021},
  howpublished   = {CERN Open Data Portal, Record 24132},
  doi            = {10.7483/OPENDATA.CMS.6ULE.YZJW},
  url            = {https://opendata.cern.ch/record/24132},
  note           = {DOI: 10.7483/OPENDATA.CMS.6ULE.YZJW}
}

@misc{CERNOpenDataMonoZAxialMx10Mv20,
  author         = {{CMS Collaboration}},
  title          = {{Simulated dataset DarkMatter\_MonoZToLL\_A\_Mx-10\_Mv-20\_gDMgQ-1\_TuneCUETP8M1\_13TeV-madgraph in MINIAODSIM format for 2015 collision data}},
  year           = {2021},
  howpublished   = {CERN Open Data Portal, Record 16575},
  doi            = {10.7483/OPENDATA.CMS.YB3Q.XTXY},
  url            = {https://opendata.cern.ch/record/16575},
  note           = {DOI: 10.7483/OPENDATA.CMS.YB3Q.XTXY}
}

@misc{CERNOpenDataMonoZAxialMx50Mv200,
  author         = {{CMS Collaboration}},
  title          = {{Simulated dataset DarkMatter\_MonoZToLL\_A\_Mx-50\_Mv-200\_gDMgQ-1\_TuneCUETP8M1\_13TeV-madgraph in MINIAODSIM format for 2015 collision data}},
  year           = {2021},
  howpublished   = {CERN Open Data Portal, Record 16597},
  doi            = {10.7483/OPENDATA.CMS.34IE.KN6I},
  url            = {https://opendata.cern.ch/record/16597},
  note           = {DOI: 10.7483/OPENDATA.CMS.34IE.KN6I}
}

@misc{CERNOpenDataMonoZScalar,
  author         = {{CMS Collaboration}},
  title          = {{Simulated dataset DarkMatter\_MonoZToLL\_EWK\_Scalar\_Mx-100\_Lambda-3000\_TuneCUETP8M1\_13TeV-madgraph in MINIAODSIM format for 2015 collision data}},
  year           = {2021},
  howpublished   = {CERN Open Data Portal, Record 16601},
  doi            = {10.7483/OPENDATA.CMS.NW7F.NFGG},
  url            = {https://opendata.cern.ch/record/16601},
  note           = {DOI: 10.7483/OPENDATA.CMS.NW7F.NFGG}
}

@misc{CERNOpenDataMonoZVector,
  author         = {{CMS Collaboration}},
  title          = {{Simulated dataset DarkMatter\_MonoZToLL\_V\_Mx-1\_Mv-500\_gDMgQ-1\_TuneCUETP8M1\_13TeV-madgraph in MINIAODSIM format for 2015 collision data}},
  year           = {2021},
  howpublished   = {CERN Open Data Portal, Record 16630},
  doi            = {10.7483/OPENDATA.CMS.YTLH.E0N7},
  url            = {https://opendata.cern.ch/record/16630},
  note           = {DOI: 10.7483/OPENDATA.CMS.YTLH.E0N7}
}

\end{document}